\documentclass[]{interact}

\usepackage{epstopdf}
\usepackage[caption=false]{subfig}

\usepackage[natbibapa,nodoi]{apacite}
\renewcommand\bibliographytypesize{\fontsize{10}{12}\selectfont}
\usepackage[]{hyperref}

\theoremstyle{plain}

\theoremstyle{definition}

\theoremstyle{remark}

\usepackage{color, soul}
\usepackage{xcolor}
\usepackage{graphicx}
\usepackage{booktabs, multirow, makecell, colortbl}
\usepackage{lscape}
\usepackage{longtable}
\newcommand{\ra}[1]{\renewcommand{\arraystretch}{#1}} 

\definecolor{Gray}{gray}{0.9}

\begin{document}


\title{Towards Responsible AI: A Design Space Exploration of Human-Centered Artificial Intelligence User Interfaces to Investigate Fairness\footnote{This is the first draft version before being reviewed. The accepted version is on \url{https://www.tandfonline.com/doi/full/10.1080/10447318.2022.2067936?src=}.}}

\author{
Yuri Nakao\textsuperscript{a}\thanks{CONTACT Yuri Nakao Email: nakao.yuri@fujitsu.com}, \name{Lorenzo Strappelli\textsuperscript{b},  Simone Stumpf\textsuperscript{b}, Aisha Naseer\textsuperscript{c}, Daniele Regoli\textsuperscript{d} and Giulia Del Gamba\textsuperscript{d}}
\affil{\textsuperscript{a}Artificial Intelligence Laboratory, Fujitsu Laboraories Ltd., Kawasaki-city, Japan; \textsuperscript{b}Centre for HCI Design, City, University of London, London, UK;  \textsuperscript{c}Fujitsu Laboratories of Europe, Hayes, UK; \textsuperscript{d}Intesa Sanpaolo S.p.A., Turin, Italy}
}

\maketitle

\begin{abstract}
With Artificial intelligence (AI) to aid or automate decision-making advancing rapidly, a particular concern is its fairness. In order to create reliable, safe and trustworthy systems through human-centred artificial intelligence (HCAI) design, recent efforts have produced  user interfaces (UIs) for AI experts to investigate the fairness of AI models. In this paper, we provide a design space exploration that supports not only data scientists but also domain experts to investigate AI fairness. Using loan applications as an example, we held a series of workshops with loan officers and data scientists to elicit their requirements. We instantiated these requirement into FairHIL, a UI to support human-in-the-loop fairness investigations, and describe how this UI could be generalized to other use cases. We evaluated FairHIL through a think-aloud user study. Our work contributes to better designs to investigate an AI model's fairness---and move closer towards responsible AI.
\end{abstract}


\section{Introduction}
Artificial intelligence (AI) to aid or automate decision-making is advancing rapidly, and there are now many systems in use in jurisprudence, medicine and finance \citep{barocas2016BigData, Corbett-Davis2017Algorighmic, barocas-hardt-narayanan, chouldechova2017fair}. However, there have been increasing calls for a human-centred artificial intelligence (HCAI) design approach that is a part of a responsible AI development process to address concerns that automated decision-making is not very reliable, safe or trustworthy because they can ignore human-values or contexts that changes depending on the situations~\citep{shneiderman_human-centered_2020, shneiderman_responsible_2021}. HCAI proposes that considerations are made for how to integrate user control into tasks, how to involve stakeholders in responsible AI development, and how to design user interfaces (UIs) that allow users to interact with AI systems. 

In recent years, a number of UIs and technologies have been proposed that aim to involve stakeholders in investigating fairness of machine learning (ML) models by making them more transparent, such as AI Fairness 360 \citep{bellamy2018ai}, the What-If Tool \citep{Wexler2020What-IfTool}, FairSight~\citep{Ahn2020FairSight}, FairVis~\citep{Cabrera2019Fairvis}, SILVA~\citep{Yan2020SILVA}, and Fairlearn~\citep{bird2020fairlearn}. However, most of this strand of HCAI research has focused on enabling data scientists or ML experts to inspect and assess their models, rather than involving other stakeholders such as domain experts or end-users. This is especially important for AI fairness, as stakeholder groups may vary in human values underlying fairness~\citep{saxena2020fairness, Veala2018Fairness}, information needs, practices, and technical abilities, and these differences need to be reflected in the design of UIs that allow users to inspect fairness of AI~\citep{Veala2018Fairness, Lee2017AHuman-Centered}. 

In this work, we provide a design space exploration of UI components that allow both domain experts and data scientists to investigate fairness. We investigated this, 1) by conducting a series of workshops to better understand domain experts' and data scientists' practices and information needs for investigating fairness and produce a set of requirements, 2) by designing FairHIL which instantiates a suite of 7 UI components based on requirements that can support these stakeholders in investigating fairness through a human-in-the-loop approach, and 3) by evaluating FairHIL through a user study. We situated our work in the loan application domain, working directly with loan officers and data scientists employed by $<$anonymized$>$, a bank who partnered with us in these studies. 

Our aims were as follows:
\begin{itemize}
    \item to understand how these different groups of users assess AI fairness in loan applications in terms of process, criteria, informational needs, and transparency, and develop requirements for UI design;
    \item to design a suite of UI components to support data scientists as well as loan officers in assessing fairness and finding potential fairness issues in datasets that underlie ML models and also ML models developed from these datasets. We show how these UI components could be generalized for other domains;
    \item to evaluate the usefulness and usability of these UIs through a user study involving data scientists and loan officers. We reflect on what these results mean for implementing these UIs into an AI development process.  
\end{itemize}

We contribute to an evolving space of HCAI design with a focus on developing responsible AI. Our work concentrates specifically on fairness which is of great concern in increasingly automated decision-making. We advance the current state-of-the-art by 
\begin{itemize}
    \item identifying the requirements of domain experts and data scientists for investigating fairness;
    \item providing a set of UI components which support the processes and practices of these stakeholder groups;
    \item clarifying the space of design options for developing UIs that allow fairness to be investigated in other domains,
    \item evaluating an instantiation of these UI components in the loan application domain.
\end{itemize}  

Our paper is structured as follows. We first review the related work in ML fairness, paying particular attention to UIs that have been developed to interactively explore fairness in ML models. We then describe our first study with data scientists and loan officers and show their UI requirements for investigating fairness. We then present in detail UI components that we developed in order to help data scientists and domain experts to investigate fairness and find potential fairness issues. We provide the findings of a second user study that evaluates these UI components with data scientists and loan officers. We conclude with a discussion of the implications and limitations of our work and future research that is warranted.

\section{Related Work}
\subsection{Assessing AI Fairness}
\label{section:relwork_fairness}
Fairness is often equated with justice which comprises two main principles, liberty and equality, in which all people should be entitled to basic rights and freedoms, and should have an equal opportunity \citep{rawls1958justice}. Scholars have attempted to differentiate and taxonomize notions of justice \citep{Lee2017AHuman-Centered, Binns2018ItsReducing, Eckhoff74Justice, Cook1983Distributive}, including equity (i.e. equal distribution), procedural (i.e. relating to a fair decision-making process), interactional (i.e. relating to people's treatment), and informational (i.e. that a decision is well-justified) fairness. 

A common way of perceiving fairness is at the group and individual level~\citep{Karen2005DoingJustice}. For example, in US law, group fairness is embodied through the concepts of disparate treatment and disparate impact~\citep{barocas2016BigData}. The former is an intentional discriminatory treatment of groups or members of groups, and the latter, the process that causes a dissimilar impact on groups. These groups are often defined using `protected attributes' which are enshrined in law. For example, the United Kingdom Equality Act 2010 states that it is illegal to discriminate against someone because of their age, disability, gender reassignment, marriage and civil partnership, pregnancy and maternity, race, religion or belief, sex and sexual orientation~\citep{EqualityandHumanRightCommission2020Protected}.
Other countries have also moved to protect certain groups from unfair treatment. However, discrimination against a group can be less obvious. Even if protected attributes are not explicitly used in decision-making, discrimination might be indirect through the link of a protected attribute (e.g. race) with non-protected attributes (e.g. zip code). A different way of judging fairness is through individual fairness, which relates to the similar treatment or outcomes of two similar individuals~\citep{Cynthia2012Fairness}. Often, one difficulty that arises in individual fairness is defining how two people are similar.  Additionally, the trade-off between individual fairness and the group fairness widely used in legal situations also have been considered as a difficult issue to solve~\citep{Cynthia2012Fairness}.

Unfairness can manifest in ML systems through biases. The ML development process presents numerous opportunities for bias to infiltrate, and biases can eventually play a role in making a decision~\citep{barocas2016BigData, Dodge2019Explaining, Hajian2016AlgorithmicBias, Olteanu2019SocialData, Friedman1996Bias, Caliskan2017Semantics}, for example, through the choice and characteristics of the dataset, feature engineering and selection, or choice of learning model. A well-known study underscores this point:  COMPAS (Correctional Offender Management Profiling for Alternative Sanctions), a system used in the US justice system, was found to predict Black defendants as a much higher risk of violent recidivism than their White counterparts~\citep{Larson2016HowWe}. Because a lot of these ML systems are not accountable or transparent, there is a strong need to investigate and then potentially mitigate fairness issues in ML. 

To measure fairness in AI systems, a number of fairness measures have been proposed. In the last few years over twenty different types have been identified \citep{narayanan2018translation, verma2018fairness}, for example:
\begin{itemize}
    \item Individual fairness: similar outcomes for similar individuals \citep{Cynthia2012Fairness}.
    \item Group fairness (statistical/demographic parity): people in both protected and unprotected groups have equal probability of having the positive outcome \citep{kamiran2009classifying, verma2018fairness}.
    \item Subgroup fairness (intersectional bias):  a combination of multiple sensitive features lead to an unfair result, which would have otherwise been considered fair when looking at them individually \citep{kearns2018preventing, yang2020fairness}.
    \item Counterfactual fairness: a comparison between predictions concerning an individual and its ``counterfactual self'' with different values for protected attributes \citep{kusner2017counterfactual}.
    \item Equalized Odds: same error rates (false positive rate and false negative rate) in groups with different values of protected attributes \citep{hardt2016equality}.
    \item Fairness Through Unawareness:	sensitive features are not explicitely employed to make decisions~\citep{kusner2017counterfactual}.
    \item Predictive Parity: same precision for groups with different values of protected attributes \citep{verma2018fairness}. 
\end{itemize}

These fairness measures can then be used to develop ways to mitigate or remove bias or unfairness. There are a number of toolkits available to investigate and mitigate fairness issues, for example, IBM’s AI Fairness 360~\citep{aif360-oct-2018} and Fairlearn~\citep{bird2020fairlearn}. There are various approaches for bias mitigation, for example, manipulating or cleaning observed data, reweighting observations, suppressing features correlated with protected attributes, or trying to remove from the data all the information of protected attributes while trying to keep as much information as possible from other variables \citep{kamiran2009classifying, zemel2013learning, calmon2017optimized}. Bias mitigation can also include adopting training approaches such as adversarial training~\citep{zhang2018mitigating}, regularization~\citep{kamishima2011fairness}, or reductions approach~\citep{agarwal2018reductions}. Finally, it is also possible to tweak the final outcomes in order to satisfy the desired fairness metric \citep{hardt2016equality, barocas-hardt-narayanan}.

Despite the exploration of fairness-aware ML, there has been little agreement which measures to use \citep{verma2018fairness} because of trade-offs between them \citep{verma2018fairness, barocas2016BigData, barocas-hardt-narayanan, chouldechova2017fair}. It has also been argued that fairness measures are poor proxies for identifying discrimination \citep{narayanan2018translation, corbettdavies2018measure, Veala2018Fairness}, and that therefore statistical bias, based on fairness measures, and societal bias, based on human values, need to be differentiated \citep{mitchell2018prediction, narayanan2018translation}. 

Whether a decision or a system is judged as fair depends very much on human values embedded in individuals and society \citep{saxena2020fairness}, and there have been increasing research efforts to investigate human fairness values empirically. For example, it has been found that while fairness outputs were perceived by participants as statistically fair, some participants were not comfortable with an AI system making high-stakes decisions without a human in the loop \citep{Binns2018ItsReducing}. In addition, stakeholders can vary in what they perceive as fair \citep{narayanan2018translation}. A spectrum of variables has been identified that influence fairness judgements in individuals. These include education, computer literacy \citep{Wang2020Factors}, gender \citep{Wang2020Factors, Mallari2020DoILook}, culture and region \citep{Bolton2010.HowDoPriceFairness, KIM200783FormingandReacting, Berman1985CrossCultural, MATTILA2006CrossCulturalComparison, Blake2015NatureOntologyOfFairness, hofstede_dimensionalizing_2011}, and environment and stakeholder role~\citep{Lee2017AHuman-Centered, Veala2018Fairness}. Hence, it is very important to study how stakeholders investigate fairness in different domains, how stakeholder groups differ within a domain, and also how to better involve them through human-in-the-loop UIs, as we do in this work.

\subsection{User Interfaces to Investigate Fairness}
\label{section:relwork_UIs}
  There has been a recent influx of fairness UIs that look to address these questions by introducing a human-in-the-loop approach in the development of ML models. Many of these efforts are based on work in Explainable AI (XAI) to communicate models and decisions to users, for example, through model-agnostic tools such as LIME \citep{Ribeiro2016WhyshouldI} and SHAP \citep{Lundberg2017UnifiedApproach}. LIME looks to explain model predictions on an local level, identifying the features that are driving the prediction \citep{Ribeiro2016WhyshouldI}. SHAP can explain the contribution of features to the model \citep{Lundberg2017UnifiedApproach}. Coupled with data visualisation, XAI approaches are often used to identify potential bias, for example, Dodge et al.~\citep{Dodge2019Explaining} investigated global and local explanation styles in fairness assessments.  

There are also now a number of toolkits being developed to support interactive exploration of fairness in datasets and ML models. We review five of the most well-known and recent ones in detail. 

\subsubsection{What-if Tool} 
\begin{figure}
\centering
\subfloat[The What-if Tool UI~\citep{Wexler2020What-IfTool}]{
\resizebox*{6.5cm}{!}{\includegraphics{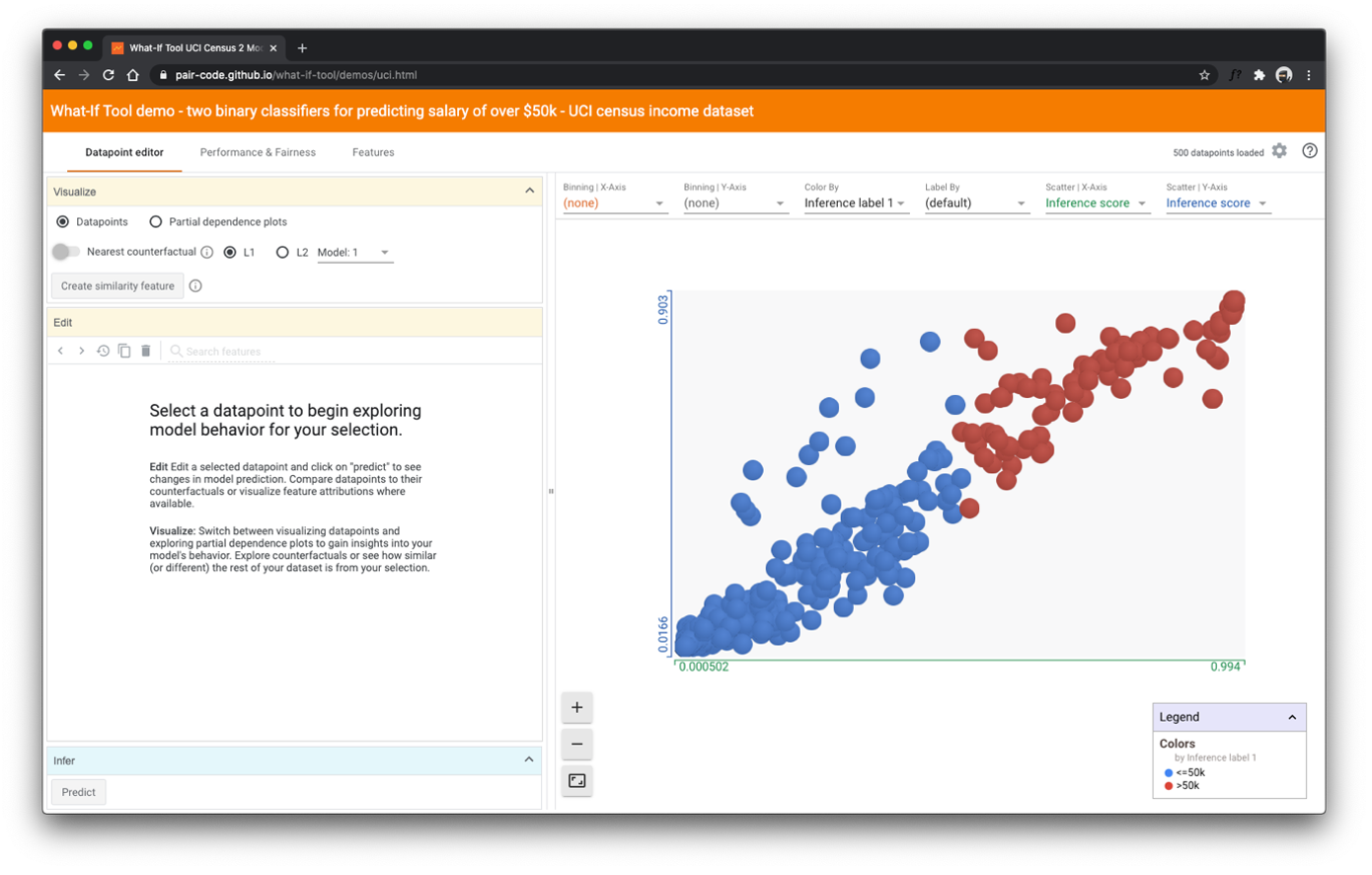}}}\hspace{5pt}
\subfloat[FairVis UI~\citep{Cabrera2019Fairvis}]{
\resizebox*{6.5cm}{!}{\includegraphics{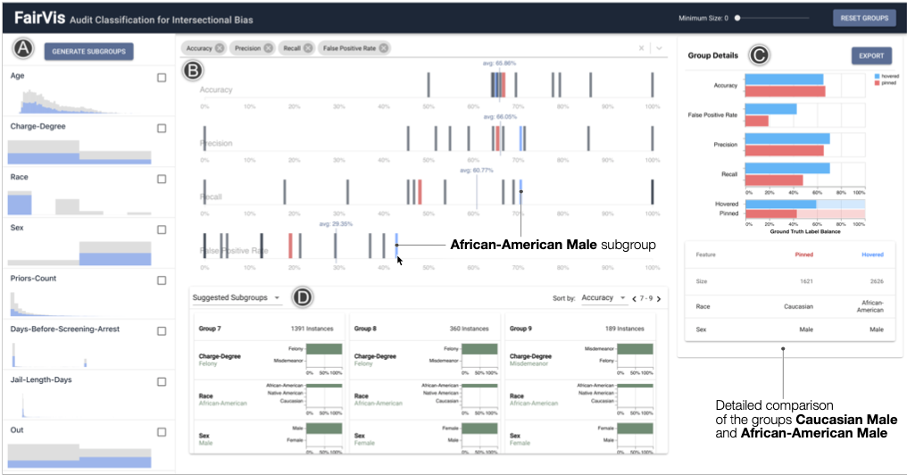}}}\hspace{5pt}
\subfloat[FairSight UI~\citep{Ahn2020FairSight}]{
\resizebox*{6.5cm}{!}{\includegraphics{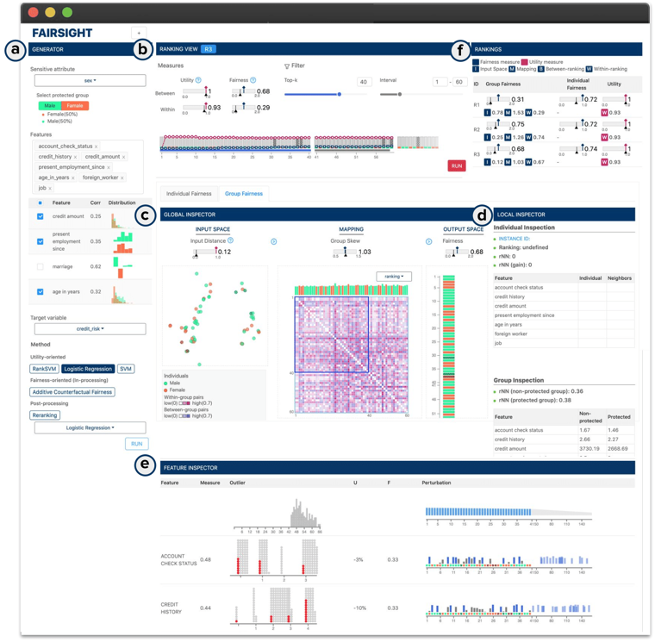}}}\hspace{5pt}
\subfloat[Silva UI~\citep{Yan2020SILVA}]{
\resizebox*{6.5cm}{!}{\includegraphics{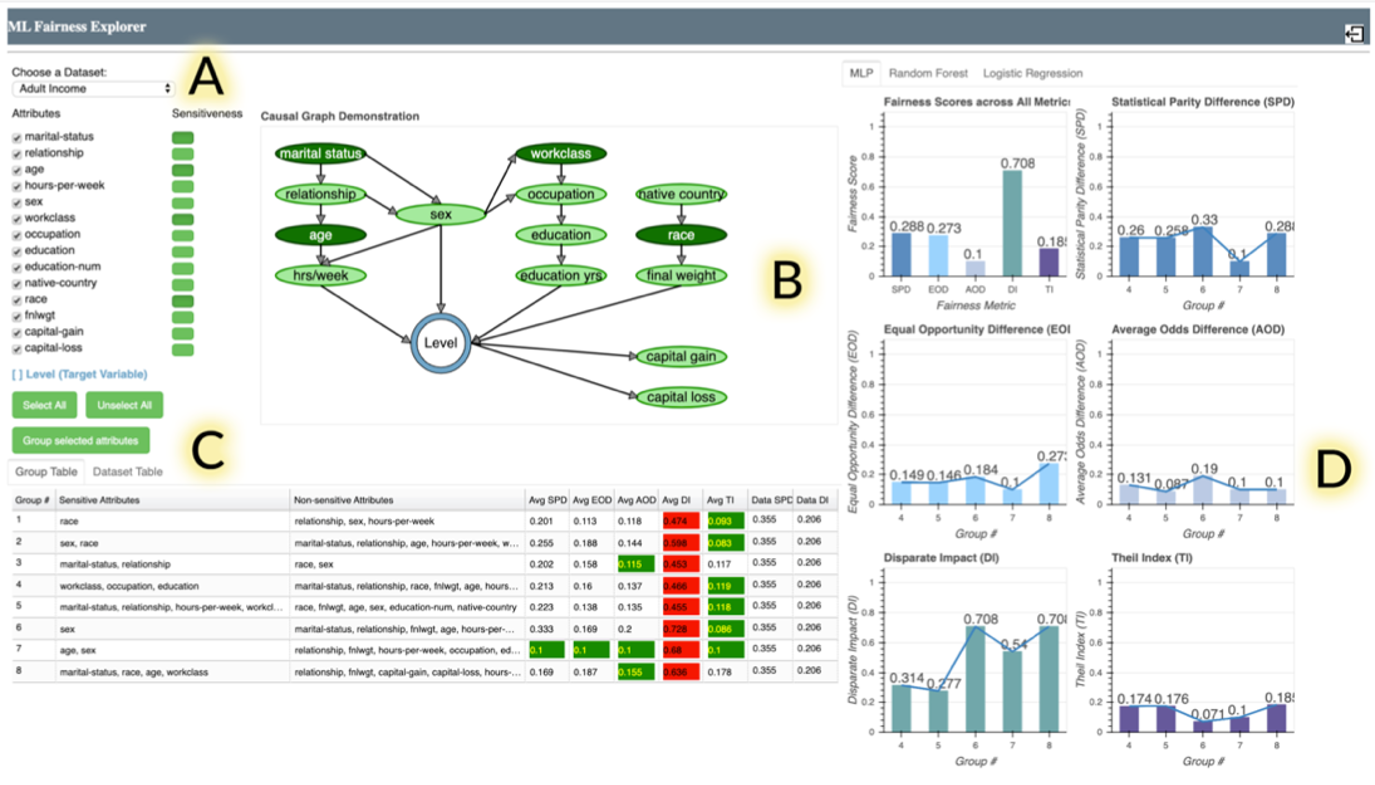}}}\hspace{5pt}
\subfloat[Fairness Elicitation Interface~\citep{cheng_soliciting_2021}]{
\resizebox*{6.5cm}{!}{\includegraphics{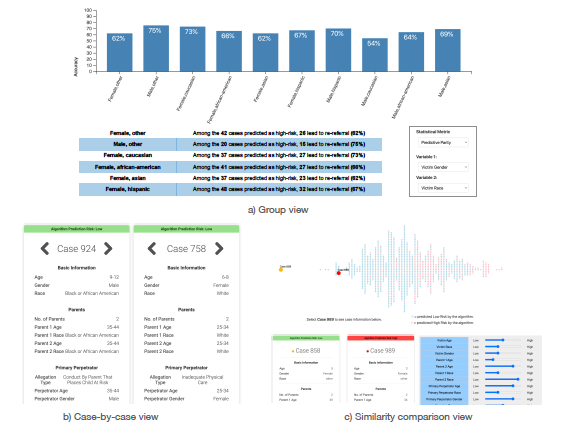}}}\hspace{5pt}
\caption{Human-in-the-loop Fairness Toolkits}  
\label{fig:FairnessUIs}

\end{figure}

The What-If Tool~\citep{Wexler2020What-IfTool} has been developed for data scientists to visualise fairness in ML (Figure~\ref{fig:FairnessUIs} a). The UI is comprised of three screens: datapoint editor, performance and fairness, and features. The datapoint editor facilitates on-the-fly visualizations of the dataset or model. Users are easily able to bin, position, color and label data points. Features can be sliced to investigate group fairness and combined for subgroup fairness. Data points can be directly investigated and compared through counterfactual reasoning; by allowing users to select a data point, visualise its counterfactual and adjust its values, they are able to see ‘what could have happened’ and subsequently infer a reason to an outcome. Through color encoding, users are also able to see decision boundaries between the (red and blue) points. The performance and fairness screen provides the user with a selection of six performance and fairness metrics. The features screen provides users with more descriptive statistics of the features in the dataset, such as their count, mean, median, and standard deviation. 

\subsubsection{FairVis}
FairVis \citep{Cabrera2019Fairvis} (Figure~\ref{fig:FairnessUIs} b) is a UI developed for data scientists to audit their models before deployment. In addition to comparing fairness between two or more groups, it also allows users to explore intersectional bias, i.e., identifying subgroup discrimination. The UI is made up of four interconnected panels. The first panel enables users to see all features (and their distributions), select them, or drill down further to select specific target classes. The second panel provides users with the ability to apply a number of predefined metrics such as accuracy, precision and recall from a drop-down list. Strip plots for each metric are displayed below. By clicking or hovering on subgroups, the user is able to see, through color encoding, the subgroup's position across the selected metrics and the four panels, therefore, comparing and contrasting the subgroup across the dataset. To explore further, the user is able to click and pin their subgroups of interest and compare metric outputs in the third panel. The fourth and last panel facilitates further exploration through a carousel of suggested subgroups, which showcases features and their classes (sorted by chosen metric) as percentile bar charts.

\subsubsection{FairSight}
FairSight \citep{Ahn2020FairSight} provides users familiar with ML with a UI that embeds a workflow to ensure fair decision-making.  The FairSight UI (Figure~\ref{fig:FairnessUIs} c) is made up of six panels; the first, ‘generator’, supports the user in setting up a model by defining sensitive features, features (including showing their distribution and correlation with the target), target and the selecting a prediction algorithm from a pre-defined list. After running the model, group and instance-based performance and fairness metrics are shown in the Ranking View, Rankings, and Global Inspector panels. The Local Inspector and Feature Inspector  panels facilitate the investigation of a specific data point and its nearest neighbor, and the features that might contribute to bias. 

\subsubsection{Silva}
 Silva \citep{Yan2020SILVA} is a UI to support the AI development workflow. Crucially it is claimed that this UI is useful for both data scientists and 'novices' unfamiliar with fairness analysis. Its main feature is a causal graph generated by structural learning algorithms (Figure~\ref{fig:FairnessUIs} d). The graph enables exploration of the features (nodes) and the relationships (edges) between them to identify potential bias in a classifier. The interface is made up of four panels; the first enables users to select and group features of interest and toggle their sensitiveness. The second panel shows the causal graph. Groups and their fairness metrics are shown in the third panel. The user is also able to access the dataset. The final panel displays the fairness metrics scores of each group for each of the machine learning algorithms.

\subsubsection{Fairness Elicitation Interface}
A UI, coupled with an interview protocol, has been proposed to elicit fairness notions from stakeholders who have no technical background in ML \citep{cheng_soliciting_2021} (Figure~\ref{fig:FairnessUIs} e).The investigations for this UI were focused on predicting child maltreatment cases, and involved social workers and parents. The UI consists of a group view which shows differences between variable groups along fairness metrics and allows division into subgroup by a further variable; a case-by-case view which shows a pair of cases and their prediction and AI model features; and a similarity comparison view which shows a scatterplot that compares a selected case with all other cases, calculated through feature similarity using weighted Euclidean distance. While this UI is encouraging as it targeted at stakeholders other than data scientists, its aim is only to function as a supportive tool to elicit fairness notions instead of providing comprehensive support to investigate fairness of an AI model.

\ \ \ \ \textbf{[Figure 1 near here]}

\subsubsection{Overview of Emerging Fairness UI Design Space}

While every tool works in different ways to support a user to explore the fairness of an AI model, some common UI functionality is emerging designed to cover various functional and informational needs. For example, all tools allow the data to be grouped in some way. Many tools either display performance or fairness metrics, or both. Table \ref{tab:UIComarison} shows a mapping of functionality to the tools we have described in 2.2.

\ \ \ \ \textbf{[Table 1 near here]}

Our work explores a design space of UI component and associated functionality to investigate fairness. We aim to support a variety of stakeholder groups that differ in their technical ability with respect to AI, and experience with formalized notions of fairness. In order to ground our exploration, we started with an empirical investigation of stakeholder functional and informational needs to investigate fairness in the loan applications domain.

\section{Study 1: Investigating stakeholder needs to assess fairness}
To explore user needs, we worked with two stakeholder groups: loan officers and data scientists employed in a partner bank, $<$anonymised for review$>$ based in Italy. Loan officers act as intermediaries between the bank and the customer, and are domain experts in this area but typically have little formal knowledge of AI or fairness metrics. Data scientists have experience in developing ML models for banking systems and processes and will typically have come across some fairness metrics. 

 We conducted two workshops with each stakeholder group to inform the exploration of this design space and uncover functional and informational needs. Due to Covid-19 restrictions, all research was conducted remotely using online meeting software. 

Workshop 1 focused on how stakeholders investigate fairness, and what information that they consider important and need. It also exposed them to some initial potential UI components, based on our literature review, as design provocations. Workshop 1 was analyzed, and the insights informed the development of an initial prototype. This prototype was then discussed in Workshop 2, in order to obtain early feedback on more detailed designs while also being open to extending functionality that we had missed previously or needs that we had misunderstood in Workshop 1. Based on the requirements and feedback from the workshops, we produced a final set of UI components (section 4) for the evaluation study (section 5). 

\subsection{Participants}
We recruited 6 loan officers (5 men, 1 woman, mean age $=36.5$, stdev $=3.39$) and 6 data scientists (3 men, 3 women, mean age $= 29.7$, stdev $=4.13$). For further background details, see Table \ref{table:participants}. All participants were over 18 years old and fluent in English. We excluded participants with any severe visual or hearing impairments, as it would have been difficult to make accommodations for these impairments in the study setup. 
Five of the 6 loan officers had previous experience with granting retail loans in bank branches across Italy, one of which had served as branch manager. The remaining participant had previously worked in Zagreb, Croatia in another role but moved to Italy where he worked in retail and wealth management for a year. Four of the 6 data scientists were very involved in considering AI fairness as part of their role. 

\ \ \ \ \textbf{[Table 2 near here]}

No incentives were offered for participation. This study was approved by the $<$anonymized$>$ Research Ethics Committee.

\subsection{Setup}
Separate workshops were held for data scientists and loan officers. 
All 6 data scientists were in one online workshop, and the 6 loan officers were in another. The sessions were run as semi-structured discussions, with the flexibility to explore relevant areas in more detail. This meant that data scientists and loan officers worked together in their respective groups and discussed their approaches and views for investigating fairness. 

\subsubsection{Workshop 1}
The aim of workshop 1 was to conduct user research into how fairness was investigated by these data scientists and loan officers, their functional and informational requirements, and to expose them to initial conceptual designs. Workshop 1 took place in September 2020 for each group and lasted 2 hours. These 2 hours were structured into 3 parts: 1) a general discussion about fairness in AI, 2) a detailed investigation of the fairness of decisions in a given dataset, and 3) initial reactions to causal reasoning graphs. 

In part 1 (approx. 30 minutes), participants had a general discussion about fairness, for example, what makes decisions in loan applications fair or unfair, and the role of ML in loan decision-making and fairness. This was to familiarize participants to the topic at hand, and to discover any general fairness values that they paid attention to.

In part 2 (approx. 1 hour), participants were asked to investigate fairness for a dataset we provided to them using their own preferred way. The dataset was developed by $<$anonymized for review$>$, a partner bank in this project. The dataset contained 1000 anonymized but real loan applications and their decisions, with 26 features. These features include demographic information of the applicant (age, gender, nationality, etc), financial information (household income, insurance, etc), loan information (amount of loan requested, purpose of loan, loan duration, monthly payments, etc), as well as some information of their financial and banking history (years of service with the bank, etc). There were also some features that related to internal bank procedures, such as a money laundering check and a credit score developed by the bank. The dataset was sent ahead of the workshop so that participants had time to familiarize themselves with it (they could but did not have to have a look at the dataset before the workshop). We created a data visualization tool, providing the ability to slice the features and present them using various chart types such as histograms, scatter plots, bar graphs and a strip plot, in order to answer questions that participants had about the data on-the-fly.  We screenshared this dataset and also provided a visualisation tool so that any questions they had could be answered on the fly, in order to further investigate the fairness of the dataset. They were asked to discuss their individual approaches of how they would examine this dataset. As such, there was no set flow; it was all directed by the participants. 

\begin{figure}
     \centering
     \includegraphics[width=.8\textwidth]{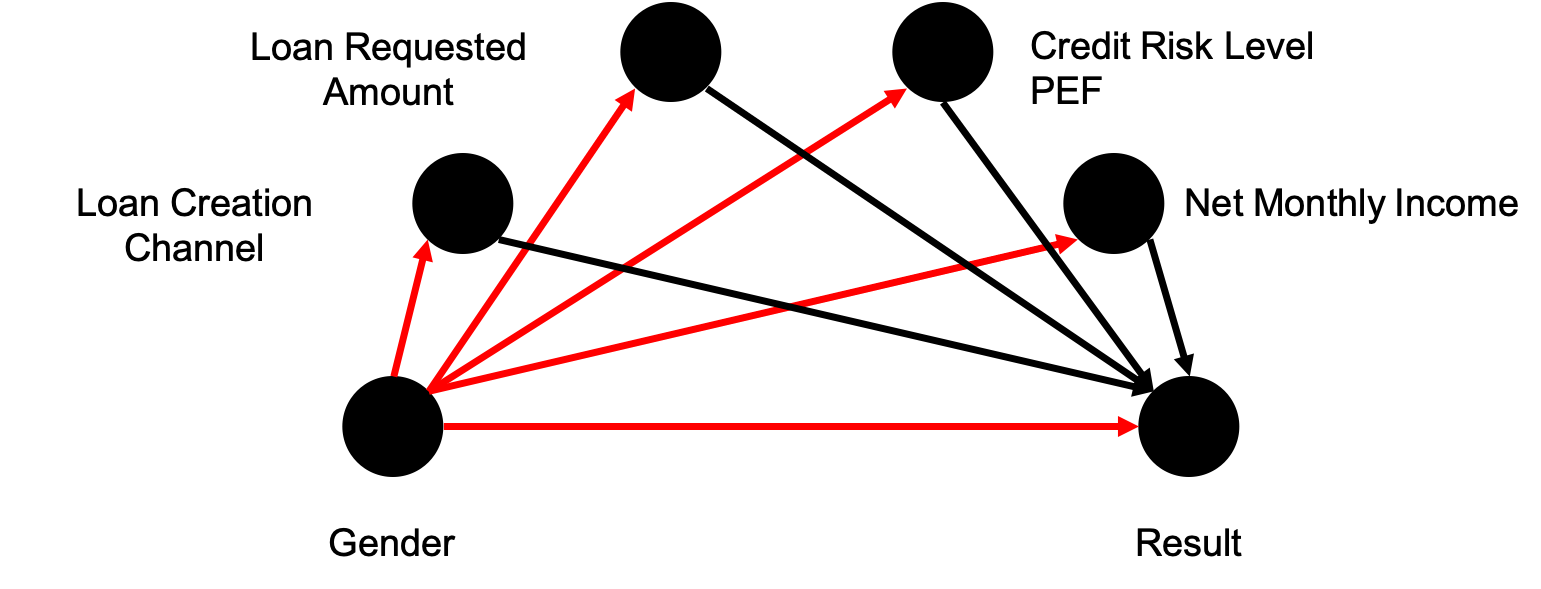} 
     \caption{Workshop causal graph example.} 
     \label{fig:workshop causal graph example}
 \end{figure}

In part 3 (approx. 30 minutes), we aimed to understand how these users would react to some initial design provocations. Based on our literature review, casual graphs, as used in \citep{Yan2020SILVA}, seemed an intuitive and well-received yet much less common way to explore fairness that could be useful for both data scientists and loan officers. We wanted to see the reactions of our participants to this. Hence, we introduced the concept of a causal graph to support fairness investigations. We only provided a simplified example (Figure~\ref{fig:workshop causal graph example}) to gauge their initial reactions. 

\ \ \ \ \textbf{[Figure 2 near here]}

After the workshop, all data was transcribed and de-identified. Thematic coding \citep{clarke2015thematic} identified themes around:
\begin{itemize}
    \item Fairness in AI:	Opinions on the limitations and benefits of AI and human involvement in loan application decision making and fairness.
    \item Fairness Criteria:	The criterion that participants use to judge and decide fairness within the dataset.
    \item Fairness Assessments:	How participants make fairness assessments within the dataset. This includes what they look at, how they manipulate data to explore and troubleshoot, and the decision points they encounter along their journey.
    \item Causal Graph:	Comments on the limitations, benefits and improvements of the causal graph presented during the workshop.
\end{itemize}

The insights from workshop 1 guided the design of an initial prototype which was discussed in Workshop 2. 

\subsubsection{Workshop 2}
The second workshop aimed to gain early feedback on an initial prototype that was developed based on insights from workshop 1, to deepen our understanding of the informational and functional requirements needed to support stakeholders to explore fairness, and to extend and change our designs based on requirements that we had missed in workshop 1 or needs that we had misunderstood. These workshops took place with the same groups and participants in November 2020, two months after Workshop 1, lasting 2 hours each.  

\begin{figure}
     \centering
     \includegraphics[width=.8\textwidth]{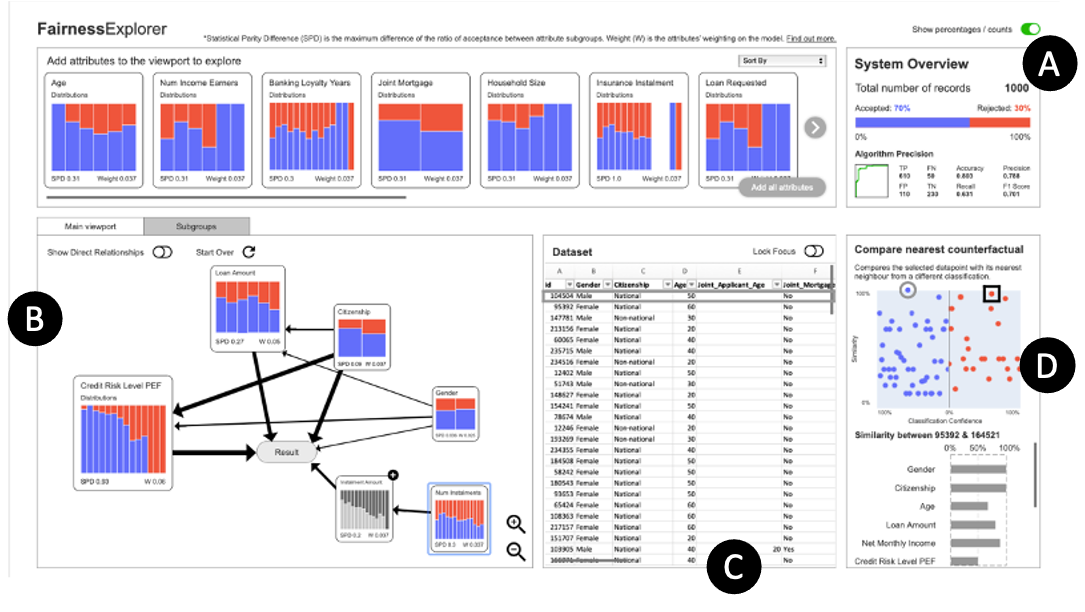}
     \caption{Initial prototype.}
     \label{fig:initial_proto}
 \end{figure}

The initial prototype contained several different UI components (Figure~\ref{fig:initial_proto}): a system overview (A),  a causal graph showing the relationships between five features (B), a dataset view (C), comparing nearest application with a different outcome (D). There was also an opportunity to see information on subgroups. Components within the UI worked in conjunction with one another, for example, selecting an application in component C highlighted this application in component D.

\ \ \ \ \textbf{[Figure 3 near here]}

We took participants through the UI to gather feedback. The prototype was screen-shared; participants did not have access to it directly, instead the researcher clicked through it and described what would happen in a fully developed prototype. 

As in workshop 1, data was collected through screen and audio recordings, transcribed and analysed through thematic coding~\citep{clarke2015thematic}. Themes relating to participants' reactions were developed around the screens and interface areas, and the clarity and usefulness of information and functionality offered in the prototype UI.

\subsection{Insights into Stakeholder Requirements} 

\subsubsection{Human-in-the-loop Fairness Process}
\label{section:fairness_process}
There were some differences between data scientists and loan officers in perceptions about the role that ML should play and how human-in-the-loop ML fairness assessments need to be situated in the decision-making process. Both stakeholder groups stated that ML systems do not currently provide the ability to explain their decisions. Participants were aware of the impact biased ML systems could have on individuals and that human intervention is needed to ensure that outputs were fair. However, they also stated that it is difficult for users to control the ML system's decision-making.

Loan officers saw ML systems as additional support for human decision-making. Rather than replacing human expertise, they thought that a ML system does not see the whole picture of the decision-making. Loan officer thought that the information that the ML system is based on might sometimes be subjective, or that important information is not available to the system at all. For example: \begin{quote}
    “I spent two years in a very small branch in very little town, the knowledge of the customer is something that is very valuable and maybe as you as an employee could provide something extra - soft information - that would help the system to get the right decision in the end.” - WL01
\end{quote}

In their view, this inability of the ML system to work with tacit information and take into account `grey areas' may then lead to unfairness. This reflects earlier findings \citep{Lee2017AHuman-Centered} that pointed out that some tasks are seen to require human skill.

Data scientists on the other hand leaned towards a process where the ML model was checked by humans before adoption.  

\textbf{What this means for user interfaces:} The context in which human-in-the-loop fairness tools are used will matter for their design.  Because data scientists would use these tools for model fairness assurance, we expect that they focus more on fairness measures of the model and relationships across attributes. Loan officers, on the other hand, want support for fair decision-making where humans have high control~\cite{shneiderman_human-centered_2020}, and hence they would focus more on information and functions that apply to individual applications. We do not suggest that these specific foci are exclusive, i.e. information about model fairness might also be useful to loan officers. What we take away from this though is that the UI should be adaptable to the needs of these user groups, or be equally usable by both. 

\subsubsection{Fairness Criteria and Metrics}
\label{section:criteria}
Our investigation found that loan officers and data scientists use different fairness criteria.

Loan officers agreed that group fairness was important: 
\begin{quote}
    “I think there is a problem with adverse selection, often related to age, for example, younger people are disadvantaged, or nationality - foreign people are also penalised and another issue that I saw in the past is the absence of a credit history.” - WL04
\end{quote}

However, they focused mainly on individual fairness, often wrestling with unintended bias creeping in. For example: 
\begin{quote}
    ``I'd like to have the same treatment if I asked for money for a car and for healthcare. Why should I receive a different result if I can afford something?'' - WL02
\end{quote} 

Perhaps because of the danger of bias based on the `worthiness' of loan purpose, loan officers focused most frequently on \emph{affordability}, by trying to understand how income and out-goings contributed to the application decisions. For example:
\begin{quote}
    ``We would have to first define what we intend for fairness, so fairness from an economic point of view from the sustainability of the numbers, or fairness from the point of view that this person is good enough to have the loan, maybe despite some short fall in the economic data, so I think this is important to set what we are looking at.'' – WL01
\end{quote}

In contrast, for data scientists group fairness played a significant role, while they did not mention individual fairness frequently. In fact, it seemed that they shied away from using individual fairness, possibly because it is more subjective, for example: 
\begin{quote}
    “How do we assess that two people are similar?” - WD06  
\end{quote}

Data scientists preferred to use metrics that captured group fairness objectively. There was also an expectation that a range of different metrics should be available in the UI, to cover different ways of assessing fairness.
\begin{quote}
    “Okay, yeah, one thing is, if this should be a fairness explorer, there must be a lot of other metrics of fairness.” – WD02
\end{quote}

\textbf{What this means for user interfaces:} Designs should support a range of popular fairness metrics, and support adding new ones if they are not already included. As previously discussed, there are quite a number of ways to measure fairness and it is sometimes unclear what measures to apply, therefore, it might also be necessary to describe these metrics and how they are derived in more detail. 

While there are readily available metrics for capturing fairness that can be applied in AI systems and which would be useful for data scientists, there is currently a lack of ways to support the fairness assessments of loan officers. This could be overcome by allowing loan officers to specify `custom' attributes, for example, to capture the notion of affordability. These attributes can then be used in ensuring fairer decision-making. 

\subsubsection{Exploring Features and Relationships}
\label{section:relationships}
To make fairness assessments, both groups looked in detail at what they considered sensitive features. Gender and citizenship were discussed the most in both groups and both considered them unfair to be used in AI decision-making. However, age as a sensitive feature proved to be more contentious. Loan Officers in particular pointed out that young people often could not provide a long credit history, so suffered from a `cold start' problem in loan applications. Similarly, older people might be deemed too much of a risk if their repayment terms were long. 

Both groups requested information on the acceptance rates for loan applications related to feature values, particularly for the ones they considered sensitive. In general, investigating relationships between attributes was very important to them, through exploring the distribution of attribute values to the target values. This allowed them to compare whether there was a bias against certain applicant groups, and to determine if there was a cause for concern in terms of group fairness: 
\begin{quote}
    “Perfect, so it's not so small the difference between 70/30 and 40/60, about 10\% higher the rate of rejected proposal of non-national customers.” – WL05
\end{quote}
 
Data scientist also wanted to explore subgroup fairness, in addition to simple group fairness. This meant combining a number of feature values and looking at the acceptance rates. 

As already highlighted, participants identified that outcomes might be related to features and feature combinations, however the stakeholders were also interested in investigating relationships between any features to see if they were linked to each other, to see how bias might have crept in unintentionally:
\begin{quote}
    “The first thing that I did was to check the correlations, mainly between the target variable, which is the result\ldots the most correlated variable is of course, the risk levels. But for example, I didn't find any correlation among the gender or the age or something like that\ldots [it would be] probably nice to deeply analyse [the dataset] better and check also the correlations between the other variables to understand if there is something biased as we said at the beginning of the discussion… I also did some plots to check for example, the percentage of accepted or rejected for each variable.” – WD01
\end{quote}

Loan officers often leveraged their domain knowledge to explore fairness. For example, they were already aware how credit risk could impact the outcome, and brought in the citizenship feature in their exploration: 
\begin{quote}
    ``Yes, this is pretty similar, yes, it's more approval in the national but it's maybe usual because national people usually earn more money than people from abroad. Nothing unusual.'' - WL03
\end{quote}

\begin{figure}
    \centering
    \includegraphics[width=.5\textwidth]{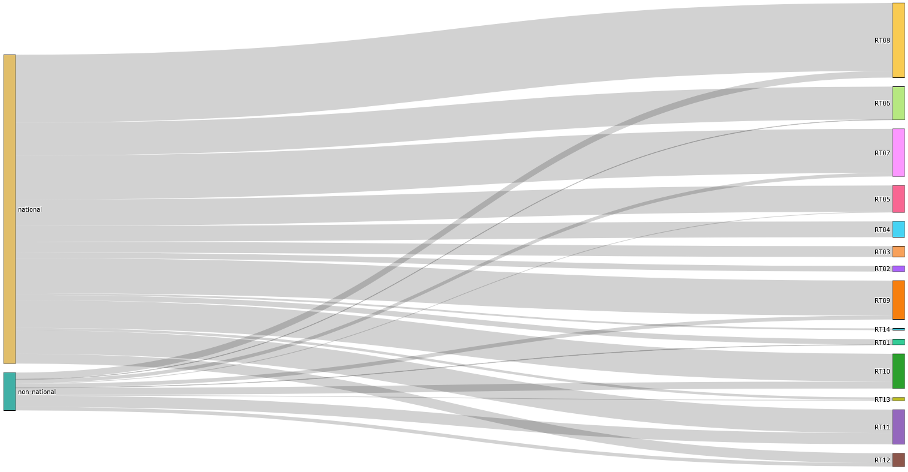}
    \caption{Sankey graph showing relationships between nationality and credit level.}
    \label{fig:sankey}
\end{figure}

In our workshops, we provided Sankey diagrams (Figure \ref{fig:sankey}) as design provocations for communicating relationships between features values. While both stakeholder groups stated that visualizing relationships between features is useful, it became clear that this choice of visualization was too complex for participants to grasp, especially for continuous, binned feature values. Instead, they requested simplified visualizations or numerical information. 

\ \ \ \ \textbf{[Figure 4 near here]}

We also showed causal relationships underlying the data, showing these in a simple node-edge graph (Figure~\ref{fig:workshop causal graph example}), to allow an alternative and easy way for stakeholder to investigate relationships between all features. Both stakeholder groups saw the value in having causal relationships shown:
\begin{quote}
    ``I think the crucial point is at the beginning to decide what is fair and what is not in that specific domain. And in this respect, maybe the graph that you're showing could help. Let's take for example, gender. This graph is telling us that gender impacts the decision, at least through the credit risk level, then you have to ask about this impact; so, the impact that goes from gender through credit risk level to result is something that I'm willing to accept as a correlation from gender to result or not? So, you have to ask yourself, how is credit risk levels calculated?'' – WD04
    \end{quote}

They also appreciated how these graphs could be integrated in the process of decision-making in loan applications and mitigating fairness issues:
\begin{quote}
``We cannot modify the graph? So for example, if we don't want gender, we can remove it?'' - WL06 
\end{quote}

\textbf{What this means for user interfaces:} Although protected features by law will need to be taken into account, there is considerable variability between the laws of countries in terms of which features count as sensitive. Further, other features might also be considered sensitive even though they are not protected by law. Hence, these features need to be selectable by the user, and then highlighted as especially important in the UI. 

Outcomes related to features values and their ratios and distributions formed a core part of the assessments of the stakeholders. This information will need to be made available in an easily understandable format, perhaps through appropriate visualizations that can support comparisons between applicant groups, or ratio percentages. 

Exploring relationships between features should also be extended to features that are not considered sensitive and features that do not directly influence the outcome. In addition, information about combination of features needs to be provided to investigate issues of subgroup fairness. This also allows intersectional issues of fairness to be explored, and would allow stakeholders to find biases that are perhaps less obvious.

Exposing causal relationships can add information that would otherwise not be available when looking at a dataset. For loan applications this is important, as the features used in calculations are often interrelated. 

\subsubsection{Exploring Individual Cases and Their Similarities}
\label{section: similarity} While a lot of the exploration of the data focused on investigating group fairness, we showed earlier (\ref{section:criteria}) that loan officers also considered individual fairness. In the workshops we also investigated how to support this. 

Feedback indicated that both stakeholder groups liked the ability to see and compare individual applications:
\begin{quote}
    “For me it's cool. Maybe the coolest part” – WD05.
\end{quote}

However, the visualizations we presented proved to be problematic as participants did not understand how similarity was calculated and how the data was mapped to the axes. What they wanted to compare was the similarity of applications, especially if they led to different outcomes, and what attributes caused this.

\textbf{What this means for user interfaces:} Comparing similar applications was appreciated and should be supported in fairness exploration, in order to investigate individual fairness. These should provide information about similarity of cases, including their attributes, across decision boundaries. 

\subsubsection{Transparency and Explanations}
\label{section:transparency}
Both groups wanted more transparency about how metrics or other information was derived or calculated. For example, they wanted information on how the causal graph was created, and how the ``similarity'' score was calculated.

In particular, data scientists wanted a clear distinction between what belongs to the ML model and what belongs to the dataset from which the ML model is derived:
\begin{quote}
    “The other thing that I noticed is I see a system overview accepted and rejected. I don't know if it's the model or the original, maybe I guess their model, but I'm not sure.” – WD06
\end{quote}

It was obvious that loan officers struggled with ML concepts that were very familiar to data scientists. For example, they had difficulty with concepts such as target, precision, etc. and also how ML algorithms worked.  

\textbf{What this means for user interfaces:} Much effort has been directed at providing explanations of ML systems and that transparency is required \citep{kulesza_principles_2015, lim_assessing_2009}. Our findings echo this, showing that users require detailed descriptions so they can understand how information and visualizations have been generated, in order to interpret them effectively. Contextual help could aid in this. In addition, explanations need to be carefully targeted at different stakeholders to extend users' understanding, and mental models \citep{Kulesza2012TellMeMore}.

\subsubsection{Summary}
Table~\ref{tab:RequirementsSummary} shows a list of requirements that we gathered through the workshops. Participants in our study supplied many informational and functional requirements that need to be supported to explore fairness. These requirements provide a much richer design space than the functionality provided by existing tools described in section 2.2.6. We found that data scientists and loan officers have slightly different priorities but overall their needs are surprisingly similar. Based on these requirements, we started to develop a prototype UI that would support data scientists and loan officers needs. We describe this UI next. 

\ \ \ \ \textbf{[Table 3 near here]}

\section{FairHIL, a UI to Support Investigating Fairness}

Our design space exploration that started with the literature review and design workshops continued with the implementation of a UI of 7 components. We now present a detailed description of these UI components we developed; for each component we reference the corresponding requirements we captured from the design workshops (see Table \ref{tab:RequirementsSummary}. Please note that we focused on informational and functional requirements that allow stakeholders to investigate fairness. Requirements that focus on mitigating fairness by adjusting the model were considered out of scope for the purposes of our research. For each of  these UI components we also show how it could be generalized for \emph{any} dataset and model. 

The UI is divided into a setup process and the main UI screen to explore the input dataset and the model, respectively. The setup comprises five steps for data scientists and four for loan officers, inspired by a wizard approach which is frequently implemented in UI setup processes (e.g.  Fairlearn~\citep{bird2020fairlearn}) and involves loading the dataset, setting the target feature, selecting the prediction model, marking up sensitive features, and choosing fairness metrics, including the ability to define a 'custom metric' which leads to a function builder. (In order to simplify the interface, we did not include a choice of prediction model or metrics for loan officers.) We will focus on the \emph{metrics choice} UI component in section~\ref{section:metrics}. 

\ \ \ \ \textbf{[Figure 5 near here]}

\begin{figure}
    \centering
    \includegraphics[width=.7\textwidth]{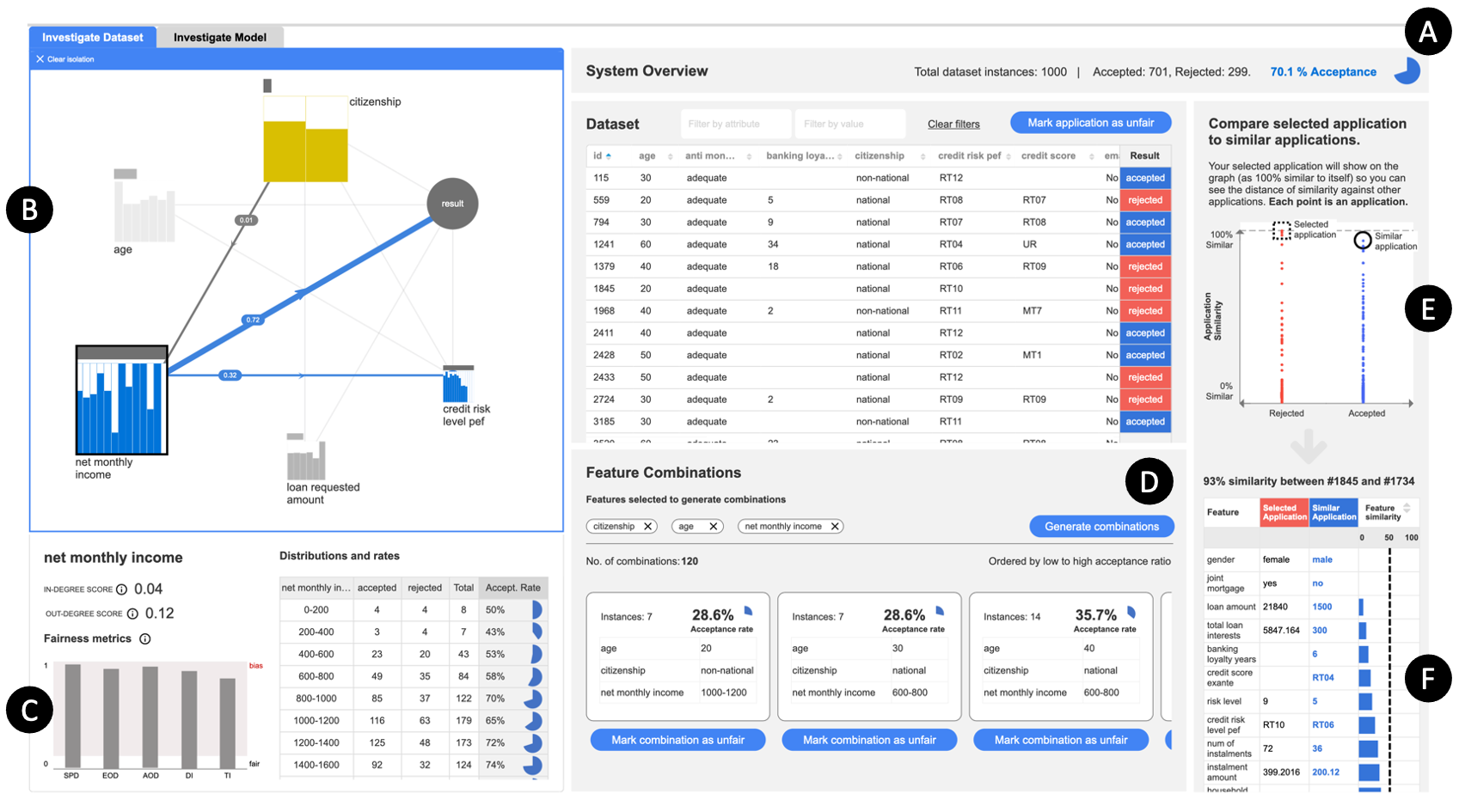}
    \caption{Main UI Components.}
    \label{fig:main_UI_overview}
\end{figure}

The main UI screen (Figure~\ref{fig:main_UI_overview}) allows users to rapidly switch between information about the dataset and the model through tabbing, and thus it supports them to easily compare the fairness of underlying loan decisions as well as the resulting predictions. The UIs to explore the dataset and the model are very similar, each with six main components that relate to each other. The \emph{system overview} (Figure~\ref{fig:main_UI_overview} A) provides overall information about the data and target, described in section~\ref{section:sys_overview}. The main component supporting users to explore fairness is a \emph{causal graph} (Figure~\ref{fig:main_UI_overview} B) which shows features and causal relationships between them and the target. We discuss this UI component in detail in section~\ref{section:causal_graph}. Detailed \emph{information about features and relationships}, such as fairness metrics, distributions, and acceptance ratios in the causal graph, is displayed in a component underneath the graph (Figure~\ref{fig:main_UI_overview} C). This UI component is described in section~\ref{section:info}. Users are also able to see how intersections of feature values affected the acceptance ratios by exploring \emph{feature combinations} (Figure~\ref{fig:main_UI_overview} D), as we will describe in section~\ref{section:combinations}. We also provide access to the underlying \emph{dataset} (Figure~\ref{fig:main_UI_overview} E) which allows filtering and more detailed explorations, see section~\ref{section:dataset}. Finally, users are able to select an application in the dataset component and compare it with other applications in the \emph{compare applications} component (Figure~\ref{fig:main_UI_overview} F), which we will detail in section~\ref{section:compare}.

Throughout we focus on providing transparency and explanations in UI components, to satisfy user requirements uncovered in section~\ref{section:transparency}. We add detailed explanations, particularly targeting loan officers who struggled in the workshops with ML terminology and concepts. We concentrate on explaining the target and the algorithm, sensitive features, and the fairness metrics, and their calculation. Furthermore, we explain how the causal graph was developed and the information it showed in detail. We also provide contextual information on individual UI elements, for example when fairness metrics are shown in the UI components.

\subsection{Metrics Choice}
\label{section:metrics}
\begin{figure}[h]
    \centering
    \includegraphics[width=.7\textwidth]{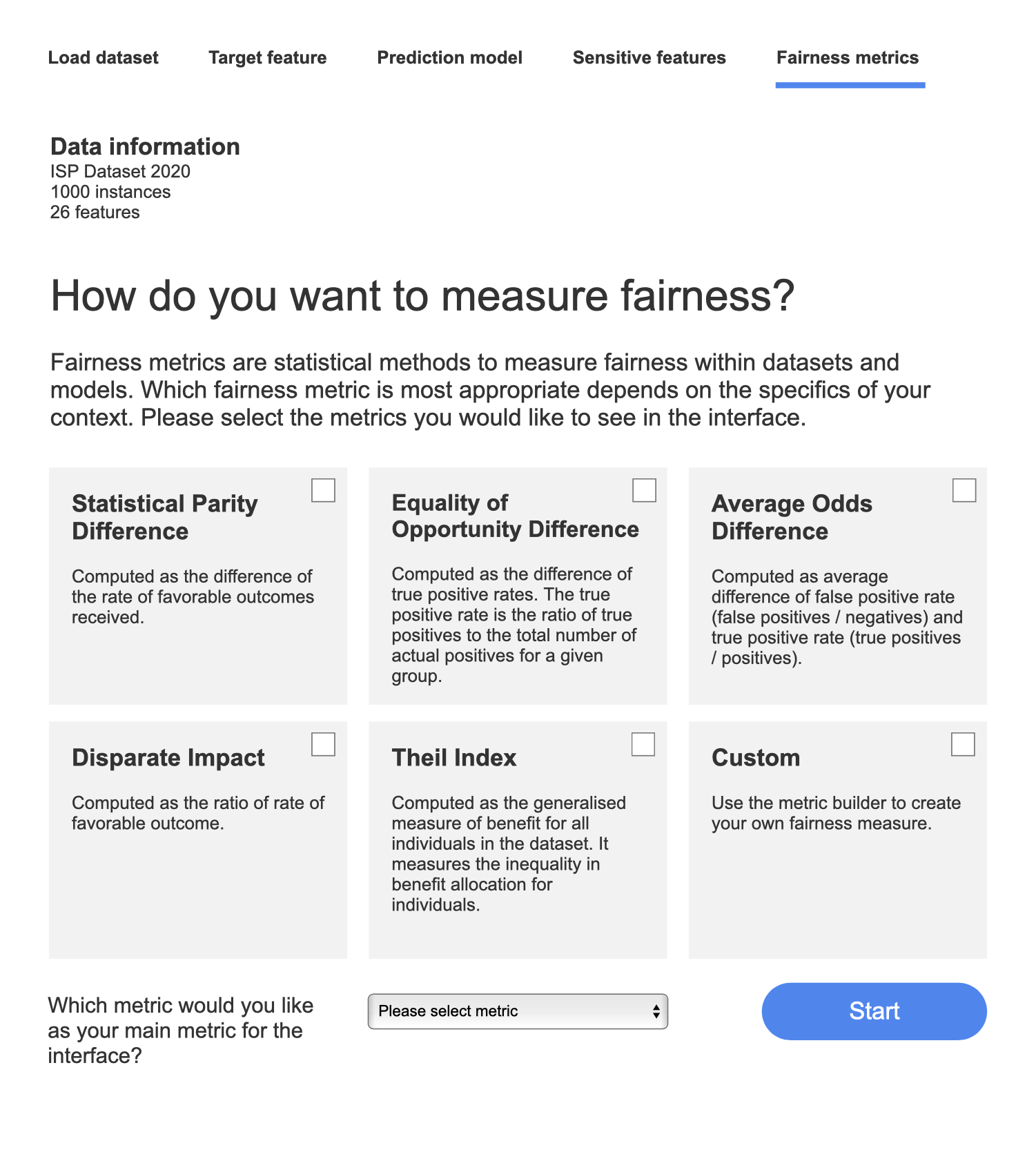}
    \caption{Setup: metrics.}
    \label{fig:setup_metrics}
\end{figure}

A main way of exploring fairness is through the use of fairness metrics (REQ1.3, REQ2.8). Providing users with a range of popular metrics is important as fairness criteria are both subjective and contextual to the problem at hand. To support this, data scientists are able to choose one or more of five metrics: statistical parity difference (SPD), equality of opportunity difference, average odds difference, disparate impact, and Theil index (see Figure~\ref{fig:setup_metrics}).  These metrics are commonly provided in other tools~\citep{aif360-oct-2018}. For loan officers, we provide only one metric, SPD, because it is relatively simple and most widely used in different contexts. To extend and generalize this UI component, further metrics could be added to the selection, e.g. drawn from~\citep{mitchell2018prediction} or \citep{verma2018fairness}. 

\ \ \ \ \textbf{[Figure 6 near here]}

\begin{figure}[t]
    \centering
    \includegraphics[width=0.7\textwidth]{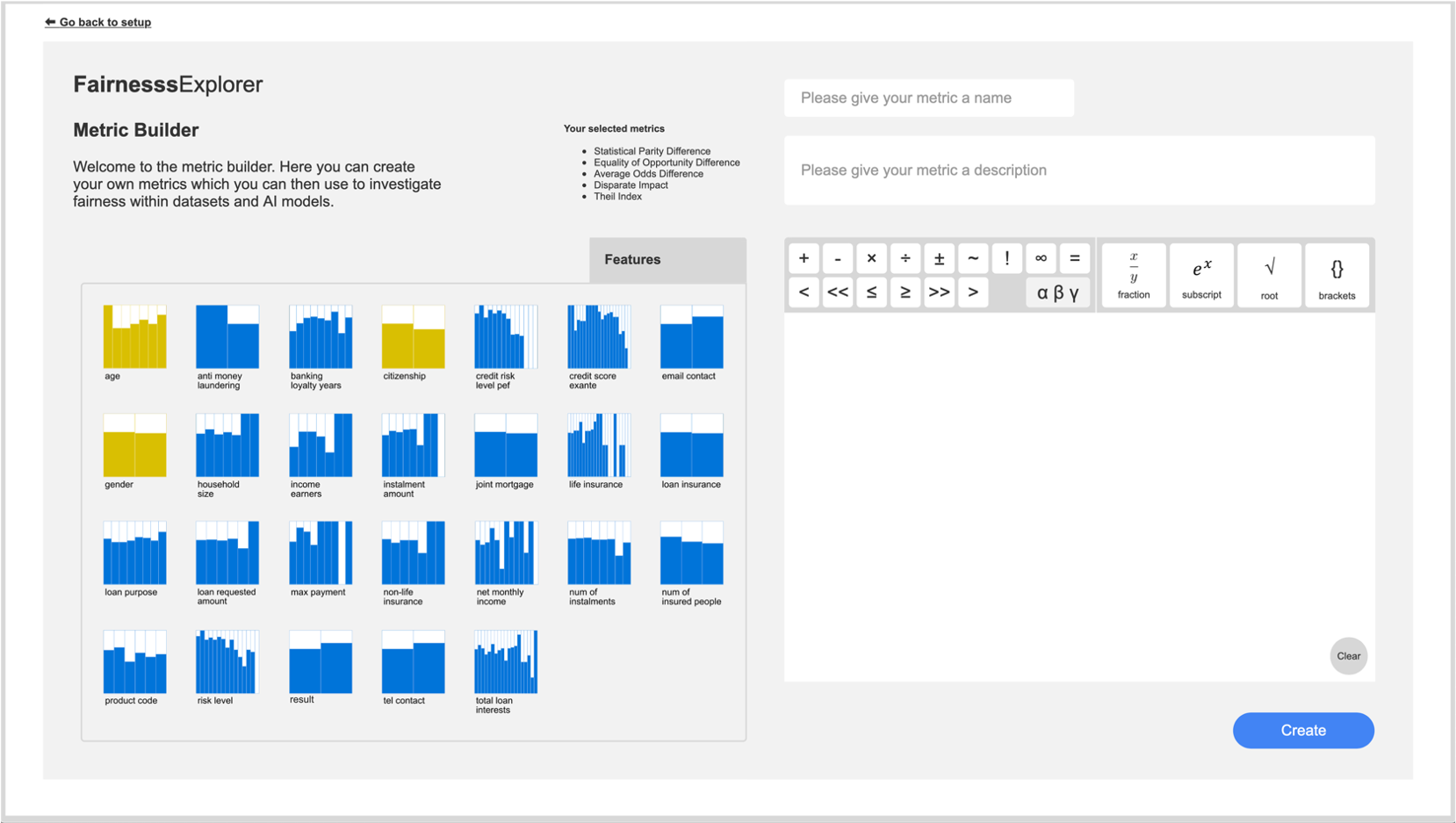}
    \caption{Metric builder.}
    \label{fig:metric_builder}
\end{figure}

We also provide the ability to specify a custom attribute which could be used as an informal 'metric' (REQ2.7) , in order to support alternative user-generated or domain specific notions of fairness, as seen in section~\ref{section:criteria}. For example, during the workshop, stakeholders repeatedly raised notions of affordability which could be added in addition to the predefined fairness metrics. The custom metric builder (Figure~\ref{fig:metric_builder}) allows users to give the metric a name, to select from existing features along with their value distributions against the target, and a work area where they are able to construct a metric using features and mathematical operators, inspired by other formula builders. Further extensions could obviously be made to the formula builder, for instance to show example output, or to incorporate more complex functions than the basic mathematical operators.

\ \ \ \ \textbf{[Figure 7 near here]}

\subsection{System Overview}
\label{section:sys_overview}
This component provides important contextual information around fairness of the data or the model to the user by showing the number of instances in the dataset (REQ1.1), an overall target ratio, shown as a percentage and a visual pie slice, which in our use case was the acceptance rate (REQ1.4). For the model, we show the same information calculated on the test data. 

The system overview could of course be extended with other global information. For example, for the model, we could add a positive prediction rate, alongside other accuracy metrics. Other forms of representing and visualizing this data could also be explored.

\subsection{Causal Graph}
\label{section:causal_graph}
\begin{figure}[t]
    \centering
    \includegraphics[width=.6\textwidth]{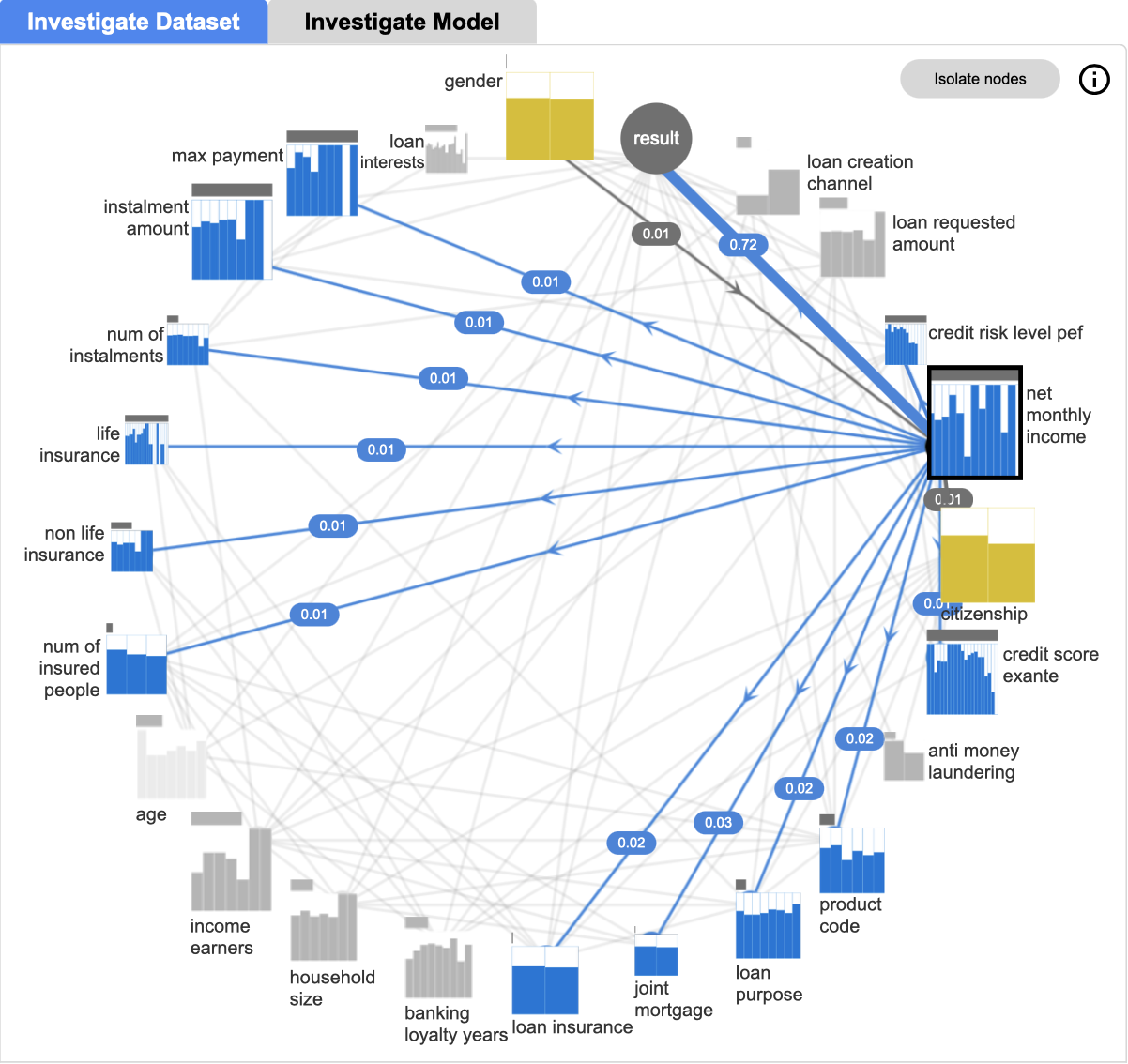}
    \caption{Causal graph in the dataset view showing a selected node: "net monthly income"}
    \label{fig:causal graph}
\end{figure}

The main way for a user to explore features and relationships between features, a key need that was highlighted in the workshops, is through a causal graph (Figure~\ref{fig:causal graph}) (REQ2, REQ3). The causal graph is made up of nodes representing features and edges representing causal relationships between features, presented in a circular layout. Causal graphs can be inferred by applying causal discovery algorithms~\citep{zheng2018dags} and then followed up and validated by domain experts. Sensitive features marked up during the setup process are highlighted in gold (REQ1.5); relationship strengths are indicated by line thickness (REQ3.2); out-degree is represented by node size. The target is shown as a circle node to differentiate it from the other features.

\ \ \ \ \textbf{[Figure 8 near here]}

In the dataset view, each feature node contains a bar graph which shows the target distribution, i.e. acceptance rate, for each feature value (REQ2.2). In the case of categorical data this is fairly straightforward, while binning is used for continuous features. We use a simple and commonly used square root approach for the binning process but other ways of determining bin size, such as Sturges approach, could be explored. Above this information, we show a small horizontal bar: its length indicates the relative value of the statistical parity difference metric for this feature, mapped to a 0 to 1 bar chart. This allows users to quickly identify features for further exploration, either through an imbalance in target distributions for feature values, or a high fairness metric value, as discussed in section~\ref{section:relationships}. We choose percentage bar graphs as they normalize the data and allow easy comparison to enable the identification of trends or possible relationships.

Using a contextual menu, users are able to add a feature to a combination (see section~\ref{section:combinations}) (REQ2.9), toggle the feature's status as sensitive (REQ1.5), or mark the feature as potentially “unfair” for further discussion (REQ1.7, REQ2.16), which set the feature color to red. 

To ease visual processing load, clicking or hovering over a feature makes all connected relationships more prominent. Outgoing edges are colored blue to encode the direction of causality (REQ3.2). In addition, if a relationship is clicked or hovered over, the relationship's strength is shown in numerical form, generated from the causal discovery algorithm. 

Initially, all features are displayed in the causal graph to get an overall sense of the data, based on feedback from the workshops (REQ1.1, REQ1.3). Users are also allowed to temporarily filter features and their relationships that interest them most for further exploration, reducing the visual `clutter' of the graph. To do so, users are able to enter a drill-down mode and select the features of interest (REQ2.1, REQ2.3). Only the selected features and their relationships are then shown in this UI component. The user is able to easily exit this drill-down view and return to the full causal graph.

\begin{figure}[t]
    \centering
    \includegraphics[width=.5\textwidth]{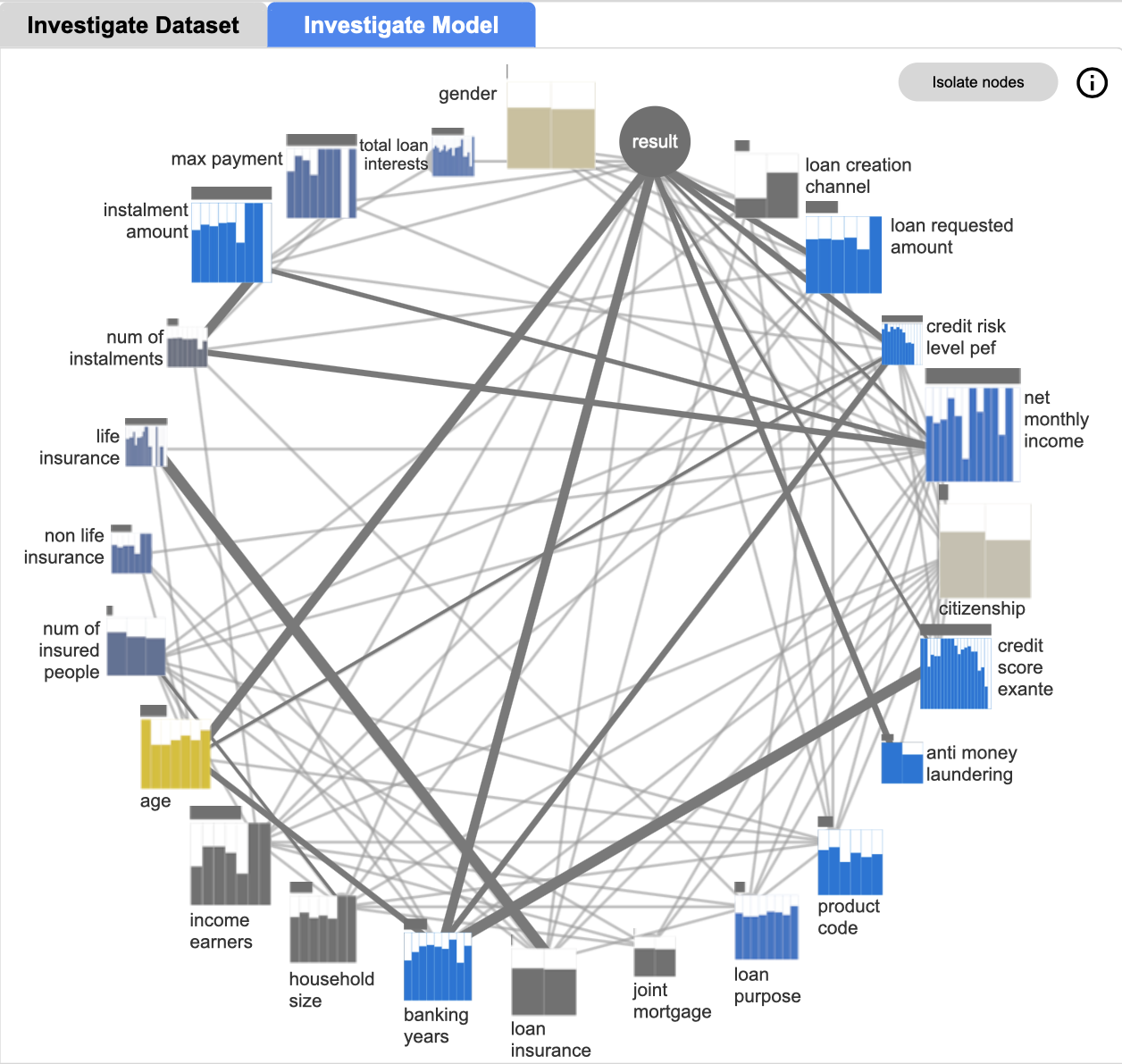}
    \caption{Causal graph in the model view showing feature importance (saturation).}
    \label{fig:feature_saturation}
\end{figure}

When investigating the model, feature importance is encoded as the feature's color saturation (Figure~\ref{fig:feature_saturation}) (REQ3.1). Feature importance in the model represents the absolute value of the feature weight and therefore can be mapped to a saturation scale between 0 to 1. The less saturated a node, the smaller of an impact it has on the model, giving users an indication of how much features contribute to the model's predictions. 

\ \ \ \ \textbf{[Figure 9 near here]}

The causal graph offers opportunities for extension based on domain and information requirements. Obviously, different choices could have been made for encoding information in the causal graph, by choosing different mappings of information to visual attributes such as colors and size, by selecting other graph types, or by highlighting other important information such as different feature types. However, we encourage designers to favor simplicity and ease of understanding in order to communicate this information to data scientists but particularly to domain experts. We also implemented a common `overview and drill-down' data visualisation strategy which was suited to our stakeholders.

\subsection{Features and Relationships Information}
\label{section:info}
The causal graph works in conjunction with a supporting information component. This displays additional contextual information triggered when selecting features and relationships (REQ1.1, REQ 1.3, REQ2.2), satisfying the main requirements outlined in section~\ref{section:relationships}. 

\begin{figure}
    \centering
    \includegraphics[width=.7\textwidth]{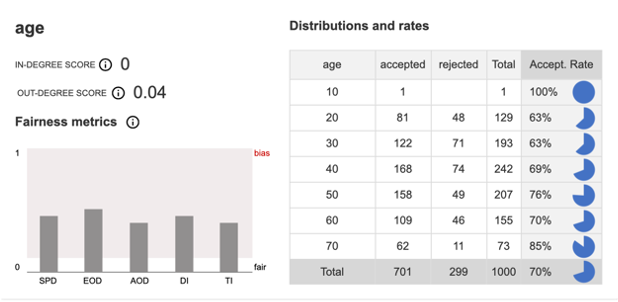}
    \caption{Detailed Information on Features.}
    \label{fig:feature_info}
\end{figure}

In the dataset view when clicking on a feature (Figure~\ref{fig:feature_info}), the component shows in-degree and out-degree values derived from the causal graph, a bar graph of fairness metric scores, and a table which shows, for each categorical or binned feature value, the number of instances in each target category -- here `accepted' and `rejected' -- and the acceptance rate in the same way as in the system overview.  In the model view, we add prediction confidence, i.e. the certainty for making this prediction represented by its closeness to the decision boundary \citep{kulesza_principles_2015}, as an additional column to this table, and display this as pie slices. This table provides the same information as the feature bar graph in the causal graph and also allows easy, visual comparison between feature values and against the overall target ratio.

\ \ \ \ \textbf{[Figure 10 near here]}

\begin{figure}
    \centering
    \includegraphics[width=.7\textwidth]{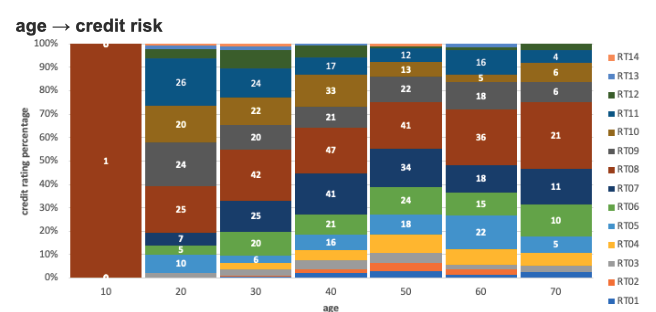}
    \caption{Detailed Information on Relationships.}
    \label{fig:rel_info}
\end{figure}

When clicking on a relationship in the causal graph, the component presents a stacked bar graph (Figure~\ref{fig:rel_info}), showing the percentage of the intersection of the two related feature values (REQ2.1). The outgoing feature's value (the cause) is mapped to the x-axis of the graph, while the other feature's values are shown on the y-axis. This simplified the Sankey diagrams originally shown in the workshops (see section~\ref{section:relationships}) while supporting easier comparisons. 

\ \ \ \ \textbf{[Figure 11 near here]}

This UI component could be extended to suit other use cases. The bar graph of fairness metrics is extensible to show more metrics, if desired. The table of detailed feature information could also include other information that is of importance to the user group. As always, information could be encoded differently. In particular, while we chose to show relationships between two features in a very simplistic way, in preference to Sankey diagrams, there might be other way of better visualizing relationships for specific user group.

The bar graphs within this component are also connected to the dataset component (see section~\ref{section:dataset}); clicking on a particular value or bin filters applications shown in the dataset component. This allows users to further explore applications that might be problematic (REQ4).  

\subsection{Feature Combinations}
\label{section:combinations}
\begin{figure}[t]
    \centering
    \includegraphics[width=.6\textwidth]{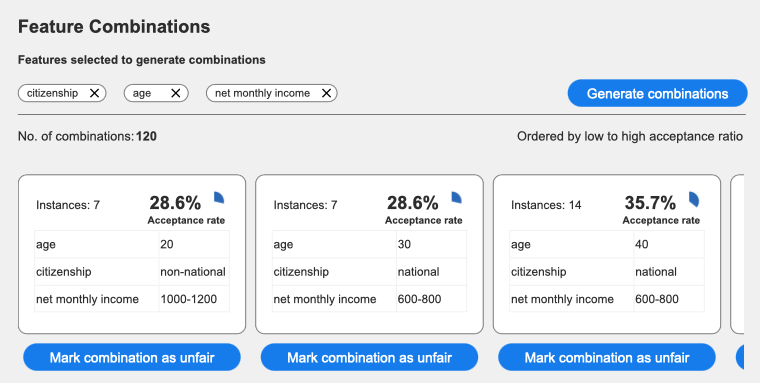}
    \caption{Feature Combinations Component.}
    \label{fig:feature combinations}
\end{figure}

To support the investigation of intersectional bias and subgroup fairness, as evidenced in the workshops (see section~\ref{section:relationships}), users are able to add features to a combination (REQ2.9). Subgroups (Figure~\ref{fig:feature combinations}) are presented as `cards' which communicate the subgroup's acceptance rate, number of application instances in this subgroup and the combination's feature values. To focus attention on possible issues of fairness, cards were ordered by low to high acceptance ratio. Users were able to highlight combinations as `unfair' for further exploration and discussion.

\ \ \ \ \textbf{[Figure 12 near here]}

In the implemented prototype, users have to select subgroups of interest. In a real working system, this could be extended through automatic calculation of all possible subgroup combinations. This might be computationally expensive, however. On the other hand, more support for focused user-driven exploration could also be provided, such as filtering against a minimum number of instances that makes up a subgroup.  

\subsection{Dataset}
\label{section:dataset}
\begin{figure}
    \centering
    \includegraphics[width=.7\textwidth]{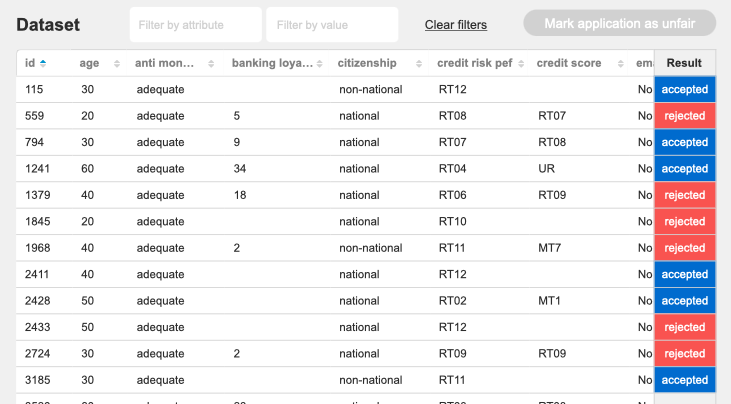}
    \caption{Dataset Component in Dataset View.}
    \label{fig:Dataset}
\end{figure}

This component enables users to see the underlying raw data and investigate individual applications (REQ4). The dataset component (Figure~\ref{fig:Dataset}) presents all features in a spreadsheet-like table, familiar to many users (REQ4.1). It allows users to filter and sort by feature and value, and is also connected to the feature information component described in section~\ref{section:info}. The target feature (here, the result column) and its values are always visible, with additional color-coding to help with rapid visual processing. 

\ \ \ \ \textbf{[Figure 13 near here]}

\begin{figure}
    \centering
    \includegraphics[width=.7\textwidth]{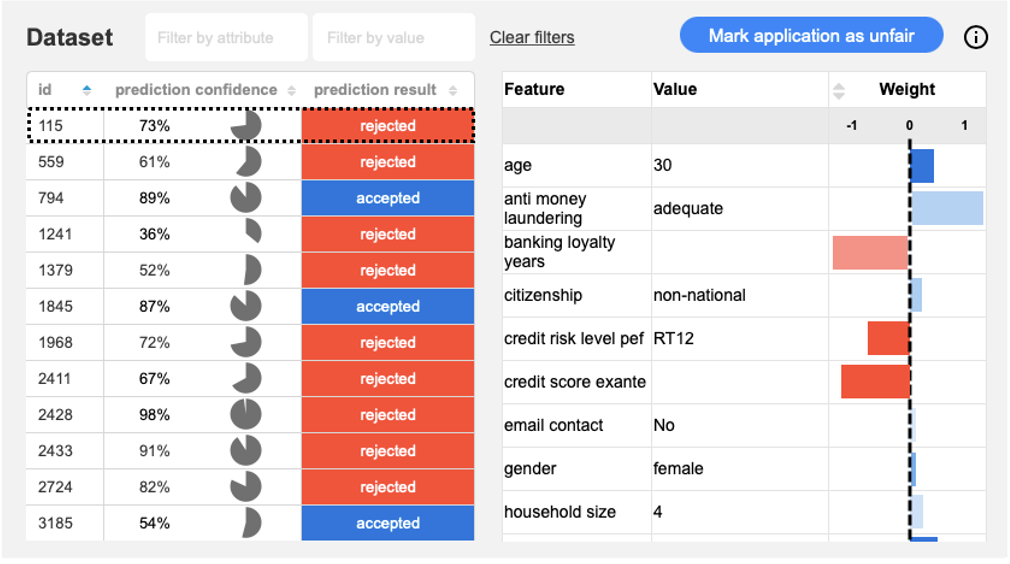}
    \caption{Dataset Component in Model View.}
    \label{fig:Model}
\end{figure}

The dataset table in the model view (Figure~\ref{fig:Model}) also includes a column for prediction confidence. i.e. the application's certainty or closeness to the decision boundary, which is shown as a percentage and a pie slice visualization. In addition, if a user selects an application we show the corresponding feature values and their relative feature weights. The weights are shown as bi-directional bars, either negative (red) or positive (blue). In addition, the criticality of the feature value, i.e. the feature value multiplied by the weight of the feature,  is encoded by color depth; the deeper the color, the more critical the feature value is in determining the target value. This enables users to understand how features contribute to the model.

\ \ \ \ \textbf{[Figure 14 near here]}

Additionally, this component is also connected to the compare applications component (section  \ref{section:compare}). By selecting an individual application, a similarity comparison graph is generated. 

\subsection{Compare Applications}
\label{section:compare}

This component is designed to give users the ability to explore similar applications that might have very different outcomes, hence uncovering potential biases or unfairness at the individual level (see section~\ref{section:relationships}) (REQ4.3, REQ4.4). This is initiated by selecting an application in the dataset component to show a similarity comparison graph. 

\begin{figure}[t]
  \centering
  \subfloat[Dataset View.]{
  \resizebox*{3cm}{!}{\includegraphics{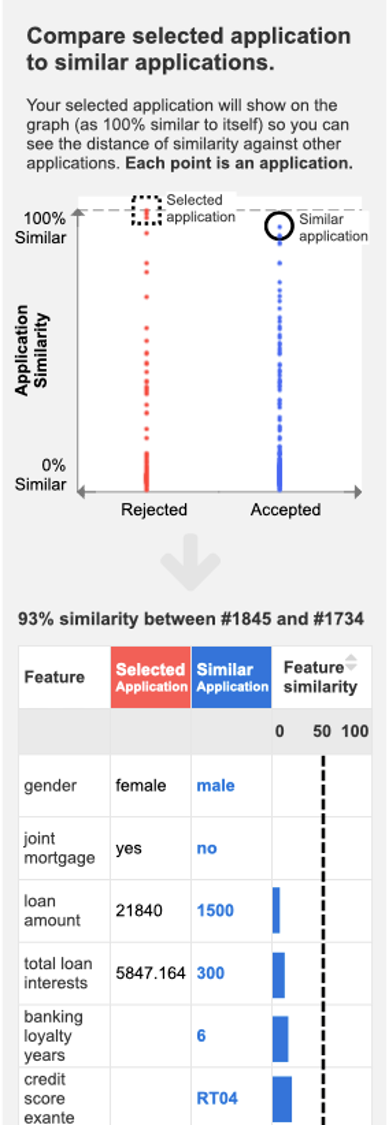}}\label{fig:Dataset_sim}}\hspace{5pt}
  \subfloat[Model View.]{
  \resizebox*{3cm}{!}{\includegraphics{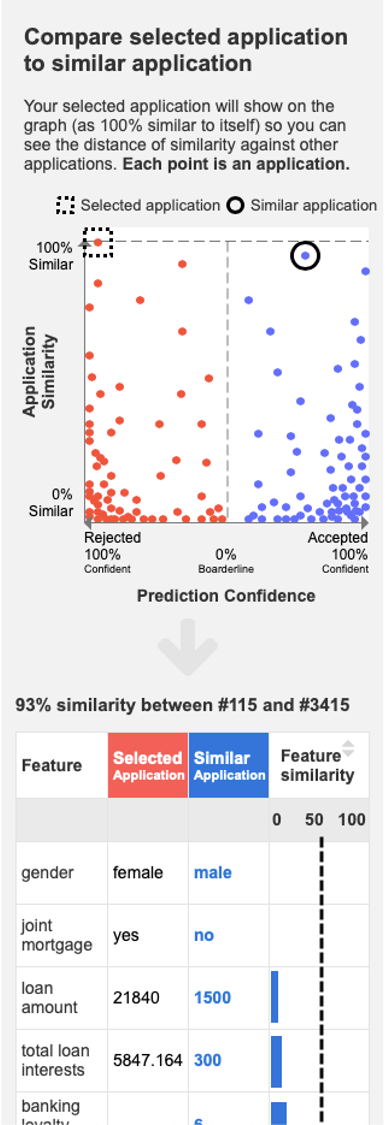}}\label{fig:Model_sim}}
  \caption{Compare Application component.}\label{fig:sim}
\end{figure}

In this graph in the dataset view (Figure~\ref{fig:sim}), we plot all applications as data points. The target categories are encoded with color (accepted: blue, rejected: red) to be consistent with the dataset table's encoding of the target. On the y-axis, the plot shows similarity of all applications to the currently selected application, shown in a square outline (REQ4.3). Different similarity metrics could be applied, we chose the Pearson correlation coefficient. When investigating similarity in the dataset, the target categories are presented on the x-axis. In the model view, the x-axis represents prediction confidence (or certainty) of the target outcome for each application; either to be accepted or rejected. This allows users to understand how close a selected application is to a decision boundary, and also its relationship to all other applications (REQ4.5).

\ \ \ \ \textbf{[Figure 15 near here]}

The user can select another application on the plot, which shows the compared features side by side, along with the feature similarity score represented as a bar chart (REQ4.4). This information is ordered from least similar values to most similar.

Constructing the plot relies to a great extent on the binary target which aids mapping this information on a 2D visualisation. However, if the target had been multi-categorical or continuous, different visualization choices are needed while still encoding similarity and prediction confidence as important information. 

\section{Study 2: FairHIL evaluation with stakeholders}
The aim of this study was to evaluate FairHIL to understand if the UI components we instantiated were useful and usable by loan officers and data scientists, our two stakeholder groups, to investigate fairness. As fairness is such a subjective notion, we did not measure any 'task success', instead we paid attention to qualitative feedback which point the way to improvements of the UI or further requirements.     

\subsection{Participants}
We conducted one-to-one 1-hour sessions with 17 participants, 8 loan officers (mean age=38, stdev=4.54) and 9 data scientists (mean age=31.8, stdev=3.56). Table~\ref{tab:participants_evaluation} shows the background details of the participants in more detail. Due to COVID-19 restrictions all sessions were held online through remote meeting software. Recruitment was facilitated by $<$anonymised$>$, a partner bank involved in the project. This study was approved by $<$anonymised$>$ Ethics Committee. No incentives were offered.

\ \ \ \ \textbf{[Table 4 near here]}

\subsection{Setup}

We focused in detail on the dataset UI components, as they were very similar to the AI model UI components. We probed participants’ reactions to each UI component as they used the prototype: first we gathered opinions about the setup process of loading the dataset and initialising the model, focusing in detail on selection of sensitive features and metrics. We then explored how they would use the prototype to investigate fairness  of a dataset, which consisted of using the causal graph, the feature and relationship information, feature combinations, and investigating the dataset and comparing applications. We repeated this for the model exploration where we focused also on the use of prediction confidence and feature weights.  

As they interacted with the prototype, we encouraged the participants to `think aloud', and verbalize if they would use the UI component, and if so, how. The study concluded with two post-session questionnaires to evaluate users’ experience in investigating fairness using the UI. First, we asked participants to answer the following questions around the \emph{usefulness} of the prototype, using a 7-point Likert scale:
\begin{itemize}
    \item The prototype supports my ability to assess fairness effectively
    \item The interface provides the information I need to assess fairness effectively
    \item The interface provides a sufficient amount of detail needed to assess fairness effectively
    \item The interface provides the functionality I need to assess fairness effectively
    \item The interface supports the way I reason when making a decision
\end{itemize}
We also allowed participants to add free-text comments on their experience of the prototype. Second, we employed the simplified NASA-TLX questionnaire \citep{hart_development_1988}, which is widely used in assessing UIs and XAI to understand workload.  

Sessions were screen and audio recorded and analyzed qualitatively through thematic analysis \citep{clarke2015thematic} focusing on themes of understanding, use and usefulness for each UI component; in addition, we also analyzed the usefulness questionnaire's free comments qualitatively. We report the quantitative results of the questionnaires through simple descriptive statistics for each stakeholder group.   

\subsection{Findings}

\begin{figure}[htb]
    \centering
    \includegraphics[width=.7\textwidth]{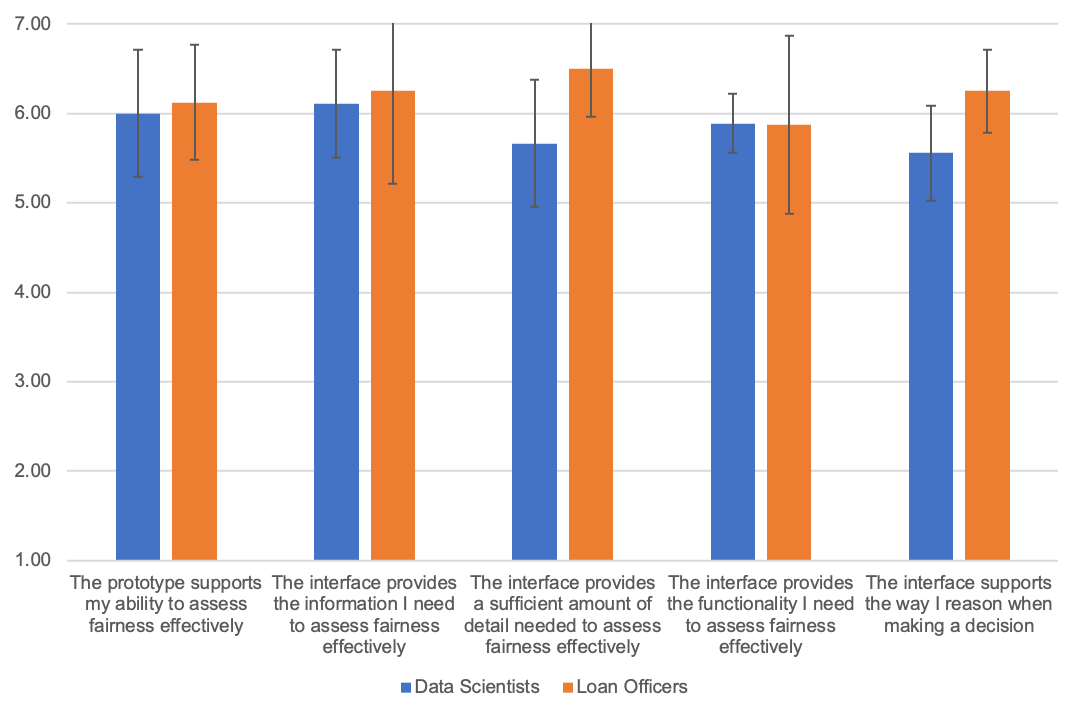}
    \caption{Mean Post-session Usefulness Questionnaire Responses. Error bars show standard deviation.}
    \label{fig:quest}
\end{figure}

Overall, the prototype was received equally well by both sets of stakeholders, as evidenced in the usefulness questionnaire responses (Figure~\ref{fig:quest}).  Loan officers viewed the prototype slightly more positively but we did not find any significant differences in the responses.

\ \ \ \ \textbf{[Figure 16 near here]}

\begin{figure}[htb]
    \centering
    \includegraphics[width=.7\textwidth]{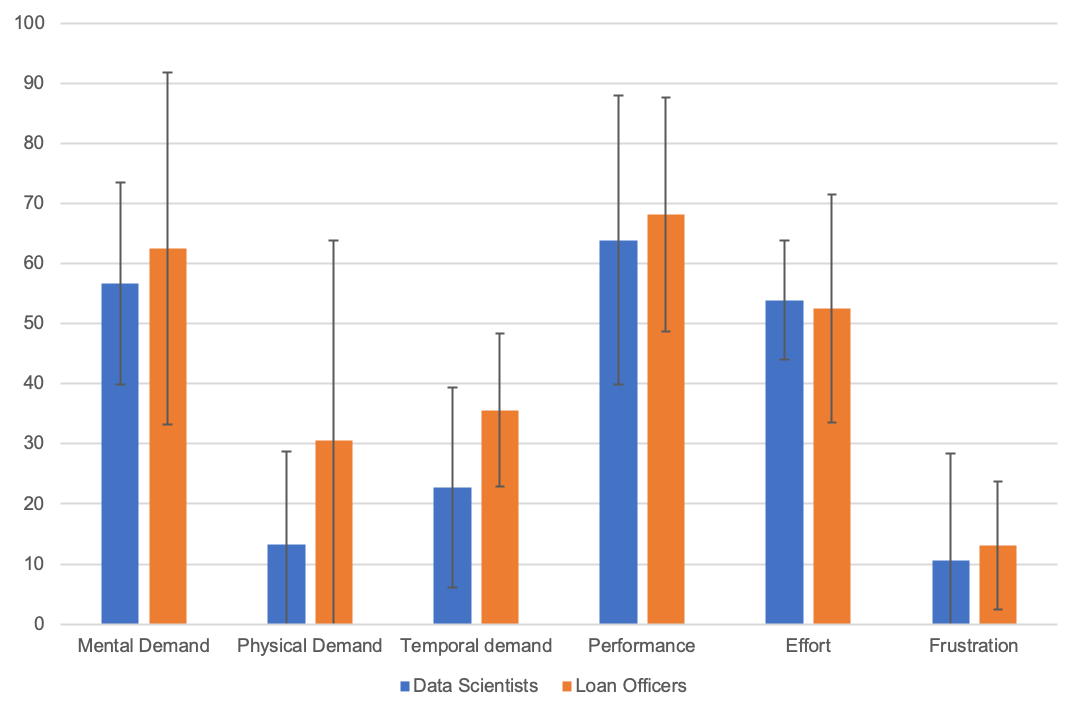}
    \caption{Mean Post-session NASA-TLX Responses. Error bars show standard deviation.}
    \label{fig:nasa}
\end{figure}

Workload as measured through the NASA-TLX questionnaire was acceptable (Figure~\ref{fig:nasa}), and there were no significant differences between data scientists and loan officers. We note that ratings on mental demand and effort were relatively high. This is likely due to the complexity of the UI and the high amount of information that needs to be processed by the user. However, this is balanced by the perceived performance which indicates that users also felt that the fairness exploration through the prototype was paying off and led to success. This shows that the UI components are suitable for both sets of stakeholders. 

\ \ \ \ \textbf{[Figure 17 near here]}

Our efforts of explaining the UI concepts in AI fairness were mainly successful, with very few areas that caused confusion. The ML and fairness concepts that were particularly difficult for loan officers to grasp were targets (5 out of 8 loan officers) and sensitive features (4  out of 8 loan officers). Frequently, targets and sensitive features were seen as \emph{any} features that mattered in decision-making, e.g. that were strongly related to the target or to sensitive features. Possibly, this would require further clarifications about input and output in ML predictive models, and what features are protected by law.   

We now turn to investigating responses to individual UI design components.

\subsubsection{Metrics Choice}
\label{section:Eval_Metrics}

An area that requires additional transparency to increase understanding is fairness metrics. Both sets of stakeholders, but in particular data scientists, requested more information about how they were calculated, which ones should be selected, and how they should be applied. This reflects the fact that in general they were not entirely trusted as a method for assessing fairness, for example:  
\begin{quote}
    ``These five metrics would have trouble letting me understand, won't allow me to understand well if I am being discriminatory because I don't see in these descriptions the way they are used.'' - ED03
\end{quote}

The custom metric builder in the UI was very popular: 7 data scientists and 4 loan officers stated that they would use it to create metrics. While data scientists stated that they would use this to create metrics that were not included in the UI by default, loan officers wanted to create a new \emph{feature} to measure affordability for \emph{each} application; it seems that they were trying to define a similarity metric in the sense of \citep{Cynthia2012Fairness}. For this, they wanted to define and compute affordability by selecting other features, and then, if the applicant was able to afford the loan, use this as an acceptance criterion, independent of other (sensitive) features.  Of course, a challenge with this approach is to define affordability in a fair way. 

\subsubsection{Causal Graph}
\label{section:Eval_CausalGraph}

Responses indicated that there were usually no problems in understanding the causal graph; all users considered it very intuitive, for example:
\begin{quote}
    "There are also indirect effects, indirect ways that age can affect the result. I think that it is a really interesting picture" - ED03
\end{quote}

Encoding important relationships as thicker lines helped users to focus on the problematic relationships, such as ones connected to sensitive features, which possibly cause fairness issues, or to the target, which directly affected the final loan decisions. Highlighting sensitive features was mentioned positively by all data scientists and 6 loan officers, while highlighting important relationships were mentioned by 8 data scientists and 7 loan officers. Users also said that the causal graph encouraged exploration of relationships between features that were not directly affecting the target, or that it encouraged exploring features that were not sensitive. 

Using the casual graph, loan officers were able to confirm or reject hypotheses they had about the data. For example: 
\begin{quote}
    ``It’s clearer for me that age is an important feature more than gender or citizenship, that's an idea I had before because of my experience, and I told you before that age is more important but now, I can realise it even looking at the chart, it's like a confirmation.'' – EL03
\end{quote}

This can be useful in investigating decision-making biases, and possible discriminatory practices, based on domain expertise.

Users could simplify the causal graph to drill down into exploring features of interest in more detail, and all participants found this useful. Participants either concentrated on sensitive features and their relationships to the target, or to hide features that were considered irrelevant or unimportant, usually because they did not have strong relationships with other features.

\subsubsection{Features and Relationships Information} 
\label{section:Eval_FeaturesAndRelationships}

Both sets of stakeholders paid attention to the fairness metrics in the information component or metric indicator in the causal graph. However, possibly because it was not clear what these fairness metrics captured or how they could be used, they were much more interested in exploring the relationships. This was useful to explore possible biases and discrimination in decision-making. For example: 
\begin{quote}
    ``ah that's interesting to see the acceptance rating on the age, so we can appreciate that it's growing, there's just one that is decreasing here. Also, we can see the number of people for each bucket, this is useful. Yeah, the majority are in their 40s and 50s.'' – ED09
    \\
\end{quote}

When exploring the model, both sets of stakeholders found the information provided useful for exploring possible issues in the decision-making process and identify fairness issues. Nine data scientists and 6 loan officers stated that feature importance was useful; 4 data scientists and 3 loan officers commented on the usefulness of the prediction confidence indicators; 7 data scientists and 5 loan officers mentioned the usefulness of seeing feature weightings for individual applications. 

\subsubsection{Feature Combinations}
\label{section:Eval_FeatureCombinations}
Users were able to combine features and explore their relationship to the target through the Group Combination UI component. Nine data scientists and 7 loan officers stated that they would use the combination feature to help them investigate bias. For example:
\begin{quote}
    \\
    ``We can understand the relation between age, citizenship, net monthly income in order to obtain a loan. And as I expected the rate is growing when net monthly income is growing, citizenship national and the age is around 40. I expected this result, this acceptance rate. I believed this acceptance rate could have been higher than 54.2\%\ldots I think this tool is fantastic because you can understand where, in what kind of area we have to concentrate in order to avoid a lot of rejected, to understand why there are a lot of rejected, and to help avoid this result.'' – EL08
\end{quote}

However, there were some concerns voiced by participants around scalability of this functionality to include more feature intersections (we limited this to 3) or how to facilitate easy comparisons between subgroups (we showed a percentage). 

\subsubsection{Dataset}
\label{section:Eval_Dataset}
Having access to the raw data was appreciated by all participants because it facilitated the investigation of individual fairness. This was especially surprising since the workshops reported in section~\ref{section:criteria} seemed to show that data scientists were not as interested in exploring the dataset from this perspective. 

Linking the dataset component to the feature and relationship information component was especially appreciated:
\begin{quote}
    ``[It would be nice to] click on this grey area and click to see how many male or how many females are in this grey area [the credit risk level group in the bar]. [\ldots] Ahh ok the dataset updates, cool, cool, this is very useful. This is useful because here I can see the records and understand why, and if there is any error or something like that\ldots to understand better the data.'' - ED06
\end{quote}

Participants stated that they liked how the dataset information was interconnected with other UI components which allowed them to `trace' the decision-making through to the raw data.

\subsubsection{Compare Applications}
\label{section:Eval_CompareApplications}
The compare applications component supported the exploration of individual fairness further. All data scientists and 6 loan officers stated that they would use this component. Participants contrasted individual applications in an attempt to explain alternative outcomes and to identify the features and their values that could have impacted the decisions, for example:
\begin{quote}
    \\
    ``I can compare the selected one that has been rejected with this one that has been approved, in order to see if there is some difference that has affected the application. This is important because for example the first thing that I see is that the loan amount is very high in the one that has been rejected, while this one is not so high, probably one of the most important reasons that the application has been rejected for the loan amount.'' – EL06
\end{quote}

\section{Discussion}
Our work has investigated the design space to support domain experts and data scientists to investigate fairness, and how a UI could be implemented for the loan applications domain and more generally to fulfill their requirements. We will now discuss, first, the limitations of our study, second, how current tools and our prototype cover this design space, and third, implications for further extensions to FairHIL. We conclude by discussing three areas of supporting human-in-the-loop fairness that warrant further attention.

\subsection{Limitations}
The work presented here has some limitations in its scope. First, our workshops and evaluation drew on a relatively small sample of participants, and thus cannot generate any quantitative findings. However, the value of a qualitative approach is to provide rich insights into users' needs that is able to generate novel UI designs. A related issue is that we have not implemented a fully working prototype that could be evaluated in a larger field trial, embedded in the practices of domain experts and data scientists. Both of these concerns would need to be addressed in further work that goes beyond the scope of this paper.

Second, our investigations were only situated in the loan applications domain. We have already highlighted that this might affect the datasets used, the ML models and hence also the UI design and visualizations. Furthermore, we involved loan officers and data scientists from one organization. While we do not anticipate that practices differ significantly between banking institutions in terms of loan applications, we do not know how fairness investigations extend to other stakeholders in loan applications, e.g. loan applicants or regulatory bodies. Similarly, domain experts and data scientists in other domains might have different criteria and ways to assess fairness which warrants further investigation.

\subsection{Coverage of design space by existing tools}
Table~\ref{tab:DesignSpace} shows how requirements that we uncovered during our workshops---which constitute the design space for tools to support investigating fairness---map to functionality by existing tools, including FairHIL. Currently, there are no tools that fulfill \emph{all} of the requirements that we uncovered during our work. However, FairHIL is so far the first to integrate most of these requirements to support both stakeholder groups to investigate fairness. Still, particular gaps exist in providing advanced data manipulation functionality, comparing individual cases, and mitigating fairness by adjusting feature weights or 'deleting' features. These are areas that future design endeavours could target.

\ \ \ \ \textbf{[Table 5 near here]}

\subsection{Design Implications for Human-in-the-loop Fairness Tools}
Our findings have concrete ramifications for the design of human-in-the-loop fairness UIs. We here summarize these insights.

While metrics are useful for assessing fairness, data scientists sometimes had difficulties in understanding fairness metrics, as found in section~\ref{section:Eval_Metrics}. Hence, in addition to describing the metrics on a high level, as we did in FairHIL, the details of how they are calculated should be given. Furthermore, users will need to be given guidance as to the expected ranges of these metrics, or when a metric indicates that there are discriminatory outcomes. This information could be integrated into visualizations. Loan officers would also benefit from this improvement, since they have less background knowledge about fairness and ML concepts.  

From \ref{section:Eval_FeaturesAndRelationships}, we found that allowing users to see the features' relationship to the target and to each other was pivotal. We suggest that this could be used to identify potential biases in decision-making, and that this could lead to further discussions within the organization for changing policy or applying fairness mitigation.  

As we have seen in \ref{section:Eval_FeatureCombinations}, exploring subgroup fairness was viewed positively by both stakeholder groups and could be a powerful way to explore intersectional bias  \citep{kearns2018preventing}. However, further work needs to consider how to extend combinations of features, possibly automatically as discussed in section~\ref{section:combinations}, while easing comparisons between subgroups through visual means. To make the assessment of feature combination more effective, it is necessary to extend the number of features taken into account when comparing the combinations, and to show the number of applications included in a specific subgroup.

Users want a way to explore the raw data and its relationship to the target, as we found in \ref{section:Eval_Dataset}. This allows users to trace decisions back to individual data instances, and gain a deeper understanding of the data as it relates to fairness. Current interfaces rarely support this \citep{Ahn2020FairSight} yet its inclusion, especially to drive further detailed comparisons with other applications is warranted.  
Our work has shown that comparing applications was intuitive and useful for assessing fairness in loan applications, as we found in \ref{section:Eval_CompareApplications}. Yet there are few human-in-the-loop fairness UIs, FairSight \citep{Ahn2020FairSight} being a notable exception, that currently support this approach. This component also clearly supports the exploration of individual fairness in loan decisions, helpful for both stakeholder groups. Yet, the compare applications component could be further extended.  While only data scientists asked to compare 'clusters' i.e. more than one application, this ability maybe useful for both stakeholder groups. Identifying these subgroups could also be automatically supported through, for example, k-nearest neighbor approaches. Comparisons could be extended even further, by letting users choose similarity metrics, or by letting the user build a custom similarity metric by re-weighting features.  

\subsection{Supporting the Human-in-the-loop Fairness Process}
Our work also raises three wider questions.
\subsubsection{How to Explain ML and Fairness Concepts?}
Our findings confirm that transparency and explanations of ML and fairness concepts are necessary \citep{lim_assessing_2009, hohman_gamut_2019}. Adopting common XAI approaches \citep{kulesza_principles_2015, Ribeiro2016WhyshouldI,Lundberg2017UnifiedApproach} to explain models in a way that is understood by non-technical domain experts and technically-savvy data scientists alike might be one way forward. To ensure understandability, information needs to be carefully encoded to support intuitive exploration and easy visual processing. While general principles for explaining ML concepts have been suggested e.g. "be iterative, sound and complete but do not overwhelm" \citep{kulesza_principles_2015}, there is still much work to be done to translate this into concrete design guidelines or design patterns.

A lack of understanding around fairness measures often meant that they are not transparent as to what they capture, how they are calculated and how they should be used. For example, in our studies, both domain experts and data scientists wanted more information and explanations of when measures indicated unfairness or discrimination. There also seemed confusion when decision-making bias tipped over into unfairness, for example, older applicants had an easier time getting a loan because their credit rating was higher due to a longer credit history yet does this mean it is unfair? Fairness measures could be made more transparent through a UI but solving these issues might also involve education of stakeholders around ML concepts and fairness principles.

\subsubsection{What is fair?}
It is becoming clear from our as well as others' work~\citep{kasinidou2021agree, saxena2020fairness, woodruff2018qualitative, lee2019procedural} that different stakeholders might have different conceptualizations of what fairness is, how it should be measured, and how fairness considerations should be integrated into the wider decision-making process when using ML. We found that in our domain, loan officers preferred to pay attention to aspects related to individual fairness while data scientists tend to employ notions of group fairness. It could be argued that existing fairness measures capture both of these perspectives, and thus could be integrated in a UI to support both domain experts and data scientists. However, as previously covered in section~\ref{section:relwork_fairness}, often these measures are not compatible or there needs to be a trade-off between fairness measures and accuracy. So who gets to decide which fairness measures to apply? We suggest that this is a question that urgently needs attention within an organizational and also a socio-technical context.

\subsubsection{How and when to investigate fairness?}
Our study suggests that requirements to support investigating fairness are very rich and varied. While we found that domain experts and data scientists might go about investigating fairness slightly differently, it was surprising to us that in the main these two stakeholder groups agreed on the information and functionality required for them to ably investigate fairness. This begs the question whether different toolkits are indeed necessary for these user groups, or whether we can find a way to communicate effectively across different user types. Current perspectives in XAI argue that different UIs might be necessary for different stakeholder groups \citep{gunning_xaiexplainable_2019} but our work suggests that it might be possible to craft UIs that suit a variety of users.

An interesting yet open question is how fairness investigations should be integrated into the AI model development. Most existing tools, such as What-if Tool and AI Fairness 360, adopt a mitigation strategy after AI models have been developed. While tools such as ours and Silva can be used on datasets as well as AI models, urgent discussion is needed on how to integrate fairness in all steps in the AI development process, from data collection to after-deployment.

\section{Conclusion}
In this paper, we presented our findings from a series of investigations to understand the design space to support investigating AI fairness for domain experts and data scientists. We ran design workshops to gather requirements for human-in-the-loop fairness UIs. We then presented 7 UI components that support data scientists as well as loan officers in assessing fairness and finding potential fairness issues, instantiated in FairHIL, a prototype UI. We addressed how these UI components could be adapted and extended for other domains. Finally, we evaluated FairHIL through a user study. 

Our findings show that:
\begin{itemize}
    \item Requirements underlying the design space are rich and varied. Surprisingly, domain experts and data scientists share many requirements to investigate fairness. 
    \item AI and fairness concepts need to be carefully explained, especially how they should be applied and what might indicate fairness issues.
    \item Causal graphs, alongside relevant and easy-to-process information about features and relationships, were an intuitive way to support fairness investigations by both stakeholder groups.
    \item Both stakeholder groups appreciated access to the underlying data, and the ability to explore and compare individual applications.
    \item Comparing subgroups is important to users.
\end{itemize}

This work will help researchers, designers and developers to better understand how domain experts and data scientists investigate fairness, and to build better human-centred AI tools to support them. We hope that this will lead to AI systems that are fairer and more transparent for everyone.

\section*{Acknowledgement}
Work by City, University of London was supported by Fujitsu Limited under a research contract.

\section*{Disclosure statement}
We don’t have a conflict of interest. 

\bibliographystyle{apacite}
\bibliography{ref}

\begin{thebibliography}{}

\bibitem [\protect \citeauthoryear {%
Agarwal%
, Beygelzimer%
, Dud{\'\i}k%
, Langford%
\BCBL {}\ \BBA {} Wallach%
}{%
Agarwal%
\ \protect \BOthers {.}}{%
{\protect \APACyear {2018}}%
}]{%
agarwal2018reductions}
\APACinsertmetastar {%
agarwal2018reductions}%
\begin{APACrefauthors}%
Agarwal, A.%
, Beygelzimer, A.%
, Dud{\'\i}k, M.%
, Langford, J.%
\BCBL {}\ \BBA {} Wallach, H.%
\end{APACrefauthors}%
\unskip\
\newblock
\APACrefYearMonthDay{2018}{}{}.
\newblock
{\BBOQ}\APACrefatitle {A reductions approach to fair classification} {A
  reductions approach to fair classification}.{\BBCQ}
\newblock
\APACjournalVolNumPages{arXiv preprint arXiv:1803.02453}{}{}{}.
\PrintBackRefs{\CurrentBib}

\bibitem [\protect \citeauthoryear {%
{Ahn}%
\ \BBA {} {Lin}%
}{%
{Ahn}%
\ \BBA {} {Lin}%
}{%
{\protect \APACyear {2020}}%
}]{%
Ahn2020FairSight}
\APACinsertmetastar {%
Ahn2020FairSight}%
\begin{APACrefauthors}%
{Ahn}, Y.%
\BCBT {}\ \BBA {} {Lin}, Y\BPBI R.%
\end{APACrefauthors}%
\unskip\
\newblock
\APACrefYearMonthDay{2020}{}{}.
\newblock
{\BBOQ}\APACrefatitle {FairSight: Visual Analytics for Fairness in Decision
  Making} {Fairsight: Visual analytics for fairness in decision making}.{\BBCQ}
\newblock
\APACjournalVolNumPages{IEEE Transactions on Visualization and Computer
  Graphics}{26}{1}{1086-1095}.
\newblock
\begin{APACrefDOI} \doi{10.1109/TVCG.2019.2934262} \end{APACrefDOI}
\PrintBackRefs{\CurrentBib}

\bibitem [\protect \citeauthoryear {%
Barocas%
, Hardt%
\BCBL {}\ \BBA {} Narayanan%
}{%
Barocas%
\ \protect \BOthers {.}}{%
{\protect \APACyear {2019}}%
}]{%
barocas-hardt-narayanan}
\APACinsertmetastar {%
barocas-hardt-narayanan}%
\begin{APACrefauthors}%
Barocas, S.%
, Hardt, M.%
\BCBL {}\ \BBA {} Narayanan, A.%
\end{APACrefauthors}%
\unskip\
\newblock
\APACrefYear{2019}.
\newblock
\APACrefbtitle {Fairness and Machine Learning} {Fairness and machine learning}.
\newblock
\APACaddressPublisher{}{fairmlbook.org}.
\newblock
\APACrefnote{\url{http://www.fairmlbook.org}}
\PrintBackRefs{\CurrentBib}

\bibitem [\protect \citeauthoryear {%
Barocas%
\ \BBA {} Selbst%
}{%
Barocas%
\ \BBA {} Selbst%
}{%
{\protect \APACyear {2016}}%
}]{%
barocas2016BigData}
\APACinsertmetastar {%
barocas2016BigData}%
\begin{APACrefauthors}%
Barocas, S.%
\BCBT {}\ \BBA {} Selbst, A\BPBI D.%
\end{APACrefauthors}%
\unskip\
\newblock
\APACrefYearMonthDay{2016}{}{}.
\newblock
{\BBOQ}\APACrefatitle {Big Data's Disparate Impact} {Big data's disparate
  impact}.{\BBCQ}
\newblock
\APACjournalVolNumPages{California Law Review}{104}{3}{671--732}.
\newblock
\begin{APACrefURL} \url{http://www.jstor.org/stable/24758720} \end{APACrefURL}
\PrintBackRefs{\CurrentBib}

\bibitem [\protect \citeauthoryear {%
Bellamy%
\ \protect \BOthers {.}}{%
Bellamy%
\ \protect \BOthers {.}}{%
{\protect \APACyear {2018}}%
{\protect \APACexlab {{\protect \BCnt {1}}}}}]{%
bellamy2018ai}
\APACinsertmetastar {%
bellamy2018ai}%
\begin{APACrefauthors}%
Bellamy, R\BPBI K\BPBI E.%
, Dey, K.%
, Hind, M.%
, Hoffman, S\BPBI C.%
, Houde, S.%
, Kannan, K.%
\BDBL {}Zhang, Y.%
\end{APACrefauthors}%
\unskip\
\newblock
\APACrefYearMonthDay{2018{\protect \BCnt {1}}}{}{}.
\newblock
\APACrefbtitle {AI Fairness 360: An Extensible Toolkit for Detecting,
  Understanding, and Mitigating Unwanted Algorithmic Bias.} {Ai fairness 360:
  An extensible toolkit for detecting, understanding, and mitigating unwanted
  algorithmic bias.}
\PrintBackRefs{\CurrentBib}

\bibitem [\protect \citeauthoryear {%
Bellamy%
\ \protect \BOthers {.}}{%
Bellamy%
\ \protect \BOthers {.}}{%
{\protect \APACyear {2018}}%
{\protect \APACexlab {{\protect \BCnt {2}}}}}]{%
aif360-oct-2018}
\APACinsertmetastar {%
aif360-oct-2018}%
\begin{APACrefauthors}%
Bellamy, R\BPBI K\BPBI E.%
, Dey, K.%
, Hind, M.%
, Hoffman, S\BPBI C.%
, Houde, S.%
, Kannan, K.%
\BDBL {}Zhang, Y.%
\end{APACrefauthors}%
\unskip\
\newblock
\APACrefYearMonthDay{2018{\protect \BCnt {2}}}{{\APACmonth{10}}}{}.
\newblock
\APACrefbtitle {{AI Fairness} 360: An Extensible Toolkit for Detecting,
  Understanding, and Mitigating Unwanted Algorithmic Bias.} {{AI Fairness} 360:
  An extensible toolkit for detecting, understanding, and mitigating unwanted
  algorithmic bias.}
\newblock
\begin{APACrefURL} \url{https://arxiv.org/abs/1810.01943} \end{APACrefURL}
\PrintBackRefs{\CurrentBib}

\bibitem [\protect \citeauthoryear {%
Berman%
, Murphy-Berman%
\BCBL {}\ \BBA {} Singh%
}{%
Berman%
\ \protect \BOthers {.}}{%
{\protect \APACyear {1985}}%
}]{%
Berman1985CrossCultural}
\APACinsertmetastar {%
Berman1985CrossCultural}%
\begin{APACrefauthors}%
Berman, J\BPBI J.%
, Murphy-Berman, V.%
\BCBL {}\ \BBA {} Singh, P.%
\end{APACrefauthors}%
\unskip\
\newblock
\APACrefYearMonthDay{1985}{}{}.
\newblock
{\BBOQ}\APACrefatitle {Cross-Cultural Similarities and Differences in
  Perceptions of Fairness} {Cross-cultural similarities and differences in
  perceptions of fairness}.{\BBCQ}
\newblock
\APACjournalVolNumPages{Journal of Cross-Cultural Psychology}{16}{1}{55-67}.
\newblock
\begin{APACrefURL} \url{https://doi.org/10.1177/0022002185016001005}
  \end{APACrefURL}
\newblock
\begin{APACrefDOI} \doi{10.1177/0022002185016001005} \end{APACrefDOI}
\PrintBackRefs{\CurrentBib}

\bibitem [\protect \citeauthoryear {%
Binns%
\ \protect \BOthers {.}}{%
Binns%
\ \protect \BOthers {.}}{%
{\protect \APACyear {2018}}%
}]{%
Binns2018ItsReducing}
\APACinsertmetastar {%
Binns2018ItsReducing}%
\begin{APACrefauthors}%
Binns, R.%
, Van~Kleek, M.%
, Veale, M.%
, Lyngs, U.%
, Zhao, J.%
\BCBL {}\ \BBA {} Shadbolt, N.%
\end{APACrefauthors}%
\unskip\
\newblock
\APACrefYearMonthDay{2018}{}{}.
\newblock
{\BBOQ}\APACrefatitle {'It's Reducing a Human Being to a Percentage':
  Perceptions of Justice in Algorithmic Decisions} {'it's reducing a human
  being to a percentage': Perceptions of justice in algorithmic
  decisions}.{\BBCQ}
\newblock
\BIn{} \APACrefbtitle {Proceedings of the 2018 CHI Conference on Human Factors
  in Computing Systems} {Proceedings of the 2018 chi conference on human
  factors in computing systems}\ (\BPG~1–14).
\newblock
\APACaddressPublisher{New York, NY, USA}{Association for Computing Machinery}.
\newblock
\begin{APACrefURL} \url{https://doi.org/10.1145/3173574.3173951}
  \end{APACrefURL}
\newblock
\begin{APACrefDOI} \doi{10.1145/3173574.3173951} \end{APACrefDOI}
\PrintBackRefs{\CurrentBib}

\bibitem [\protect \citeauthoryear {%
Bird%
\ \protect \BOthers {.}}{%
Bird%
\ \protect \BOthers {.}}{%
{\protect \APACyear {2020}}%
}]{%
bird2020fairlearn}
\APACinsertmetastar {%
bird2020fairlearn}%
\begin{APACrefauthors}%
Bird, S.%
, Dud{\'i}k, M.%
, Edgar, R.%
, Horn, B.%
, Lutz, R.%
, Milan, V.%
\BDBL {}Walker, K.%
\end{APACrefauthors}%
\unskip\
\newblock
\APACrefYearMonthDay{2020}{May}{}.
\newblock
\APACrefbtitle {Fairlearn: A toolkit for assessing and improving fairness in
  {AI}} {Fairlearn: A toolkit for assessing and improving fairness in {AI}}\
  \APACbVolEdTR{}{\BTR{}\ \BNUM\ MSR-TR-2020-32}.
\newblock
\APACaddressInstitution{}{Microsoft}.
\newblock
\begin{APACrefURL}
  \url{https://www.microsoft.com/en-us/research/publication/fairlearn-a-toolkit-for-assessing-and-improving-fairness-in-ai/}
  \end{APACrefURL}
\PrintBackRefs{\CurrentBib}

\bibitem [\protect \citeauthoryear {%
Blake%
\ \protect \BOthers {.}}{%
Blake%
\ \protect \BOthers {.}}{%
{\protect \APACyear {2015}}%
}]{%
Blake2015NatureOntologyOfFairness}
\APACinsertmetastar {%
Blake2015NatureOntologyOfFairness}%
\begin{APACrefauthors}%
Blake, P\BPBI R.%
, McAuliffe, K.%
, Corbit, J.%
, Callaghan, T\BPBI C.%
, Barry, O.%
, Bowie, A.%
\BDBL {}Warneken, F.%
\end{APACrefauthors}%
\unskip\
\newblock
\APACrefYearMonthDay{2015}{November}{}.
\newblock
{\BBOQ}\APACrefatitle {The ontogeny of fairness in seven societies} {The
  ontogeny of fairness in seven societies}.{\BBCQ}
\newblock
\APACjournalVolNumPages{Nature}{528}{}{258-261}.
\newblock
\begin{APACrefDOI} \doi{10.1038/nature15703} \end{APACrefDOI}
\PrintBackRefs{\CurrentBib}

\bibitem [\protect \citeauthoryear {%
Bolton%
, Keh%
\BCBL {}\ \BBA {} Alba%
}{%
Bolton%
\ \protect \BOthers {.}}{%
{\protect \APACyear {2010}}%
}]{%
Bolton2010.HowDoPriceFairness}
\APACinsertmetastar {%
Bolton2010.HowDoPriceFairness}%
\begin{APACrefauthors}%
Bolton, L\BPBI E.%
, Keh, H\BPBI T.%
\BCBL {}\ \BBA {} Alba, J\BPBI W.%
\end{APACrefauthors}%
\unskip\
\newblock
\APACrefYearMonthDay{2010}{}{}.
\newblock
{\BBOQ}\APACrefatitle {How Do Price Fairness Perceptions Differ across
  Culture?} {How do price fairness perceptions differ across culture?}{\BBCQ}
\newblock
\APACjournalVolNumPages{Journal of Marketing Research}{47}{3}{564-576}.
\newblock
\begin{APACrefURL} \url{https://doi.org/10.1509/jmkr.47.3.564} \end{APACrefURL}
\newblock
\begin{APACrefDOI} \doi{10.1509/jmkr.47.3.564} \end{APACrefDOI}
\PrintBackRefs{\CurrentBib}

\bibitem [\protect \citeauthoryear {%
{Cabrera}%
\ \protect \BOthers {.}}{%
{Cabrera}%
\ \protect \BOthers {.}}{%
{\protect \APACyear {2019}}%
}]{%
Cabrera2019Fairvis}
\APACinsertmetastar {%
Cabrera2019Fairvis}%
\begin{APACrefauthors}%
{Cabrera}, A\BPBI A.%
, {Epperson}, W.%
, {Hohman}, F.%
, {Kahng}, M.%
, {Morgenstern}, J.%
\BCBL {}\ \BBA {} {Chau}, D\BPBI H.%
\end{APACrefauthors}%
\unskip\
\newblock
\APACrefYearMonthDay{2019}{}{}.
\newblock
{\BBOQ}\APACrefatitle {FAIRVIS: Visual Analytics for Discovering Intersectional
  Bias in Machine Learning} {Fairvis: Visual analytics for discovering
  intersectional bias in machine learning}.{\BBCQ}
\newblock
\BIn{} \APACrefbtitle {2019 IEEE Conference on Visual Analytics Science and
  Technology (VAST)} {2019 ieee conference on visual analytics science and
  technology (vast)}\ (\BPG~46-56).
\newblock
\begin{APACrefDOI} \doi{10.1109/VAST47406.2019.8986948} \end{APACrefDOI}
\PrintBackRefs{\CurrentBib}

\bibitem [\protect \citeauthoryear {%
Caliskan%
, Bryson%
\BCBL {}\ \BBA {} Narayanan%
}{%
Caliskan%
\ \protect \BOthers {.}}{%
{\protect \APACyear {2017}}%
}]{%
Caliskan2017Semantics}
\APACinsertmetastar {%
Caliskan2017Semantics}%
\begin{APACrefauthors}%
Caliskan, A.%
, Bryson, J\BPBI J.%
\BCBL {}\ \BBA {} Narayanan, A.%
\end{APACrefauthors}%
\unskip\
\newblock
\APACrefYearMonthDay{2017}{}{}.
\newblock
{\BBOQ}\APACrefatitle {Semantics derived automatically from language corpora
  contain human-like biases} {Semantics derived automatically from language
  corpora contain human-like biases}.{\BBCQ}
\newblock
\APACjournalVolNumPages{Science}{356}{6334}{183--186}.
\newblock
\begin{APACrefURL} \url{https://science.sciencemag.org/content/356/6334/183}
  \end{APACrefURL}
\newblock
\begin{APACrefDOI} \doi{10.1126/science.aal4230} \end{APACrefDOI}
\PrintBackRefs{\CurrentBib}

\bibitem [\protect \citeauthoryear {%
Calmon%
, Wei%
, Vinzamuri%
, Ramamurthy%
\BCBL {}\ \BBA {} Varshney%
}{%
Calmon%
\ \protect \BOthers {.}}{%
{\protect \APACyear {2017}}%
}]{%
calmon2017optimized}
\APACinsertmetastar {%
calmon2017optimized}%
\begin{APACrefauthors}%
Calmon, F.%
, Wei, D.%
, Vinzamuri, B.%
, Ramamurthy, K\BPBI N.%
\BCBL {}\ \BBA {} Varshney, K\BPBI R.%
\end{APACrefauthors}%
\unskip\
\newblock
\APACrefYearMonthDay{2017}{}{}.
\newblock
{\BBOQ}\APACrefatitle {Optimized pre-processing for discrimination prevention}
  {Optimized pre-processing for discrimination prevention}.{\BBCQ}
\newblock
\BIn{} \APACrefbtitle {Advances in Neural Information Processing Systems}
  {Advances in neural information processing systems}\ (\BPGS\ 3992--4001).
\PrintBackRefs{\CurrentBib}

\bibitem [\protect \citeauthoryear {%
Cheng%
\ \protect \BOthers {.}}{%
Cheng%
\ \protect \BOthers {.}}{%
{\protect \APACyear {2021}}%
}]{%
cheng_soliciting_2021}
\APACinsertmetastar {%
cheng_soliciting_2021}%
\begin{APACrefauthors}%
Cheng, H\BHBI F.%
, Stapleton, L.%
, Wang, R.%
, Bullock, P.%
, Chouldechova, A.%
, Wu, Z\BPBI S\BPBI S.%
\BCBL {}\ \BBA {} Zhu, H.%
\end{APACrefauthors}%
\unskip\
\newblock
\APACrefYearMonthDay{2021}{{\APACmonth{05}}}{}.
\newblock
{\BBOQ}\APACrefatitle {Soliciting {Stakeholders}\&\#x2019; {Fairness} {Notions}
  in {Child} {Maltreatment} {Predictive} {Systems}} {Soliciting
  {Stakeholders}\&\#x2019; {Fairness} {Notions} in {Child} {Maltreatment}
  {Predictive} {Systems}}.{\BBCQ}
\newblock
\BIn{} \APACrefbtitle {Proceedings of the 2021 {CHI} {Conference} on {Human}
  {Factors} in {Computing} {Systems}} {Proceedings of the 2021 {CHI}
  {Conference} on {Human} {Factors} in {Computing} {Systems}}\ (\BPGS\ 1--17).
\newblock
\APACaddressPublisher{New York, NY, USA}{Association for Computing Machinery}.
\newblock
\begin{APACrefURL} [{2021-08-18}]\url{https://doi.org/10.1145/3411764.3445308}
  \end{APACrefURL}
\PrintBackRefs{\CurrentBib}

\bibitem [\protect \citeauthoryear {%
Chouldechova%
}{%
Chouldechova%
}{%
{\protect \APACyear {2017}}%
}]{%
chouldechova2017fair}
\APACinsertmetastar {%
chouldechova2017fair}%
\begin{APACrefauthors}%
Chouldechova, A.%
\end{APACrefauthors}%
\unskip\
\newblock
\APACrefYearMonthDay{2017}{}{}.
\newblock
{\BBOQ}\APACrefatitle {Fair prediction with disparate impact: A study of bias
  in recidivism prediction instruments} {Fair prediction with disparate impact:
  A study of bias in recidivism prediction instruments}.{\BBCQ}
\newblock
\APACjournalVolNumPages{Big data}{5}{2}{153--163}.
\PrintBackRefs{\CurrentBib}

\bibitem [\protect \citeauthoryear {%
Clarke%
, Braun%
\BCBL {}\ \BBA {} Hayfield%
}{%
Clarke%
\ \protect \BOthers {.}}{%
{\protect \APACyear {2015}}%
}]{%
clarke2015thematic}
\APACinsertmetastar {%
clarke2015thematic}%
\begin{APACrefauthors}%
Clarke, V.%
, Braun, V.%
\BCBL {}\ \BBA {} Hayfield, N.%
\end{APACrefauthors}%
\unskip\
\newblock
\APACrefYearMonthDay{2015}{}{}.
\newblock
{\BBOQ}\APACrefatitle {Thematic analysis} {Thematic analysis}.{\BBCQ}
\newblock
\APACjournalVolNumPages{Qualitative psychology: A practical guide to research
  methods}{}{}{222--248}.
\PrintBackRefs{\CurrentBib}

\bibitem [\protect \citeauthoryear {%
Cook%
\ \BBA {} Hegtvedt%
}{%
Cook%
\ \BBA {} Hegtvedt%
}{%
{\protect \APACyear {1983}}%
}]{%
Cook1983Distributive}
\APACinsertmetastar {%
Cook1983Distributive}%
\begin{APACrefauthors}%
Cook, K\BPBI S.%
\BCBT {}\ \BBA {} Hegtvedt, K\BPBI A.%
\end{APACrefauthors}%
\unskip\
\newblock
\APACrefYearMonthDay{1983}{}{}.
\newblock
{\BBOQ}\APACrefatitle {Distributive Justice, Equity, and Equality}
  {Distributive justice, equity, and equality}.{\BBCQ}
\newblock
\APACjournalVolNumPages{Annual Review of Sociology}{9}{1}{217-241}.
\newblock
\begin{APACrefURL} \url{https://doi.org/10.1146/annurev.so.09.080183.001245}
  \end{APACrefURL}
\newblock
\begin{APACrefDOI} \doi{10.1146/annurev.so.09.080183.001245} \end{APACrefDOI}
\PrintBackRefs{\CurrentBib}

\bibitem [\protect \citeauthoryear {%
Corbett-Davies%
\ \BBA {} Goel%
}{%
Corbett-Davies%
\ \BBA {} Goel%
}{%
{\protect \APACyear {2018}}%
}]{%
corbettdavies2018measure}
\APACinsertmetastar {%
corbettdavies2018measure}%
\begin{APACrefauthors}%
Corbett-Davies, S.%
\BCBT {}\ \BBA {} Goel, S.%
\end{APACrefauthors}%
\unskip\
\newblock
\APACrefYearMonthDay{2018}{}{}.
\newblock
\APACrefbtitle {The Measure and Mismeasure of Fairness: A Critical Review of
  Fair Machine Learning.} {The measure and mismeasure of fairness: A critical
  review of fair machine learning.}
\PrintBackRefs{\CurrentBib}

\bibitem [\protect \citeauthoryear {%
Corbett-Davies%
, Pierson%
, Feller%
, Goel%
\BCBL {}\ \BBA {} Huq%
}{%
Corbett-Davies%
\ \protect \BOthers {.}}{%
{\protect \APACyear {2017}}%
}]{%
Corbett-Davis2017Algorighmic}
\APACinsertmetastar {%
Corbett-Davis2017Algorighmic}%
\begin{APACrefauthors}%
Corbett-Davies, S.%
, Pierson, E.%
, Feller, A.%
, Goel, S.%
\BCBL {}\ \BBA {} Huq, A.%
\end{APACrefauthors}%
\unskip\
\newblock
\APACrefYearMonthDay{2017}{}{}.
\newblock
{\BBOQ}\APACrefatitle {Algorithmic Decision Making and the Cost of Fairness}
  {Algorithmic decision making and the cost of fairness}.{\BBCQ}
\newblock
\BIn{} \APACrefbtitle {Proceedings of the 23rd ACM SIGKDD International
  Conference on Knowledge Discovery and Data Mining} {Proceedings of the 23rd
  acm sigkdd international conference on knowledge discovery and data mining}\
  (\BPG~797–806).
\newblock
\APACaddressPublisher{New York, NY, USA}{Association for Computing Machinery}.
\newblock
\begin{APACrefURL} \url{https://doi.org/10.1145/3097983.3098095}
  \end{APACrefURL}
\newblock
\begin{APACrefDOI} \doi{10.1145/3097983.3098095} \end{APACrefDOI}
\PrintBackRefs{\CurrentBib}

\bibitem [\protect \citeauthoryear {%
Dodge%
, Liao%
, Zhang%
, Bellamy%
\BCBL {}\ \BBA {} Dugan%
}{%
Dodge%
\ \protect \BOthers {.}}{%
{\protect \APACyear {2019}}%
}]{%
Dodge2019Explaining}
\APACinsertmetastar {%
Dodge2019Explaining}%
\begin{APACrefauthors}%
Dodge, J.%
, Liao, Q\BPBI V.%
, Zhang, Y.%
, Bellamy, R\BPBI K\BPBI E.%
\BCBL {}\ \BBA {} Dugan, C.%
\end{APACrefauthors}%
\unskip\
\newblock
\APACrefYearMonthDay{2019}{}{}.
\newblock
{\BBOQ}\APACrefatitle {Explaining Models: An Empirical Study of How
  Explanations Impact Fairness Judgment} {Explaining models: An empirical study
  of how explanations impact fairness judgment}.{\BBCQ}
\newblock
\BIn{} \APACrefbtitle {Proceedings of the 24th International Conference on
  Intelligent User Interfaces} {Proceedings of the 24th international
  conference on intelligent user interfaces}\ (\BPG~275–285).
\newblock
\APACaddressPublisher{New York, NY, USA}{Association for Computing Machinery}.
\newblock
\begin{APACrefURL} \url{https://doi.org/10.1145/3301275.3302310}
  \end{APACrefURL}
\newblock
\begin{APACrefDOI} \doi{10.1145/3301275.3302310} \end{APACrefDOI}
\PrintBackRefs{\CurrentBib}

\bibitem [\protect \citeauthoryear {%
Dwork%
, Hardt%
, Pitassi%
, Reingold%
\BCBL {}\ \BBA {} Zemel%
}{%
Dwork%
\ \protect \BOthers {.}}{%
{\protect \APACyear {2012}}%
}]{%
Cynthia2012Fairness}
\APACinsertmetastar {%
Cynthia2012Fairness}%
\begin{APACrefauthors}%
Dwork, C.%
, Hardt, M.%
, Pitassi, T.%
, Reingold, O.%
\BCBL {}\ \BBA {} Zemel, R.%
\end{APACrefauthors}%
\unskip\
\newblock
\APACrefYearMonthDay{2012}{}{}.
\newblock
{\BBOQ}\APACrefatitle {Fairness through Awareness} {Fairness through
  awareness}.{\BBCQ}
\newblock
\BIn{} \APACrefbtitle {Proceedings of the 3rd Innovations in Theoretical
  Computer Science Conference} {Proceedings of the 3rd innovations in
  theoretical computer science conference}\ (\BPG~214–226).
\newblock
\APACaddressPublisher{New York, NY, USA}{Association for Computing Machinery}.
\newblock
\begin{APACrefURL} \url{https://doi.org/10.1145/2090236.2090255}
  \end{APACrefURL}
\newblock
\begin{APACrefDOI} \doi{10.1145/2090236.2090255} \end{APACrefDOI}
\PrintBackRefs{\CurrentBib}

\bibitem [\protect \citeauthoryear {%
Eckhoff%
}{%
Eckhoff%
}{%
{\protect \APACyear {1974}}%
}]{%
Eckhoff74Justice}
\APACinsertmetastar {%
Eckhoff74Justice}%
\begin{APACrefauthors}%
Eckhoff, T.%
\end{APACrefauthors}%
\unskip\
\newblock
\APACrefYear{1974}.
\newblock
\APACrefbtitle {Justice: Its determinants in social interaction} {Justice: Its
  determinants in social interaction}.
\newblock
\APACaddressPublisher{}{Rotterdam University Press}.
\PrintBackRefs{\CurrentBib}

\bibitem [\protect \citeauthoryear {%
Equality%
\ \BBA {} Commission%
}{%
Equality%
\ \BBA {} Commission%
}{%
{\protect \APACyear {2020}}%
}]{%
EqualityandHumanRightCommission2020Protected}
\APACinsertmetastar {%
EqualityandHumanRightCommission2020Protected}%
\begin{APACrefauthors}%
Equality%
\BCBT {}\ \BBA {} Commission, H\BPBI R.%
\end{APACrefauthors}%
\unskip\
\newblock
\APACrefYearMonthDay{2020}{}{}.
\newblock
\APACrefbtitle {Protected characteristics.} {Protected characteristics.}
\newblock
\begin{APACrefURL}
  \url{https://www.equalityhumanrights.com/en/equality-act/protected-characteristics}
  \end{APACrefURL}
\PrintBackRefs{\CurrentBib}

\bibitem [\protect \citeauthoryear {%
Friedman%
\ \BBA {} Nissenbaum%
}{%
Friedman%
\ \BBA {} Nissenbaum%
}{%
{\protect \APACyear {1996}}%
}]{%
Friedman1996Bias}
\APACinsertmetastar {%
Friedman1996Bias}%
\begin{APACrefauthors}%
Friedman, B.%
\BCBT {}\ \BBA {} Nissenbaum, H.%
\end{APACrefauthors}%
\unskip\
\newblock
\APACrefYearMonthDay{1996}{{\APACmonth{07}}}{}.
\newblock
{\BBOQ}\APACrefatitle {Bias in Computer Systems} {Bias in computer
  systems}.{\BBCQ}
\newblock
\APACjournalVolNumPages{ACM Trans. Inf. Syst.}{14}{3}{330–347}.
\newblock
\begin{APACrefURL} \url{https://doi.org/10.1145/230538.230561} \end{APACrefURL}
\newblock
\begin{APACrefDOI} \doi{10.1145/230538.230561} \end{APACrefDOI}
\PrintBackRefs{\CurrentBib}

\bibitem [\protect \citeauthoryear {%
Gunning%
\ \protect \BOthers {.}}{%
Gunning%
\ \protect \BOthers {.}}{%
{\protect \APACyear {2019}}%
}]{%
gunning_xaiexplainable_2019}
\APACinsertmetastar {%
gunning_xaiexplainable_2019}%
\begin{APACrefauthors}%
Gunning, D.%
, Stefik, M.%
, Choi, J.%
, Miller, T.%
, Stumpf, S.%
\BCBL {}\ \BBA {} Yang, G\BHBI Z.%
\end{APACrefauthors}%
\unskip\
\newblock
\APACrefYearMonthDay{2019}{{\APACmonth{12}}}{}.
\newblock
{\BBOQ}\APACrefatitle {{XAI}—{Explainable} artificial intelligence}
  {{XAI}—{Explainable} artificial intelligence}.{\BBCQ}
\newblock
\APACjournalVolNumPages{Science Robotics}{4}{37}{}.
\newblock
\begin{APACrefURL}
  [{2020-02-05}]\url{https://robotics.sciencemag.org/content/4/37/eaay7120}
  \end{APACrefURL}
\newblock
\begin{APACrefDOI} \doi{10.1126/scirobotics.aay7120} \end{APACrefDOI}
\PrintBackRefs{\CurrentBib}

\bibitem [\protect \citeauthoryear {%
Hajian%
, Bonchi%
\BCBL {}\ \BBA {} Castillo%
}{%
Hajian%
\ \protect \BOthers {.}}{%
{\protect \APACyear {2016}}%
}]{%
Hajian2016AlgorithmicBias}
\APACinsertmetastar {%
Hajian2016AlgorithmicBias}%
\begin{APACrefauthors}%
Hajian, S.%
, Bonchi, F.%
\BCBL {}\ \BBA {} Castillo, C.%
\end{APACrefauthors}%
\unskip\
\newblock
\APACrefYearMonthDay{2016}{}{}.
\newblock
{\BBOQ}\APACrefatitle {Algorithmic Bias: From Discrimination Discovery to
  Fairness-Aware Data Mining} {Algorithmic bias: From discrimination discovery
  to fairness-aware data mining}.{\BBCQ}
\newblock
\BIn{} \APACrefbtitle {Proceedings of the 22nd ACM SIGKDD International
  Conference on Knowledge Discovery and Data Mining} {Proceedings of the 22nd
  acm sigkdd international conference on knowledge discovery and data mining}\
  (\BPG~2125–2126).
\newblock
\APACaddressPublisher{New York, NY, USA}{Association for Computing Machinery}.
\newblock
\begin{APACrefURL} \url{https://doi.org/10.1145/2939672.2945386}
  \end{APACrefURL}
\newblock
\begin{APACrefDOI} \doi{10.1145/2939672.2945386} \end{APACrefDOI}
\PrintBackRefs{\CurrentBib}

\bibitem [\protect \citeauthoryear {%
Hardt%
, Price%
\BCBL {}\ \BBA {} Srebro%
}{%
Hardt%
\ \protect \BOthers {.}}{%
{\protect \APACyear {2016}}%
}]{%
hardt2016equality}
\APACinsertmetastar {%
hardt2016equality}%
\begin{APACrefauthors}%
Hardt, M.%
, Price, E.%
\BCBL {}\ \BBA {} Srebro, N.%
\end{APACrefauthors}%
\unskip\
\newblock
\APACrefYearMonthDay{2016}{}{}.
\newblock
{\BBOQ}\APACrefatitle {Equality of opportunity in supervised learning}
  {Equality of opportunity in supervised learning}.{\BBCQ}
\newblock
\BIn{} \APACrefbtitle {Advances in neural information processing systems}
  {Advances in neural information processing systems}\ (\BPGS\ 3315--3323).
\PrintBackRefs{\CurrentBib}

\bibitem [\protect \citeauthoryear {%
Hart%
\ \BBA {} Staveland%
}{%
Hart%
\ \BBA {} Staveland%
}{%
{\protect \APACyear {1988}}%
}]{%
hart_development_1988}
\APACinsertmetastar {%
hart_development_1988}%
\begin{APACrefauthors}%
Hart, S\BPBI G.%
\BCBT {}\ \BBA {} Staveland, L\BPBI E.%
\end{APACrefauthors}%
\unskip\
\newblock
\APACrefYearMonthDay{1988}{}{}.
\newblock
{\BBOQ}\APACrefatitle {Development of {NASA}-{TLX} ({Task} {Load} {Index}):
  {Results} of {Empirical} and {Theoretical} {Research}} {Development of
  {NASA}-{TLX} ({Task} {Load} {Index}): {Results} of {Empirical} and
  {Theoretical} {Research}}.{\BBCQ}
\newblock
\BIn{} {Peter A. Hancock and Najmedin Meshkati}\ (\BED), \APACrefbtitle
  {Advances in {Psychology}} {Advances in {Psychology}}\ (\BVOL\ Volume 52,
  \BPGS\ 139--183).
\newblock
\APACaddressPublisher{}{North-Holland}.
\newblock
\begin{APACrefURL}
  [{2013-11-06}]\url{http://www.sciencedirect.com/science/article/pii/S0166411508623869}
  \end{APACrefURL}
\PrintBackRefs{\CurrentBib}

\bibitem [\protect \citeauthoryear {%
Hegtvedt%
}{%
Hegtvedt%
}{%
{\protect \APACyear {2005}}%
}]{%
Karen2005DoingJustice}
\APACinsertmetastar {%
Karen2005DoingJustice}%
\begin{APACrefauthors}%
Hegtvedt, K\BPBI A.%
\end{APACrefauthors}%
\unskip\
\newblock
\APACrefYearMonthDay{2005}{}{}.
\newblock
{\BBOQ}\APACrefatitle {Doing Justice to the Group: Examining the Roles of the
  Group in Justice Research} {Doing justice to the group: Examining the roles
  of the group in justice research}.{\BBCQ}
\newblock
\APACjournalVolNumPages{Annual Review of Sociology}{31}{}{25--45}.
\newblock
\begin{APACrefURL} \url{http://www.jstor.org/stable/29737710} \end{APACrefURL}
\PrintBackRefs{\CurrentBib}

\bibitem [\protect \citeauthoryear {%
Hofstede%
}{%
Hofstede%
}{%
{\protect \APACyear {2011}}%
}]{%
hofstede_dimensionalizing_2011}
\APACinsertmetastar {%
hofstede_dimensionalizing_2011}%
\begin{APACrefauthors}%
Hofstede, G.%
\end{APACrefauthors}%
\unskip\
\newblock
\APACrefYearMonthDay{2011}{}{}.
\newblock
{\BBOQ}\APACrefatitle {Dimensionalizing cultures: {The} {Hofstede} model in
  context} {Dimensionalizing cultures: {The} {Hofstede} model in
  context}.{\BBCQ}
\newblock
\APACjournalVolNumPages{Online readings in psychology and
  culture}{2}{1}{2307--0919}.
\PrintBackRefs{\CurrentBib}

\bibitem [\protect \citeauthoryear {%
Hohman%
, Head%
, Caruana%
, DeLine%
\BCBL {}\ \BBA {} Drucker%
}{%
Hohman%
\ \protect \BOthers {.}}{%
{\protect \APACyear {2019}}%
}]{%
hohman_gamut_2019}
\APACinsertmetastar {%
hohman_gamut_2019}%
\begin{APACrefauthors}%
Hohman, F.%
, Head, A.%
, Caruana, R.%
, DeLine, R.%
\BCBL {}\ \BBA {} Drucker, S\BPBI M.%
\end{APACrefauthors}%
\unskip\
\newblock
\APACrefYearMonthDay{2019}{{\APACmonth{05}}}{}.
\newblock
{\BBOQ}\APACrefatitle {Gamut: {A} {Design} {Probe} to {Understand} {How} {Data}
  {Scientists} {Understand} {Machine} {Learning} {Models}} {Gamut: {A} {Design}
  {Probe} to {Understand} {How} {Data} {Scientists} {Understand} {Machine}
  {Learning} {Models}}.{\BBCQ}
\newblock
\BIn{} \APACrefbtitle {Proceedings of the 2019 {CHI} {Conference} on {Human}
  {Factors} in {Computing} {Systems}} {Proceedings of the 2019 {CHI}
  {Conference} on {Human} {Factors} in {Computing} {Systems}}\ (\BPGS\ 1--13).
\newblock
\APACaddressPublisher{New York, NY, USA}{Association for Computing Machinery}.
\newblock
\begin{APACrefURL} [{2021-09-19}]\url{https://doi.org/10.1145/3290605.3300809}
  \end{APACrefURL}
\PrintBackRefs{\CurrentBib}

\bibitem [\protect \citeauthoryear {%
Jeff~Larson%
\ \BBA {} Angwin%
}{%
Jeff~Larson%
\ \BBA {} Angwin%
}{%
{\protect \APACyear {2016}}%
}]{%
Larson2016HowWe}
\APACinsertmetastar {%
Larson2016HowWe}%
\begin{APACrefauthors}%
Jeff~Larson, L\BPBI K., Surya~Mattu%
\BCBT {}\ \BBA {} Angwin, J.%
\end{APACrefauthors}%
\unskip\
\newblock
\APACrefYearMonthDay{2016}{May}{}.
\newblock
\APACrefbtitle {How We Analyzed the COMPAS Recidivism Algorithm.} {How we
  analyzed the compas recidivism algorithm.}
\newblock
\begin{APACrefURL}
  \url{https://www.propublica.org/article/how-we-analyzed-the-compas-recidivism-algorithm}
  \end{APACrefURL}
\PrintBackRefs{\CurrentBib}

\bibitem [\protect \citeauthoryear {%
Kamiran%
\ \BBA {} Calders%
}{%
Kamiran%
\ \BBA {} Calders%
}{%
{\protect \APACyear {2009}}%
}]{%
kamiran2009classifying}
\APACinsertmetastar {%
kamiran2009classifying}%
\begin{APACrefauthors}%
Kamiran, F.%
\BCBT {}\ \BBA {} Calders, T.%
\end{APACrefauthors}%
\unskip\
\newblock
\APACrefYearMonthDay{2009}{}{}.
\newblock
{\BBOQ}\APACrefatitle {Classifying without discriminating} {Classifying without
  discriminating}.{\BBCQ}
\newblock
\BIn{} \APACrefbtitle {2009 2nd International Conference on Computer, Control
  and Communication} {2009 2nd international conference on computer, control
  and communication}\ (\BPGS\ 1--6).
\PrintBackRefs{\CurrentBib}

\bibitem [\protect \citeauthoryear {%
Kamishima%
, Akaho%
\BCBL {}\ \BBA {} Sakuma%
}{%
Kamishima%
\ \protect \BOthers {.}}{%
{\protect \APACyear {2011}}%
}]{%
kamishima2011fairness}
\APACinsertmetastar {%
kamishima2011fairness}%
\begin{APACrefauthors}%
Kamishima, T.%
, Akaho, S.%
\BCBL {}\ \BBA {} Sakuma, J.%
\end{APACrefauthors}%
\unskip\
\newblock
\APACrefYearMonthDay{2011}{}{}.
\newblock
{\BBOQ}\APACrefatitle {Fairness-aware learning through regularization approach}
  {Fairness-aware learning through regularization approach}.{\BBCQ}
\newblock
\BIn{} \APACrefbtitle {2011 IEEE 11th International Conference on Data Mining
  Workshops} {2011 ieee 11th international conference on data mining
  workshops}\ (\BPGS\ 643--650).
\PrintBackRefs{\CurrentBib}

\bibitem [\protect \citeauthoryear {%
Kasinidou%
, Kleanthous%
, Barlas%
\BCBL {}\ \BBA {} Otterbacher%
}{%
Kasinidou%
\ \protect \BOthers {.}}{%
{\protect \APACyear {2021}}%
}]{%
kasinidou2021agree}
\APACinsertmetastar {%
kasinidou2021agree}%
\begin{APACrefauthors}%
Kasinidou, M.%
, Kleanthous, S.%
, Barlas, P.%
\BCBL {}\ \BBA {} Otterbacher, J.%
\end{APACrefauthors}%
\unskip\
\newblock
\APACrefYearMonthDay{2021}{}{}.
\newblock
{\BBOQ}\APACrefatitle {I agree with the decision, but they didn't deserve this:
  Future Developers' Perception of Fairness in Algorithmic Decisions} {I agree
  with the decision, but they didn't deserve this: Future developers'
  perception of fairness in algorithmic decisions}.{\BBCQ}
\newblock
\BIn{} \APACrefbtitle {Proceedings of the 2021 ACM Conference on Fairness,
  Accountability, and Transparency} {Proceedings of the 2021 acm conference on
  fairness, accountability, and transparency}\ (\BPGS\ 690--700).
\PrintBackRefs{\CurrentBib}

\bibitem [\protect \citeauthoryear {%
Kearns%
, Neel%
, Roth%
\BCBL {}\ \BBA {} Wu%
}{%
Kearns%
\ \protect \BOthers {.}}{%
{\protect \APACyear {2018}}%
}]{%
kearns2018preventing}
\APACinsertmetastar {%
kearns2018preventing}%
\begin{APACrefauthors}%
Kearns, M.%
, Neel, S.%
, Roth, A.%
\BCBL {}\ \BBA {} Wu, Z\BPBI S.%
\end{APACrefauthors}%
\unskip\
\newblock
\APACrefYearMonthDay{2018}{}{}.
\newblock
{\BBOQ}\APACrefatitle {Preventing fairness gerrymandering: Auditing and
  learning for subgroup fairness} {Preventing fairness gerrymandering: Auditing
  and learning for subgroup fairness}.{\BBCQ}
\newblock
\BIn{} \APACrefbtitle {International Conference on Machine Learning}
  {International conference on machine learning}\ (\BPGS\ 2564--2572).
\PrintBackRefs{\CurrentBib}

\bibitem [\protect \citeauthoryear {%
Kim%
\ \BBA {} Leung%
}{%
Kim%
\ \BBA {} Leung%
}{%
{\protect \APACyear {2007}}%
}]{%
KIM200783FormingandReacting}
\APACinsertmetastar {%
KIM200783FormingandReacting}%
\begin{APACrefauthors}%
Kim, T\BHBI Y.%
\BCBT {}\ \BBA {} Leung, K.%
\end{APACrefauthors}%
\unskip\
\newblock
\APACrefYearMonthDay{2007}{}{}.
\newblock
{\BBOQ}\APACrefatitle {Forming and reacting to overall fairness: A
  cross-cultural comparison} {Forming and reacting to overall fairness: A
  cross-cultural comparison}.{\BBCQ}
\newblock
\APACjournalVolNumPages{Organizational Behavior and Human Decision
  Processes}{104}{1}{83 - 95}.
\newblock
\begin{APACrefURL}
  \url{http://www.sciencedirect.com/science/article/pii/S0749597807000076}
  \end{APACrefURL}
\newblock
\begin{APACrefDOI} \doi{https://doi.org/10.1016/j.obhdp.2007.01.004}
  \end{APACrefDOI}
\PrintBackRefs{\CurrentBib}

\bibitem [\protect \citeauthoryear {%
Kulesza%
, Burnett%
, Wong%
\BCBL {}\ \BBA {} Stumpf%
}{%
Kulesza%
\ \protect \BOthers {.}}{%
{\protect \APACyear {2015}}%
}]{%
kulesza_principles_2015}
\APACinsertmetastar {%
kulesza_principles_2015}%
\begin{APACrefauthors}%
Kulesza, T.%
, Burnett, M.%
, Wong, W\BHBI K.%
\BCBL {}\ \BBA {} Stumpf, S.%
\end{APACrefauthors}%
\unskip\
\newblock
\APACrefYearMonthDay{2015}{}{}.
\newblock
{\BBOQ}\APACrefatitle {Principles of {Explanatory} {Debugging} to {Personalize}
  {Interactive} {Machine} {Learning}} {Principles of {Explanatory} {Debugging}
  to {Personalize} {Interactive} {Machine} {Learning}}.{\BBCQ}
\newblock
\BIn{} \APACrefbtitle {Proceedings of the 20th {International} {Conference} on
  {Intelligent} {User} {Interfaces}} {Proceedings of the 20th {International}
  {Conference} on {Intelligent} {User} {Interfaces}}\ (\BPGS\ 126--137).
\newblock
\APACaddressPublisher{New York, NY, USA}{ACM}.
\newblock
\begin{APACrefURL}
  [{2015-12-14}]\url{http://doi.acm.org/10.1145/2678025.2701399}
  \end{APACrefURL}
\newblock
\begin{APACrefDOI} \doi{10.1145/2678025.2701399} \end{APACrefDOI}
\PrintBackRefs{\CurrentBib}

\bibitem [\protect \citeauthoryear {%
Kulesza%
, Stumpf%
, Burnett%
\BCBL {}\ \BBA {} Kwan%
}{%
Kulesza%
\ \protect \BOthers {.}}{%
{\protect \APACyear {2012}}%
}]{%
Kulesza2012TellMeMore}
\APACinsertmetastar {%
Kulesza2012TellMeMore}%
\begin{APACrefauthors}%
Kulesza, T.%
, Stumpf, S.%
, Burnett, M.%
\BCBL {}\ \BBA {} Kwan, I.%
\end{APACrefauthors}%
\unskip\
\newblock
\APACrefYearMonthDay{2012}{}{}.
\newblock
{\BBOQ}\APACrefatitle {Tell Me More? The Effects of Mental Model Soundness on
  Personalizing an Intelligent Agent} {Tell me more? the effects of mental
  model soundness on personalizing an intelligent agent}.{\BBCQ}
\newblock
\BIn{} \APACrefbtitle {Proceedings of the SIGCHI Conference on Human Factors in
  Computing Systems} {Proceedings of the sigchi conference on human factors in
  computing systems}\ (\BPG~1–10).
\newblock
\APACaddressPublisher{New York, NY, USA}{Association for Computing Machinery}.
\newblock
\begin{APACrefURL} \url{https://doi.org/10.1145/2207676.2207678}
  \end{APACrefURL}
\newblock
\begin{APACrefDOI} \doi{10.1145/2207676.2207678} \end{APACrefDOI}
\PrintBackRefs{\CurrentBib}

\bibitem [\protect \citeauthoryear {%
Kusner%
, Loftus%
, Russell%
\BCBL {}\ \BBA {} Silva%
}{%
Kusner%
\ \protect \BOthers {.}}{%
{\protect \APACyear {2017}}%
}]{%
kusner2017counterfactual}
\APACinsertmetastar {%
kusner2017counterfactual}%
\begin{APACrefauthors}%
Kusner, M\BPBI J.%
, Loftus, J.%
, Russell, C.%
\BCBL {}\ \BBA {} Silva, R.%
\end{APACrefauthors}%
\unskip\
\newblock
\APACrefYearMonthDay{2017}{}{}.
\newblock
{\BBOQ}\APACrefatitle {Counterfactual fairness} {Counterfactual
  fairness}.{\BBCQ}
\newblock
\BIn{} \APACrefbtitle {Advances in neural information processing systems}
  {Advances in neural information processing systems}\ (\BPGS\ 4066--4076).
\PrintBackRefs{\CurrentBib}

\bibitem [\protect \citeauthoryear {%
Lee%
, Jain%
, Cha%
, Ojha%
\BCBL {}\ \BBA {} Kusbit%
}{%
Lee%
\ \protect \BOthers {.}}{%
{\protect \APACyear {2019}}%
}]{%
lee2019procedural}
\APACinsertmetastar {%
lee2019procedural}%
\begin{APACrefauthors}%
Lee, M\BPBI K.%
, Jain, A.%
, Cha, H\BPBI J.%
, Ojha, S.%
\BCBL {}\ \BBA {} Kusbit, D.%
\end{APACrefauthors}%
\unskip\
\newblock
\APACrefYearMonthDay{2019}{}{}.
\newblock
{\BBOQ}\APACrefatitle {Procedural justice in algorithmic fairness: Leveraging
  transparency and outcome control for fair algorithmic mediation} {Procedural
  justice in algorithmic fairness: Leveraging transparency and outcome control
  for fair algorithmic mediation}.{\BBCQ}
\newblock
\APACjournalVolNumPages{Proceedings of the ACM on Human-Computer
  Interaction}{3}{CSCW}{1--26}.
\PrintBackRefs{\CurrentBib}

\bibitem [\protect \citeauthoryear {%
Lee%
, Kim%
\BCBL {}\ \BBA {} Lizarondo%
}{%
Lee%
\ \protect \BOthers {.}}{%
{\protect \APACyear {2017}}%
}]{%
Lee2017AHuman-Centered}
\APACinsertmetastar {%
Lee2017AHuman-Centered}%
\begin{APACrefauthors}%
Lee, M\BPBI K.%
, Kim, J\BPBI T.%
\BCBL {}\ \BBA {} Lizarondo, L.%
\end{APACrefauthors}%
\unskip\
\newblock
\APACrefYearMonthDay{2017}{}{}.
\newblock
{\BBOQ}\APACrefatitle {A Human-Centered Approach to Algorithmic Services:
  Considerations for Fair and Motivating Smart Community Service Management
  That Allocates Donations to Non-Profit Organizations} {A human-centered
  approach to algorithmic services: Considerations for fair and motivating
  smart community service management that allocates donations to non-profit
  organizations}.{\BBCQ}
\newblock
\BIn{} \APACrefbtitle {Proceedings of the 2017 CHI Conference on Human Factors
  in Computing Systems} {Proceedings of the 2017 chi conference on human
  factors in computing systems}\ (\BPG~3365–3376).
\newblock
\APACaddressPublisher{New York, NY, USA}{Association for Computing Machinery}.
\newblock
\begin{APACrefURL} \url{https://doi.org/10.1145/3025453.3025884}
  \end{APACrefURL}
\newblock
\begin{APACrefDOI} \doi{10.1145/3025453.3025884} \end{APACrefDOI}
\PrintBackRefs{\CurrentBib}

\bibitem [\protect \citeauthoryear {%
Lim%
\ \BBA {} Dey%
}{%
Lim%
\ \BBA {} Dey%
}{%
{\protect \APACyear {2009}}%
}]{%
lim_assessing_2009}
\APACinsertmetastar {%
lim_assessing_2009}%
\begin{APACrefauthors}%
Lim, B\BPBI Y.%
\BCBT {}\ \BBA {} Dey, A\BPBI K.%
\end{APACrefauthors}%
\unskip\
\newblock
\APACrefYearMonthDay{2009}{}{}.
\newblock
{\BBOQ}\APACrefatitle {Assessing {Demand} for {Intelligibility} in
  {Context}-aware {Applications}} {Assessing {Demand} for {Intelligibility} in
  {Context}-aware {Applications}}.{\BBCQ}
\newblock
\BIn{} \APACrefbtitle {Proceedings of the 11th {International} {Conference} on
  {Ubiquitous} {Computing}} {Proceedings of the 11th {International}
  {Conference} on {Ubiquitous} {Computing}}\ (\BPGS\ 195--204).
\newblock
\APACaddressPublisher{New York, NY, USA}{ACM}.
\newblock
\begin{APACrefURL} \url{http://doi.acm.org/10.1145/1620545.1620576}
  \end{APACrefURL}
\newblock
\begin{APACrefDOI} \doi{10.1145/1620545.1620576} \end{APACrefDOI}
\PrintBackRefs{\CurrentBib}

\bibitem [\protect \citeauthoryear {%
Lundberg%
\ \BBA {} Lee%
}{%
Lundberg%
\ \BBA {} Lee%
}{%
{\protect \APACyear {2017}}%
}]{%
Lundberg2017UnifiedApproach}
\APACinsertmetastar {%
Lundberg2017UnifiedApproach}%
\begin{APACrefauthors}%
Lundberg, S\BPBI M.%
\BCBT {}\ \BBA {} Lee, S\BHBI I.%
\end{APACrefauthors}%
\unskip\
\newblock
\APACrefYearMonthDay{2017}{}{}.
\newblock
{\BBOQ}\APACrefatitle {A Unified Approach to Interpreting Model Predictions} {A
  unified approach to interpreting model predictions}.{\BBCQ}
\newblock
\BIn{} I.~Guyon\ \BOthers {.}\ (\BEDS), \APACrefbtitle {Advances in Neural
  Information Processing Systems} {Advances in neural information processing
  systems}\ (\BVOL~30).
\newblock
\APACaddressPublisher{}{Curran Associates, Inc.}
\newblock
\begin{APACrefURL}
  \url{https://proceedings.neurips.cc/paper/2017/file/8a20a8621978632d76c43dfd28b67767-Paper.pdf}
  \end{APACrefURL}
\PrintBackRefs{\CurrentBib}

\bibitem [\protect \citeauthoryear {%
Mallari%
\ \protect \BOthers {.}}{%
Mallari%
\ \protect \BOthers {.}}{%
{\protect \APACyear {2020}}%
}]{%
Mallari2020DoILook}
\APACinsertmetastar {%
Mallari2020DoILook}%
\begin{APACrefauthors}%
Mallari, K.%
, Inkpen, K.%
, Johns, P.%
, Tan, S.%
, Ramesh, D.%
\BCBL {}\ \BBA {} Kamar, E.%
\end{APACrefauthors}%
\unskip\
\newblock
\APACrefYearMonthDay{2020}{}{}.
\newblock
{\BBOQ}\APACrefatitle {Do I Look Like a Criminal? Examining How Race
  Presentation Impacts Human Judgement of Recidivism} {Do i look like a
  criminal? examining how race presentation impacts human judgement of
  recidivism}.{\BBCQ}
\newblock
\BIn{} \APACrefbtitle {Proceedings of the 2020 CHI Conference on Human Factors
  in Computing Systems} {Proceedings of the 2020 chi conference on human
  factors in computing systems}\ (\BPG~1–13).
\newblock
\APACaddressPublisher{New York, NY, USA}{Association for Computing Machinery}.
\newblock
\begin{APACrefURL} \url{https://doi.org/10.1145/3313831.3376257}
  \end{APACrefURL}
\newblock
\begin{APACrefDOI} \doi{10.1145/3313831.3376257} \end{APACrefDOI}
\PrintBackRefs{\CurrentBib}

\bibitem [\protect \citeauthoryear {%
Mattila%
\ \BBA {} Choi%
}{%
Mattila%
\ \BBA {} Choi%
}{%
{\protect \APACyear {2006}}%
}]{%
MATTILA2006CrossCulturalComparison}
\APACinsertmetastar {%
MATTILA2006CrossCulturalComparison}%
\begin{APACrefauthors}%
Mattila, A\BPBI S.%
\BCBT {}\ \BBA {} Choi, S.%
\end{APACrefauthors}%
\unskip\
\newblock
\APACrefYearMonthDay{2006}{}{}.
\newblock
{\BBOQ}\APACrefatitle {A cross-cultural comparison of perceived fairness and
  satisfaction in the context of hotel room pricing} {A cross-cultural
  comparison of perceived fairness and satisfaction in the context of hotel
  room pricing}.{\BBCQ}
\newblock
\APACjournalVolNumPages{International Journal of Hospitality
  Management}{25}{1}{146 - 153}.
\newblock
\begin{APACrefURL}
  \url{http://www.sciencedirect.com/science/article/pii/S0278431904001240}
  \end{APACrefURL}
\newblock
\begin{APACrefDOI} \doi{https://doi.org/10.1016/j.ijhm.2004.12.003}
  \end{APACrefDOI}
\PrintBackRefs{\CurrentBib}

\bibitem [\protect \citeauthoryear {%
Mitchell%
, Potash%
, Barocas%
, D'Amour%
\BCBL {}\ \BBA {} Lum%
}{%
Mitchell%
\ \protect \BOthers {.}}{%
{\protect \APACyear {2018}}%
}]{%
mitchell2018prediction}
\APACinsertmetastar {%
mitchell2018prediction}%
\begin{APACrefauthors}%
Mitchell, S.%
, Potash, E.%
, Barocas, S.%
, D'Amour, A.%
\BCBL {}\ \BBA {} Lum, K.%
\end{APACrefauthors}%
\unskip\
\newblock
\APACrefYearMonthDay{2018}{}{}.
\newblock
{\BBOQ}\APACrefatitle {Prediction-based decisions and fairness: A catalogue of
  choices, assumptions, and definitions} {Prediction-based decisions and
  fairness: A catalogue of choices, assumptions, and definitions}.{\BBCQ}
\newblock
\APACjournalVolNumPages{arXiv preprint arXiv:1811.07867}{}{}{}.
\PrintBackRefs{\CurrentBib}

\bibitem [\protect \citeauthoryear {%
Narayanan%
}{%
Narayanan%
}{%
{\protect \APACyear {2018}}%
}]{%
narayanan2018translation}
\APACinsertmetastar {%
narayanan2018translation}%
\begin{APACrefauthors}%
Narayanan, A.%
\end{APACrefauthors}%
\unskip\
\newblock
\APACrefYearMonthDay{2018}{}{}.
\newblock
{\BBOQ}\APACrefatitle {Translation tutorial: 21 fairness definitions and their
  politics} {Translation tutorial: 21 fairness definitions and their
  politics}.{\BBCQ}
\newblock
\BIn{} \APACrefbtitle {Proc. Conf. Fairness Accountability Transp., New York,
  USA} {Proc. conf. fairness accountability transp., new york, usa}\ (\BVOL~2,
  \BPGS\ 6--2).
\PrintBackRefs{\CurrentBib}

\bibitem [\protect \citeauthoryear {%
Olteanu%
, Castillo%
, Diaz%
\BCBL {}\ \BBA {} Kıcıman%
}{%
Olteanu%
\ \protect \BOthers {.}}{%
{\protect \APACyear {2019}}%
}]{%
Olteanu2019SocialData}
\APACinsertmetastar {%
Olteanu2019SocialData}%
\begin{APACrefauthors}%
Olteanu, A.%
, Castillo, C.%
, Diaz, F.%
\BCBL {}\ \BBA {} Kıcıman, E.%
\end{APACrefauthors}%
\unskip\
\newblock
\APACrefYearMonthDay{2019}{}{}.
\newblock
{\BBOQ}\APACrefatitle {Social Data: Biases, Methodological Pitfalls, and
  Ethical Boundaries} {Social data: Biases, methodological pitfalls, and
  ethical boundaries}.{\BBCQ}
\newblock
\APACjournalVolNumPages{Frontiers in Big Data}{2}{}{13}.
\newblock
\begin{APACrefURL}
  \url{https://www.frontiersin.org/article/10.3389/fdata.2019.00013}
  \end{APACrefURL}
\newblock
\begin{APACrefDOI} \doi{10.3389/fdata.2019.00013} \end{APACrefDOI}
\PrintBackRefs{\CurrentBib}

\bibitem [\protect \citeauthoryear {%
Rawls%
}{%
Rawls%
}{%
{\protect \APACyear {1958}}%
}]{%
rawls1958justice}
\APACinsertmetastar {%
rawls1958justice}%
\begin{APACrefauthors}%
Rawls, J.%
\end{APACrefauthors}%
\unskip\
\newblock
\APACrefYearMonthDay{1958}{}{}.
\newblock
{\BBOQ}\APACrefatitle {Justice as fairness} {Justice as fairness}.{\BBCQ}
\newblock
\APACjournalVolNumPages{The philosophical review}{67}{2}{164--194}.
\PrintBackRefs{\CurrentBib}

\bibitem [\protect \citeauthoryear {%
Ribeiro%
, Singh%
\BCBL {}\ \BBA {} Guestrin%
}{%
Ribeiro%
\ \protect \BOthers {.}}{%
{\protect \APACyear {2016}}%
}]{%
Ribeiro2016WhyshouldI}
\APACinsertmetastar {%
Ribeiro2016WhyshouldI}%
\begin{APACrefauthors}%
Ribeiro, M\BPBI T.%
, Singh, S.%
\BCBL {}\ \BBA {} Guestrin, C.%
\end{APACrefauthors}%
\unskip\
\newblock
\APACrefYearMonthDay{2016}{}{}.
\newblock
{\BBOQ}\APACrefatitle {"Why Should I Trust You?": Explaining the Predictions of
  Any Classifier} {"why should i trust you?": Explaining the predictions of any
  classifier}.{\BBCQ}
\newblock
\BIn{} \APACrefbtitle {Proceedings of the 22nd ACM SIGKDD International
  Conference on Knowledge Discovery and Data Mining} {Proceedings of the 22nd
  acm sigkdd international conference on knowledge discovery and data mining}\
  (\BPG~1135–1144).
\newblock
\APACaddressPublisher{New York, NY, USA}{Association for Computing Machinery}.
\newblock
\begin{APACrefURL} \url{https://doi.org/10.1145/2939672.2939778}
  \end{APACrefURL}
\newblock
\begin{APACrefDOI} \doi{10.1145/2939672.2939778} \end{APACrefDOI}
\PrintBackRefs{\CurrentBib}

\bibitem [\protect \citeauthoryear {%
Saxena%
\ \protect \BOthers {.}}{%
Saxena%
\ \protect \BOthers {.}}{%
{\protect \APACyear {2020}}%
}]{%
saxena2020fairness}
\APACinsertmetastar {%
saxena2020fairness}%
\begin{APACrefauthors}%
Saxena, N\BPBI A.%
, Huang, K.%
, DeFilippis, E.%
, Radanovic, G.%
, Parkes, D\BPBI C.%
\BCBL {}\ \BBA {} Liu, Y.%
\end{APACrefauthors}%
\unskip\
\newblock
\APACrefYearMonthDay{2020}{}{}.
\newblock
{\BBOQ}\APACrefatitle {How do fairness definitions fare? Testing public
  attitudes towards three algorithmic definitions of fairness in loan
  allocations} {How do fairness definitions fare? testing public attitudes
  towards three algorithmic definitions of fairness in loan
  allocations}.{\BBCQ}
\newblock
\APACjournalVolNumPages{Artificial Intelligence}{283}{}{103238}.
\PrintBackRefs{\CurrentBib}

\bibitem [\protect \citeauthoryear {%
Shneiderman%
}{%
Shneiderman%
}{%
{\protect \APACyear {2020}}%
}]{%
shneiderman_human-centered_2020}
\APACinsertmetastar {%
shneiderman_human-centered_2020}%
\begin{APACrefauthors}%
Shneiderman, B.%
\end{APACrefauthors}%
\unskip\
\newblock
\APACrefYearMonthDay{2020}{{\APACmonth{04}}}{}.
\newblock
{\BBOQ}\APACrefatitle {Human-{Centered} {Artificial} {Intelligence}:
  {Reliable}, {Safe} \& {Trustworthy}} {Human-{Centered} {Artificial}
  {Intelligence}: {Reliable}, {Safe} \& {Trustworthy}}.{\BBCQ}
\newblock
\APACjournalVolNumPages{International Journal of Human–Computer
  Interaction}{36}{6}{495--504}.
\newblock
\begin{APACrefURL}
  [{2021-08-18}]\url{https://doi.org/10.1080/10447318.2020.1741118}
  \end{APACrefURL}
\newblock
\APACrefnote{Publisher: Taylor \& Francis \_eprint:
  https://doi.org/10.1080/10447318.2020.1741118}
\newblock
\begin{APACrefDOI} \doi{10.1080/10447318.2020.1741118} \end{APACrefDOI}
\PrintBackRefs{\CurrentBib}

\bibitem [\protect \citeauthoryear {%
Shneiderman%
}{%
Shneiderman%
}{%
{\protect \APACyear {2021}}%
}]{%
shneiderman_responsible_2021}
\APACinsertmetastar {%
shneiderman_responsible_2021}%
\begin{APACrefauthors}%
Shneiderman, B.%
\end{APACrefauthors}%
\unskip\
\newblock
\APACrefYearMonthDay{2021}{{\APACmonth{08}}}{}.
\newblock
{\BBOQ}\APACrefatitle {Responsible {AI}: {Bridging} {From} {Ethics} to
  {Practice}} {Responsible {AI}: {Bridging} {From} {Ethics} to
  {Practice}}.{\BBCQ}
\newblock
\APACjournalVolNumPages{Communications of the ACM}{64}{8}{32--35}.
\newblock
\begin{APACrefURL}
  [{2021-08-18}]\url{https://cacm.acm.org/magazines/2021/8/254306-responsible-ai/fulltext?mobile=false}
  \end{APACrefURL}
\PrintBackRefs{\CurrentBib}

\bibitem [\protect \citeauthoryear {%
Veale%
, Van~Kleek%
\BCBL {}\ \BBA {} Binns%
}{%
Veale%
\ \protect \BOthers {.}}{%
{\protect \APACyear {2018}}%
}]{%
Veala2018Fairness}
\APACinsertmetastar {%
Veala2018Fairness}%
\begin{APACrefauthors}%
Veale, M.%
, Van~Kleek, M.%
\BCBL {}\ \BBA {} Binns, R.%
\end{APACrefauthors}%
\unskip\
\newblock
\APACrefYearMonthDay{2018}{}{}.
\newblock
{\BBOQ}\APACrefatitle {Fairness and Accountability Design Needs for Algorithmic
  Support in High-Stakes Public Sector Decision-Making} {Fairness and
  accountability design needs for algorithmic support in high-stakes public
  sector decision-making}.{\BBCQ}
\newblock
\BIn{} \APACrefbtitle {Proceedings of the 2018 CHI Conference on Human Factors
  in Computing Systems} {Proceedings of the 2018 chi conference on human
  factors in computing systems}\ (\BPG~1–14).
\newblock
\APACaddressPublisher{New York, NY, USA}{Association for Computing Machinery}.
\newblock
\begin{APACrefURL} \url{https://doi.org/10.1145/3173574.3174014}
  \end{APACrefURL}
\newblock
\begin{APACrefDOI} \doi{10.1145/3173574.3174014} \end{APACrefDOI}
\PrintBackRefs{\CurrentBib}

\bibitem [\protect \citeauthoryear {%
Verma%
\ \BBA {} Rubin%
}{%
Verma%
\ \BBA {} Rubin%
}{%
{\protect \APACyear {2018}}%
}]{%
verma2018fairness}
\APACinsertmetastar {%
verma2018fairness}%
\begin{APACrefauthors}%
Verma, S.%
\BCBT {}\ \BBA {} Rubin, J.%
\end{APACrefauthors}%
\unskip\
\newblock
\APACrefYearMonthDay{2018}{}{}.
\newblock
{\BBOQ}\APACrefatitle {Fairness definitions explained} {Fairness definitions
  explained}.{\BBCQ}
\newblock
\BIn{} \APACrefbtitle {2018 IEEE/ACM International Workshop on Software
  Fairness (FairWare)} {2018 ieee/acm international workshop on software
  fairness (fairware)}\ (\BPGS\ 1--7).
\PrintBackRefs{\CurrentBib}

\bibitem [\protect \citeauthoryear {%
Wang%
, Harper%
\BCBL {}\ \BBA {} Zhu%
}{%
Wang%
\ \protect \BOthers {.}}{%
{\protect \APACyear {2020}}%
}]{%
Wang2020Factors}
\APACinsertmetastar {%
Wang2020Factors}%
\begin{APACrefauthors}%
Wang, R.%
, Harper, F\BPBI M.%
\BCBL {}\ \BBA {} Zhu, H.%
\end{APACrefauthors}%
\unskip\
\newblock
\APACrefYearMonthDay{2020}{}{}.
\newblock
{\BBOQ}\APACrefatitle {Factors Influencing Perceived Fairness in Algorithmic
  Decision-Making: Algorithm Outcomes, Development Procedures, and Individual
  Differences} {Factors influencing perceived fairness in algorithmic
  decision-making: Algorithm outcomes, development procedures, and individual
  differences}.{\BBCQ}
\newblock
\BIn{} \APACrefbtitle {Proceedings of the 2020 CHI Conference on Human Factors
  in Computing Systems} {Proceedings of the 2020 chi conference on human
  factors in computing systems}\ (\BPG~1–14).
\newblock
\APACaddressPublisher{New York, NY, USA}{Association for Computing Machinery}.
\newblock
\begin{APACrefURL} \url{https://doi.org/10.1145/3313831.3376813}
  \end{APACrefURL}
\newblock
\begin{APACrefDOI} \doi{10.1145/3313831.3376813} \end{APACrefDOI}
\PrintBackRefs{\CurrentBib}

\bibitem [\protect \citeauthoryear {%
{Wexler}%
\ \protect \BOthers {.}}{%
{Wexler}%
\ \protect \BOthers {.}}{%
{\protect \APACyear {2020}}%
}]{%
Wexler2020What-IfTool}
\APACinsertmetastar {%
Wexler2020What-IfTool}%
\begin{APACrefauthors}%
{Wexler}, J.%
, {Pushkarna}, M.%
, {Bolukbasi}, T.%
, {Wattenberg}, M.%
, {Viégas}, F.%
\BCBL {}\ \BBA {} {Wilson}, J.%
\end{APACrefauthors}%
\unskip\
\newblock
\APACrefYearMonthDay{2020}{}{}.
\newblock
{\BBOQ}\APACrefatitle {The What-If Tool: Interactive Probing of Machine
  Learning Models} {The what-if tool: Interactive probing of machine learning
  models}.{\BBCQ}
\newblock
\APACjournalVolNumPages{IEEE Transactions on Visualization and Computer
  Graphics}{26}{1}{56-65}.
\newblock
\begin{APACrefDOI} \doi{10.1109/TVCG.2019.2934619} \end{APACrefDOI}
\PrintBackRefs{\CurrentBib}

\bibitem [\protect \citeauthoryear {%
Woodruff%
, Fox%
, Rousso-Schindler%
\BCBL {}\ \BBA {} Warshaw%
}{%
Woodruff%
\ \protect \BOthers {.}}{%
{\protect \APACyear {2018}}%
}]{%
woodruff2018qualitative}
\APACinsertmetastar {%
woodruff2018qualitative}%
\begin{APACrefauthors}%
Woodruff, A.%
, Fox, S\BPBI E.%
, Rousso-Schindler, S.%
\BCBL {}\ \BBA {} Warshaw, J.%
\end{APACrefauthors}%
\unskip\
\newblock
\APACrefYearMonthDay{2018}{}{}.
\newblock
{\BBOQ}\APACrefatitle {A qualitative exploration of perceptions of algorithmic
  fairness} {A qualitative exploration of perceptions of algorithmic
  fairness}.{\BBCQ}
\newblock
\BIn{} \APACrefbtitle {Proceedings of the 2018 CHI conference on human factors
  in computing systems} {Proceedings of the 2018 chi conference on human
  factors in computing systems}\ (\BPGS\ 1--14).
\PrintBackRefs{\CurrentBib}

\bibitem [\protect \citeauthoryear {%
Yan%
, Gu%
, Lin%
\BCBL {}\ \BBA {} Rzeszotarski%
}{%
Yan%
\ \protect \BOthers {.}}{%
{\protect \APACyear {2020}}%
}]{%
Yan2020SILVA}
\APACinsertmetastar {%
Yan2020SILVA}%
\begin{APACrefauthors}%
Yan, J\BPBI N.%
, Gu, Z.%
, Lin, H.%
\BCBL {}\ \BBA {} Rzeszotarski, J\BPBI M.%
\end{APACrefauthors}%
\unskip\
\newblock
\APACrefYearMonthDay{2020}{}{}.
\newblock
{\BBOQ}\APACrefatitle {Silva: Interactively Assessing Machine Learning Fairness
  Using Causality} {Silva: Interactively assessing machine learning fairness
  using causality}.{\BBCQ}
\newblock
\BIn{} \APACrefbtitle {Proceedings of the 2020 CHI Conference on Human Factors
  in Computing Systems} {Proceedings of the 2020 chi conference on human
  factors in computing systems}\ (\BPG~1–13).
\newblock
\APACaddressPublisher{New York, NY, USA}{Association for Computing Machinery}.
\newblock
\begin{APACrefURL} \url{https://doi.org/10.1145/3313831.3376447}
  \end{APACrefURL}
\newblock
\begin{APACrefDOI} \doi{10.1145/3313831.3376447} \end{APACrefDOI}
\PrintBackRefs{\CurrentBib}

\bibitem [\protect \citeauthoryear {%
Yang%
, Cisse%
\BCBL {}\ \BBA {} Koyejo%
}{%
Yang%
\ \protect \BOthers {.}}{%
{\protect \APACyear {2020}}%
}]{%
yang2020fairness}
\APACinsertmetastar {%
yang2020fairness}%
\begin{APACrefauthors}%
Yang, F.%
, Cisse, M.%
\BCBL {}\ \BBA {} Koyejo, O\BPBI O.%
\end{APACrefauthors}%
\unskip\
\newblock
\APACrefYearMonthDay{2020}{}{}.
\newblock
{\BBOQ}\APACrefatitle {Fairness with Overlapping Groups; a Probabilistic
  Perspective} {Fairness with overlapping groups; a probabilistic
  perspective}.{\BBCQ}
\newblock
\APACjournalVolNumPages{Advances in Neural Information Processing
  Systems}{33}{}{}.
\PrintBackRefs{\CurrentBib}

\bibitem [\protect \citeauthoryear {%
Zemel%
, Wu%
, Swersky%
, Pitassi%
\BCBL {}\ \BBA {} Dwork%
}{%
Zemel%
\ \protect \BOthers {.}}{%
{\protect \APACyear {2013}}%
}]{%
zemel2013learning}
\APACinsertmetastar {%
zemel2013learning}%
\begin{APACrefauthors}%
Zemel, R.%
, Wu, Y.%
, Swersky, K.%
, Pitassi, T.%
\BCBL {}\ \BBA {} Dwork, C.%
\end{APACrefauthors}%
\unskip\
\newblock
\APACrefYearMonthDay{2013}{}{}.
\newblock
{\BBOQ}\APACrefatitle {Learning fair representations} {Learning fair
  representations}.{\BBCQ}
\newblock
\BIn{} \APACrefbtitle {International Conference on Machine Learning}
  {International conference on machine learning}\ (\BPGS\ 325--333).
\PrintBackRefs{\CurrentBib}

\bibitem [\protect \citeauthoryear {%
Zhang%
, Lemoine%
\BCBL {}\ \BBA {} Mitchell%
}{%
Zhang%
\ \protect \BOthers {.}}{%
{\protect \APACyear {2018}}%
}]{%
zhang2018mitigating}
\APACinsertmetastar {%
zhang2018mitigating}%
\begin{APACrefauthors}%
Zhang, B\BPBI H.%
, Lemoine, B.%
\BCBL {}\ \BBA {} Mitchell, M.%
\end{APACrefauthors}%
\unskip\
\newblock
\APACrefYearMonthDay{2018}{}{}.
\newblock
{\BBOQ}\APACrefatitle {Mitigating unwanted biases with adversarial learning}
  {Mitigating unwanted biases with adversarial learning}.{\BBCQ}
\newblock
\BIn{} \APACrefbtitle {Proceedings of the 2018 AAAI/ACM Conference on AI,
  Ethics, and Society} {Proceedings of the 2018 aaai/acm conference on ai,
  ethics, and society}\ (\BPGS\ 335--340).
\PrintBackRefs{\CurrentBib}

\bibitem [\protect \citeauthoryear {%
Zheng%
, Aragam%
, Ravikumar%
\BCBL {}\ \BBA {} Xing%
}{%
Zheng%
\ \protect \BOthers {.}}{%
{\protect \APACyear {2018}}%
}]{%
zheng2018dags}
\APACinsertmetastar {%
zheng2018dags}%
\begin{APACrefauthors}%
Zheng, X.%
, Aragam, B.%
, Ravikumar, P\BPBI K.%
\BCBL {}\ \BBA {} Xing, E\BPBI P.%
\end{APACrefauthors}%
\unskip\
\newblock
\APACrefYearMonthDay{2018}{}{}.
\newblock
{\BBOQ}\APACrefatitle {DAGs with NO TEARS: Continuous optimization for
  structure learning} {Dags with no tears: Continuous optimization for
  structure learning}.{\BBCQ}
\newblock
\BIn{} \APACrefbtitle {Advances in Neural Information Processing Systems}
  {Advances in neural information processing systems}\ (\BPGS\ 9472--9483).
\PrintBackRefs{\CurrentBib}

\end{thebibliography}

\newpage

\begin{table}[h!]
\caption{A comparison of UI functionality among existing tools described in section 2.2. }
\centering
\ra{1.2}
\resizebox{\textwidth}{!}{
\begin{tabular}{@{}lccccccc@{}}
\toprule
Functionality & \begin{tabular}{c}What-if Tool \\ (\citeauthor{Wexler2020What-IfTool}, \\ \citeyear{Wexler2020What-IfTool})\end{tabular} & \begin{tabular}{c}FairVis \\ (\citeauthor{Cabrera2019Fairvis},\\ \citeyear{Cabrera2019Fairvis})\end{tabular} & \begin{tabular}{c}FairSight \\ 
(\citeauthor{Ahn2020FairSight}, \\ \citeyear{Ahn2020FairSight}) \end{tabular} & \begin{tabular}{c}Silva 
\\ (\citeauthor{Yan2020SILVA}, \\ \citeyear{Yan2020SILVA})\end{tabular} & \begin{tabular}{c}
    Fairness \\
    Elicitation Tool\\ 
    (\citeauthor{cheng_soliciting_2021}, \\
    \citeyear{cheng_soliciting_2021})\end{tabular} \\
\midrule
\begin{tabular}{l}Show performance \\ metrics\end{tabular} &  \checkmark &  \checkmark &  \checkmark &  & \\
\rowcolor{Gray}\begin{tabular}{l}Show statistical \\ fairness  metrics\end{tabular}       &  \checkmark &  &  \checkmark &  \checkmark &  \checkmark\\
\begin{tabular}{l}Select sensitive/ \\ protected features\end{tabular}     &  &  &  \checkmark &  \checkmark & \\
\rowcolor{Gray}\begin{tabular}{l}Show data in groups\end{tabular}                     & \checkmark & \checkmark & \checkmark & \checkmark & \checkmark\\
\begin{tabular}{l}Show sub-groups\end{tabular}                         & \checkmark &  &  &  & \checkmark\\
\rowcolor{Gray}\begin{tabular}{l}Show counterfactuals \\ to data points\end{tabular}     & \checkmark & \checkmark &  &  & \\
\begin{tabular}{l}Show decision \\ boundaries\end{tabular}                & \checkmark &  &  &  & \\
\rowcolor{Gray}\begin{tabular}{l}Provide descriptive \\ stats for features\end{tabular}  & \checkmark &  & \checkmark &  & \\
\begin{tabular}{l}Show feature \\ contributions\end{tabular}              &  & \checkmark & \checkmark &  & \checkmark\\
\rowcolor{Gray}\begin{tabular}{l}Show causal graph\end{tabular}                       &  &  &  & \checkmark & \\
\begin{tabular}{l}Show data similarity\end{tabular}                    &  &  &  \checkmark &  & \checkmark\\
\bottomrule
\end{tabular}
}
\label{tab:UIComarison}
\end{table}

\newpage

\begin{table}[h!]
\caption{Overview of workshop participants' details.}
\centering
\ra{1.2}
\begin{tabular}{@{}lccccr@{}}
\toprule
ID	& Gender & Age &  Education & Role \\
 \midrule
WL01 &	Male &	36 &		Bachelor &	Loan Officer \\
WL02 &	Male &	33	& 	Master	& Loan Officer \\
WL03 &	Male &	39	& 	Master &	Loan Officer \\
WL04 &	Male &	42 &	 Master &	Loan Officer \\
WL05 &	Male &	34 &	 Master &	Loan Officer \\
WL06 &	Female &	35	& 	Master &	Loan Officer \\
WD01 &	Female &	28	 &	Master &Data Scientist \\
WD02 &	Male &	38 &	 Doctorate	& Data Scientist \\
WD03 &	Female &	27 &	 Master &	Data Scientist \\
WD04 &	Male &	28	&  Master &	Data Scientist \\
WD05 &	Male &	28 &		Master &	Data Scientist \\
WD06 &	Female &	29 &		Master &	Data Scientist \\
\bottomrule
\end{tabular}
\label{table:participants}
\end{table}

\newpage

\begin{table}[h!]
\caption{Requirements arising from design workshops. In the Stakeholder row, L stands for the requirements from loan officers, D from data scientists, and B from both stakeholders.}
\centering
\ra{1.2}
\resizebox{\textwidth}{!}{
\begin{tabular}{@{}lllc@{}}
\toprule
\textbf{Area} & \textbf{Use} & \textbf{Requirement} & \textbf{Stakeholder}\\
\hline
1. Attribute  & \multirow{8}{*}{Informational}&  \makecell[l]{1.1 attributes, number of records and attribute value distributions} & B\\
overviews&& \makecell[l]{1.2 amount of missing data}  & B\\
&& \makecell[l]{1.3 fairness metrics for model and individual protected attributes} & B \\
&& \makecell[l]{1.4 target distribution} & B \\  
&& \makecell[l]{1.5 protected attributes} & D  \\
&& \makecell[l]{1.6 explanation how attribute values  are calculated or derived}  & D \\
&& \makecell[l]{1.7 Identify if the data is subjective/ objective} & D  \\
&& \makecell[l]{1.8 Identify where the data has come from (applicant, bank, third \\ party) / attribute  provenance} & D  \\\hline

2. Investigate& \multirow{4}{*}{Informational} & \makecell[l]{2.1 distribution of protected attributes with other attributes} & B\\
Relationships&& \makecell[l]{2.2 distribution of user-selected attribute values (e.g. credit risk \\ ratings) and target values} & B\\ 
between attributes&& \makecell[l]{2.3 distribution of two user-selected attributes’ values} & B\\
&& \makecell[l]{2.4 Credit risk rating traffic light system} & D \\ \cline{2-4}
& \multirow{6}{*}{Functional} & \makecell[l]{2.5 Support data transformations (e.g. categorical into numerical, \\ binning)} & D \\
&& \makecell[l]{2.6 support filling in missing values} & D \\
&& \makecell[l]{2.7 Support creation of new attributes (i.e. calculated from other \\ attributes  e.g. affordability)}  & B\\
&& \makecell[l]{2.8 Ability to create/include own fairness metric (if not already in \\ system) } & B\\
&& \makecell[l]{2.9 Allow creation of subgroups based on a combination of \\ attributes and see their distribution on target} & B\\ \cline{2-4}
& \multirow{7}{*}{Adjust model} & \makecell[l]{2.10 Input custom thresholds to affect AI model} & B \\
&& \makecell[l]{2.11 Change weights on attributes to adjust AI model} & B\\
&& \makecell[l]{2.12 Remove attributes} & B\\ 
&& \makecell[l]{2.13 Change weightings of attribute variables (the variables that \\ make up the attribute) on attributes} & D\\
&& \makecell[l]{2.14 Identify similar attributes which  do not contain  protected \\ attributes and  substitute attribute from these choices} & D\\
&& \makecell[l]{2.15 Optimise model against fairness  metrics and accuracy \\ automatically} & D\\
&& \makecell[l]{2.16 Feedback to data scientists on ‘questionable’ attributes \\ that should  not be used for decision-making} & L\\\hline

3. Causal Graph & \multirow{3}{*}{Informational} & \makecell[l]{3.1 Node weight and impact on target} & B\\
&& \makecell[l]{3.2 Relationships between nodes, their ‘strength’ and direction} & B\\
&& \makecell[l]{3.3 Explanation of how the graph was derived} & B\\ \cline{2-4}
& Adjust model & \makecell[l]{3.4 The ability to remove nodes and relationships to adjust \\ AI model} & B\\\hline

4. Individual cases & \multirow{6}{*}{Informational} & \makecell[l]{4.1 specific application and attribute values} & B\\ 
&& \makecell[l]{4.2 fairness metric for individual case} & B\\  
&& \makecell[l]{4.3 Level of similarity between cases} & B\\  
&& \makecell[l]{4.4 Select specific cases to compare and show which attributes \\ are similar}  & B\\
&& \makecell[l]{4.5 Show decision boundaries} & B\\ \cline{2-4}
& Functional & \makecell[l]{4.6 See 'What If' results on target based on changes to \\ attribute values} & L \\\hline

5. Model & \multirow{3}{*}{Informational} & \makecell[l]{5.1 how model works} & B\\
&&\makecell[l]{ 5.2 how it was created, rationale for decisions in modelling}  & B\\
&& \makecell[l]{5.3 who created it}& B \\
\hline
\end{tabular}
}
\label{tab:RequirementsSummary}
\end{table}

\newpage

\begin{table}[h!]
\caption{Overview of evaluation participants' details.}
\centering
\ra{1.2}
\resizebox{.8\textwidth}{!}{
\begin{tabular}{@{}lcccccr@{}}
\toprule
ID	& Gender	& Age	&		Education	&	AI Fairness Knowledge & Role\\
\midrule

EL01	&	Female	&	37	&		Doctorate	&	I am aware of fair machine learning & Loan Officer \\
EL02	&	Female	&	40	&		Master	&	None & Loan Officer\\
EL03	&	Male	&	42	&		Master	&	None & Loan Officer\\
EL04	&	Male	&	43	&		Master	&	None & Loan Officer\\
EL05	&	Male	&	41	&		Master	&	I am aware of fair machine learning & Loan Officer\\
EL06	&	Female	&	30	&		Master	&	None & Loan Officer\\
EL07	&	Male	&	38	&		Master	&	None & Loan Officer\\
EL08	&	Male	&	33	&	Master	&	None & Loan Officer\\

ED01	&	Female	&	32	&		Master	&	I am aware of fair machine learning & Data Scientist\\
ED02	&	Female	&	29	&		Master	&	I actively work in fair machine learning & Data Scientist\\
ED03	&	Male	&	34	&		Doctorate	&	I am aware of fair machine learning & Data Scientist\\
ED04	&	Female	&	31	&		Master	&	None & Data Scientist\\
ED05	&	Male	&	34	&		Doctorate	&	I am aware of fair machine learning & Data Scientist\\
ED06	&	Male	&	35	&		Master	&	None & Data Scientist\\
ED07	&	Male	&	27	&		Master	&	I am aware of fair machine learning & Data Scientist\\
ED08	&	Female	&	37	&		Doctorate	&	\makecell[r]{I consider myself up to date with what's\\ in the fair machine learning space} & Data Scientist\\
ED09	&	Male	&	27	&		Master	&	I am aware of fair machine learning & Data Scientist\\
\bottomrule

\end{tabular}
}
\label{tab:participants_evaluation}
\end{table}

\newpage

\begin{table}[h!]
\caption{Requirements arising from design workshops covered by conventional tools and ours. In the Stakeholder row, L stands for the requirements from loan officers, D from data scientists, and B from both stakeholders. WiT, FV, FS, SV, FET, and HFIL stand for What-if Tool~\citep{Wexler2020What-IfTool}, FairVis \cite{Cabrera2019Fairvis}, FairSight~\citep{Ahn2020FairSight}, Silva~\citep{Yan2020SILVA}, Fairness Elicitation Tool~\citep{cheng_soliciting_2021}, and FairHIL (our tool), respectively.}
\centering
\ra{1.2}
\resizebox{\textwidth}{!}{
\begin{tabular}{@{}lllccccccc@{}}
\toprule
\textbf{Area} & \textbf{Use} & \textbf{Requirement} & \makecell[l]{\textbf{Stake-}\\\textbf{holder}} & \makecell[c]{\textbf{WiT}} & \makecell[c]{\textbf{FV}} & \makecell[c]{\textbf{FS}} & \makecell[c]{\textbf{SV}} & \makecell[c]{
    \textbf{FET}} & \makecell[c]{\textbf{FHIL}\\ (ours)} \\
\hline
\makecell[l]{1. Attribute\\overviews}  & \multirow{8}{*}{Informational}& \makecell[l]{1.1 attributes, number of records \\ and  attribute value  distributions} & B & \checkmark & \checkmark & \checkmark &  & \checkmark & \checkmark \\
&& \makecell[l]{1.2 amount of missing data}  & B &  &  &  &  &  &  \\
&& \makecell[l]{1.3 fairness metrics for model and \\ individual  protected attributes} & B  & \checkmark & \checkmark & \checkmark & \checkmark & \checkmark & \checkmark \\
&& \makecell[l]{1.4 target distribution} & B  & \checkmark & \checkmark &  &  &  & \checkmark \\  
&& \makecell[l]{1.5 protected attributes} & D  &  &  & \checkmark & \checkmark &  &\checkmark   \\
&& \makecell[l]{1.6 explanation how attribute \\ values are calculated or derived}  & D  &  &  &  &  &  &  \\
&& \makecell[l]{1.7 Identify if the data is \\ subjective/  objective} & D  &  &  &  &  &  &   \\
&& \makecell[l]{1.8 Identify where the data has \\ come from (applicant, bank, third \\ party) / attribute provenance} & D  &  &  &  &  &  &   \\
\hline

\makecell[l]{2. Investigate \\ Relationships \\ between attributes }  & \multirow{6}{*}{Informational} & \makecell[l]{2.1 distribution of protected\\ attributes with other attributes} & B & \checkmark &  & \checkmark &  &  & \checkmark \\
&& \makecell[l]{2.2 distribution of user-selected \\ attribute values (e.g. credit \\ risk  ratings) and  target values} & B & \checkmark &  & \checkmark &  &  & \checkmark \\ 
&& \makecell[l]{2.3 distribution of two user-\\selected attributes’  values} & B & \checkmark &  &  &  &  & \checkmark \\
&& \makecell[l]{2.4 Credit risk rating traffic light system} & D  &  &  &  &  &  &  \\ \cline{2-10}
& \multirow{8}{*}{Functional} & \makecell[l]{2.5 Support data transformations \\ (e.g. categorical into numerical, binning)} & D  & \checkmark &  &  &  &  &  \\
&& \makecell[l]{2.6 support filling in missing values} & D  & \checkmark &  &  &  &  &  \\
&& \makecell[l]{2.7 Support creation of new \\attributes (i.e.  calculated from other \\  attributes  e.g. affordability)}  & B & \checkmark &  &  &  &  & \checkmark \\
&& \makecell[l]{2.8 Ability to create/include own fairness \\ metric (if  not already in system) } & B &  &  &  &  &  & \checkmark \\
\hline
\makecell[l]{2. Investigate\\ Relationships\\between attributes}&\multirow{1}{*}{Functional}& \makecell[l]{2.9 Allow creation of subgroups based \\ on a combination of attributes and \\ see their distribution on target} & B & \checkmark & \checkmark &  &  & \checkmark & \checkmark \\ \cline{2-10}
& \multirow{11}{*}{Adjust model} & \makecell[l]{2.10 Input custom thresholds to affect AI model} & B  &  &  &  &  &  &  \\
&& \makecell[l]{2.11 Change weights on  attributes  to \\ adjust  AI model} & B & \checkmark &  &  &  &  &  \\
&& \makecell[l]{2.12 Remove attributes} & B & \checkmark &  &  &  &  &  \\ 
&& \makecell[l]{2.13 Change weightings of attribute \\ variables  (the  variables that make \\ up the  attribute) on attributes} & D & \checkmark &  &  &  &  &  \\
&& \makecell[l]{2.14 Identify similar  attributes which \\  do not contain  protected attributes \\ and  substitute attribute from these  choices} & D &  &  &  &  &  &  \\
&& \makecell[l]{2.15 Optimise model against fairness  \\ metrics and accuracy  automatically} & D &  &  &  &  &  &  \\
&& \makecell[l]{2.16 Feedback to data scientists on \\ ‘questionable’ attributes that should  \\ not be used  for decision-making} & L &  &  &  &  &  &  \\\hline

3. Causal Graph & \multirow{3}{*}{Informational} & \makecell[l]{3.1 Node weight and  impact on target} & B &  &  &  &  &  &  \checkmark\\
&& \makecell[l]{3.2 Relationships between nodes, their \\ ‘strength’ and direction} & B &  &  &  & \checkmark &  & \checkmark \\
&& \makecell[l]{3.3 Explanation of how  the graph was derived} & B &  &  &  &  &  &  \\ \cline{2-10}
& Adjust model & \makecell[l]{3.4 The ability to remove nodes and \\ relationships to adjust  AI model} & B &  &  &  &  &  &  \\\hline

4. individual cases & \multirow{6}{*}{Informational} & \makecell[l]{4.1 specific application and attribute values} & B & \checkmark &  & \checkmark & \checkmark & \checkmark & \checkmark \\ 
&& \makecell[l]{4.2 fairness metric for individual case} & B &  &  &  &  &  &  \\  
&& \makecell[l]{4.3 Level of similarity between cases} & B &  &  &  &  & \checkmark & \checkmark \\  
&& \makecell[l]{4.4 Select specific cases to compare \\ and show which attributes are similar}  & B &  &  & \checkmark &  & \checkmark & \checkmark \\
&& \makecell[l]{4.5 Show decision boundaries} & B & \checkmark &  &  &  &  & \checkmark \\ \cline{2-10}
& Functional & \makecell[l]{4.6 See 'What If' results on target \\ based  on changes to  attribute values} & L  & \checkmark &  &  &  &  &  \\\hline

5. Model & \multirow{3}{*}{Informational} & \makecell[l]{5.1 how model works} & B &  &  &  &  &  & \checkmark \\
&&\makecell[l]{ 5.2 how it was created, rationale for \\ decisions  in  modelling}  & B &  &  &  &  &  &  \\
&& \makecell[l]{5.3 who created it}& B  &  &  &  &  &  &  \\
\hline
\end{tabular}
}
\label{tab:DesignSpace}
\end{table}

\newpage

\clearpage
\begin{description}
    \item[Figure 1.]Human-in-the-loop Fairness Toolkits
    \item[Figure 2.]Workshop causal graph example.
    \item[Figure 3.]Initial prototype.
    \item[Figure 4.]Sankey graph showing relationships between nationality and credit level.
    \item[Figure 5.]Main UI Components.
    \item[Figure 6.]Setup: metrics.
    \item[Figure 7.]Metric builder.
    \item[Figure 8.]Causal graph in the dataset view showing a selected node: "net monthly income"
    \item[Figure 9.]Causal graph in the model view showing feature importance (saturation).
    \item[Figure 10.]Detailed Information on Features.
    \item[Figure 11.]Detailed Information on Relationships.
    \item[Figure 12.]Feature Combinations Component.
    \item[Figure 13.]Dataset Component in Dataset View.
    \item[Figure 14.]Dataset Component in Model View.
    \item[Figure 15.]Compare Application component.
    \item[Figure 16.]Mean Post-session Usefulness Questionnaire Responses. Error bars show standard deviation.
    \item[Figure 17.]Mean Post-session NASA-TLX Responses. Error bars show standard deviation.
\end{description}

\end{document}